\documentclass[11pt, a4paper,  notitlepage]{extarticle}

\usepackage[left=3cm, right=3cm, top=3.5cm, bottom=3.5cm, bindingoffset=0cm]{geometry}
\usepackage[T2A]{fontenc}
\usepackage[utf8]{inputenc}
\usepackage{graphicx}
\usepackage{mathtools}
\usepackage[affil-it]{authblk}
\usepackage{url}
\usepackage{amsmath}
\usepackage{amssymb}
\usepackage{multicol}
\usepackage{float}
\usepackage[section]{placeins}
\usepackage{listings}
\usepackage{changepage}
\usepackage{hyperref}
\numberwithin{equation}{section}
\numberwithin{figure}{section}
\numberwithin{table}{section}
\title{Analytics of Business Time Series Using Machine Learning and Bayesian Inference}
\author{Bohdan M.  Pavlyshenko  \\ Ivan Franko National University of Lviv, Ukraine \\ b.pavlyshenko@gmail.com,  \href{http://www.linkedin.com/in/bpavlyshenko/}{linkedin.com/in/bpavlyshenko/}}

\begin{document}

\date{}
\maketitle
\sloppy

In the survey we  consider the case studies on sales time series forecasting~\cite{pavlyshenko2019machine, pavlyshenko2016linear,pavlyshenko2018using,pavlyshenko2020bayesian, pavlyshenko2020using},
 the deep learning approach for forecasting non-stationary time series using time trend correction~\cite{pavlyshenko2022forecasting},
dynamic price and supply optimization using Q-learning~\cite{pavlyshenko2020salests},
Bitcoin price modeling~\cite{pavlyshenko2019bitcoin}, COVID-19 spread impact on stock market~\cite{pavlyshenko2020regression}, using social networks signals in analytics~\cite{pavlyshenko2021forming,pavlyshenko2022methods,
pavlyshenko2019forecasting}.  The use of machine learning and Bayesian inference in predictive  analytics has been analyzed.
\tableofcontents
\section{Machine Learning  Models for  Sales Time Series~Forecasting}
In this case study,  we consider the usage of machine-learning models for sales predictive analytics. The main goal of this paper is to consider main approaches and case studies of using machine learning for sales forecasting. 
The effect of machine learning generalization has been considered.  This effect can be used  to make sales predictions when there is a small amount of historical data for specific sales time series in the case when a new  product or store is launched. A stacking approach for building regression ensemble of single models has been studied.  
The results show that using stacking techniques, we can improve the performance of predictive models for sales time series forecasting.

\subsection{Introduction}
Sales prediction is an important part of modern business intelligence~\cite{mentzer2004sales,efendigil2009decision,zhang2004neural}. 
It can be a  complex problem, especially in the case of lack of data, 
missing data,  and the presence of outliers.
 Sales~can be considered as a time series. At present time, different time series  models have been developed, for~example, by Holt-Winters, ARIMA, SARIMA, SARIMAX, GARCH,  etc. 
Different time series approaches can be found in~\cite{chatfield2000time,brockwell2002introduction,box2015time,doganis2006time,hyndman2018forecasting,
tsay2005analysis,wei2006time,cerqueira2018arbitrage,hyndman2007automatic,papacharalampous2017comparison,tyralis2017variable,
tyralis2018large}.
 In~\cite{papacharalampous2018predictability} authors investigate the predictability of time series, and study the performance of
different time series forecasting methods.  In~\cite{taieb2012review},  different approaches for 
multi-step ahead time series forecasting are considered and compared. In~\cite{graefe2014combining}, different forecasting methods combining have been investigated. 
It is shown that in the case when different models are based on different algorithms and data, one can receive 
essential gain in the accuracy. Accuracy improving is essential in the cases with large uncertainty.
In~\cite{wolpert1992stacked, rokach2010ensemble, sagi2018ensemble, gomes2017survey, dietterich2000ensemble, rokach2005ensemble},   
 different ensemble-based methods for classification problems are considered. 
In~\cite{armstrong1989combining}, it is shown that by combining forecasts produced by different algorithms, it is possible to improve forecasting accuracy. 
In the work, different conditions  for effective forecast combining were considered. 
In~\cite{papacharalampous2018univariate}   authors considered 
lagged variable selection, hyperparameter optimization, comparison between classical algorithms and machine learning 
based algorithms for time series. On the  temperature time series datasets, the authors showed that classical algorithms and machine-learning-based 
algorithms can be equally used.  
There are some limitations of time series approaches for sales forecasting. Here are some of them:
\begin{itemize} 
\item We need to have historical data for a long time period to capture seasonality. 
However, often we do not have historical data for  a target variable, for example in case when a new product is launched. 
At the same time we have sales time series for a similar product and we can expect that our new product will have  a similar  sales pattern. 
\item Sales data can have  a lot of outliers and missing data. We must clean outliers and interpolate data before using a time series approach.
\item We need to take into account a lot of exogenous factors which have impact on sales.   
\end{itemize}

Sales prediction is rather a regression problem than a time series problem. 
Practice shows that the use of regression approaches can often give us  better results compared to time series methods. 
Machine-learning algorithms make it possible to find patterns in the time series.
We can find complicated patterns in the sales dynamics, using supervised machine-learning methods. Some of the most popular are  tree-based machine-learning algorithms~\cite{james2013introduction}, e.g., Random Forest~\cite{breiman2001random}, Gradiend Boosting Machine~\cite{friedman2001greedy, friedman2002stochastic}.  
One of the main assumptions of regression methods is that the patterns in the past data will be repeated in future. 
 In~\cite{pavlyshenko2016linear}, we studied linear models, machine learning, and probabilistic models for time series modeling. 
For probabilistic modeling, we considered the use of copulas and Bayesian inference approaches. 
In~\cite{pavlyshenko2016machine}, we studied the logistic regression in the problem of  detecting manufacturing failures. 
For logistic regression, we considered a generalized linear model, machine learning  and Bayesian models. 
In~\cite{pavlyshenko2018using}, we studied stacking approaches for time series forecasting and logistic regression with highly imbalanced data. 
 In the sales data, we can observe  several types of patterns and effects.  They are: trend, seasonality, autocorrelation, patterns caused by the impact of such external factors as promo, pricing, competitors' behavior. We also observe noise in the sales. Noise is caused by the factors which are not included into our consideration. In the sales data, we  can also observe extreme values -- outliers. If we need to perform risk assessment, we should to take into account noise and extreme values. 
Outliers can be caused by some specific factors, e.g., promo events, price reduction, weather conditions, etc. If these specific events are repeated periodically, we~can add a new feature which will indicate these special events and describe the extreme values of 
the target variable.  

In this work, we study the usage of machine-learning models for sales time series forecasting. \mbox{We will} consider a single  model, the effect of machine-learning generalization and stacking of multiple~models.

\subsection{Machine Learning Predictive Models}
\begin{figure}[H]
\center
\includegraphics[width=0.85\linewidth]{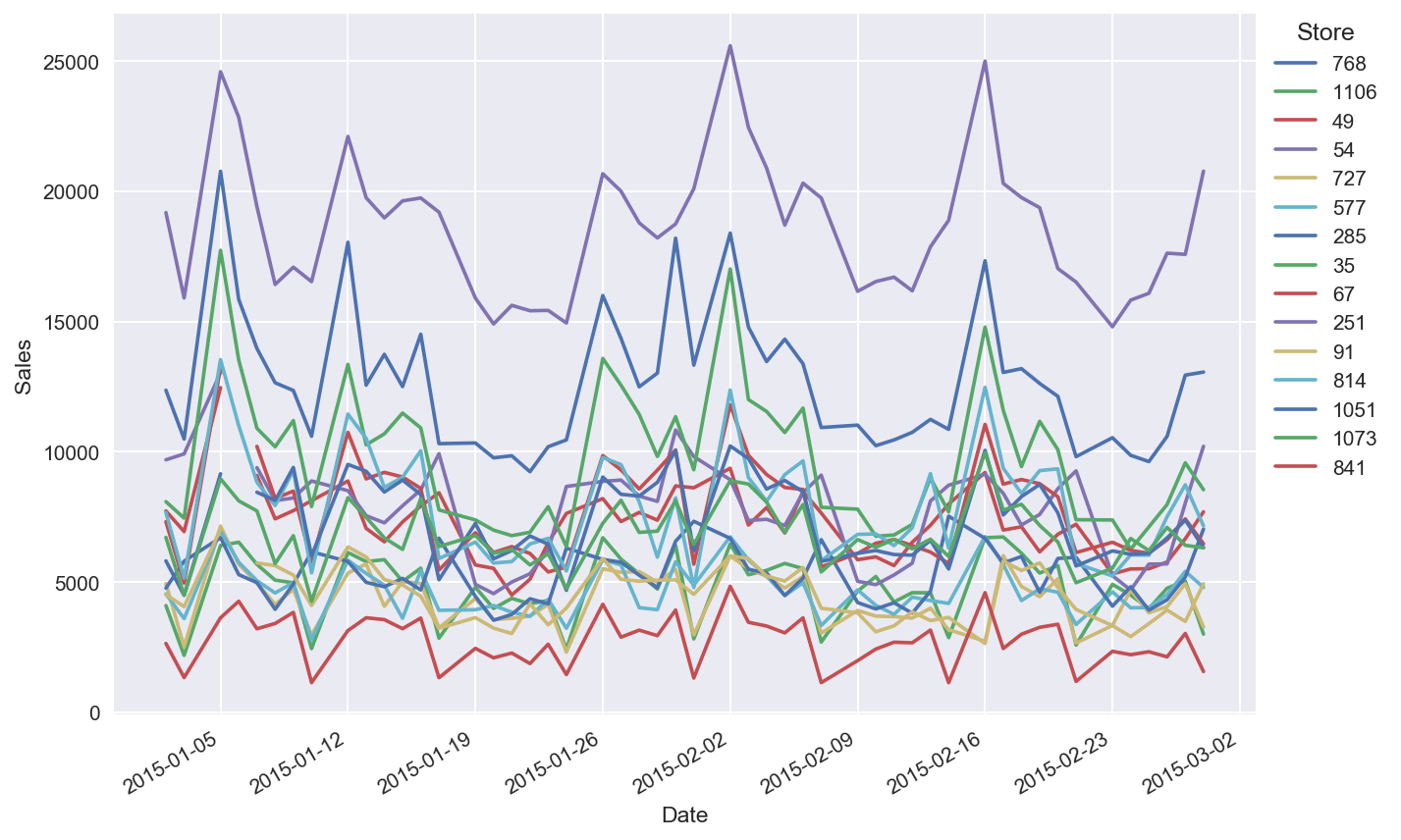}
\caption{Typical time series for sales.}
\label{ml_ts_fig1}
\end{figure}

\begin{figure}[H]
\center
\includegraphics[width=0.75\linewidth]{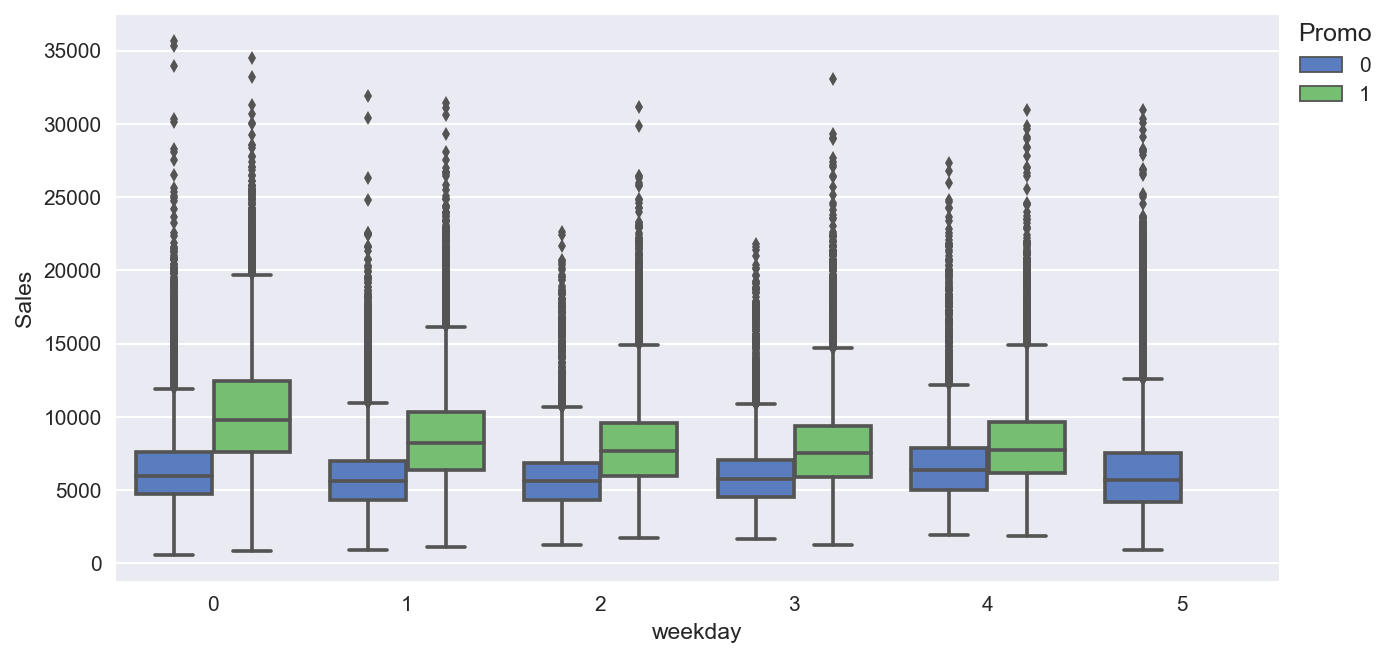}
\caption{Boxplots for sales distribution vs. day of week.}
\label{ml_ts_fig2}
\end{figure}

\begin{figure}[H]
\center
\includegraphics[width=0.75\linewidth]{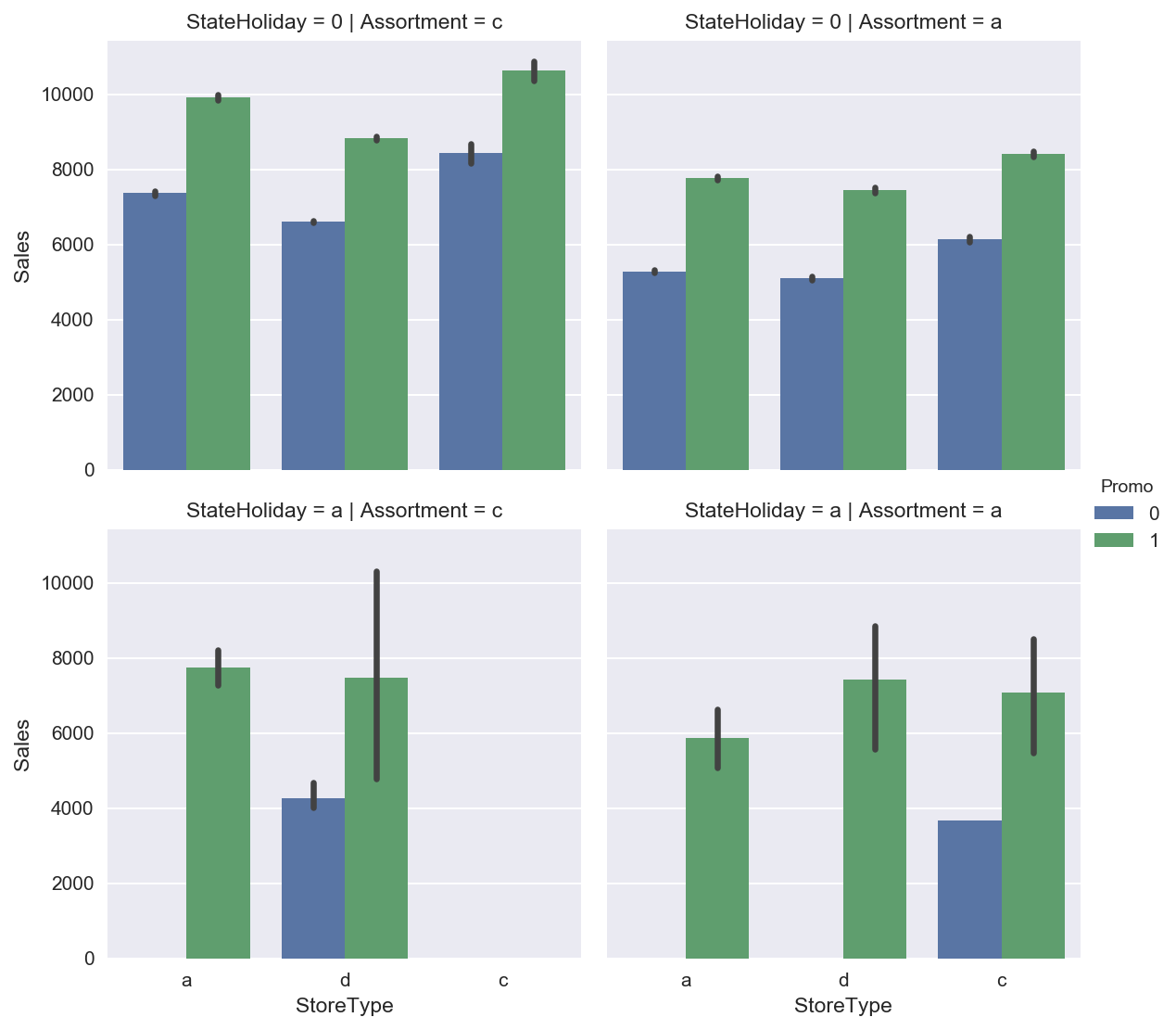}
\caption{Factor plots for aggregated sales.}
\label{ml_ts_fig3}
\end{figure}

\begin{figure}[H]
\center
\includegraphics[width=0.85\linewidth]{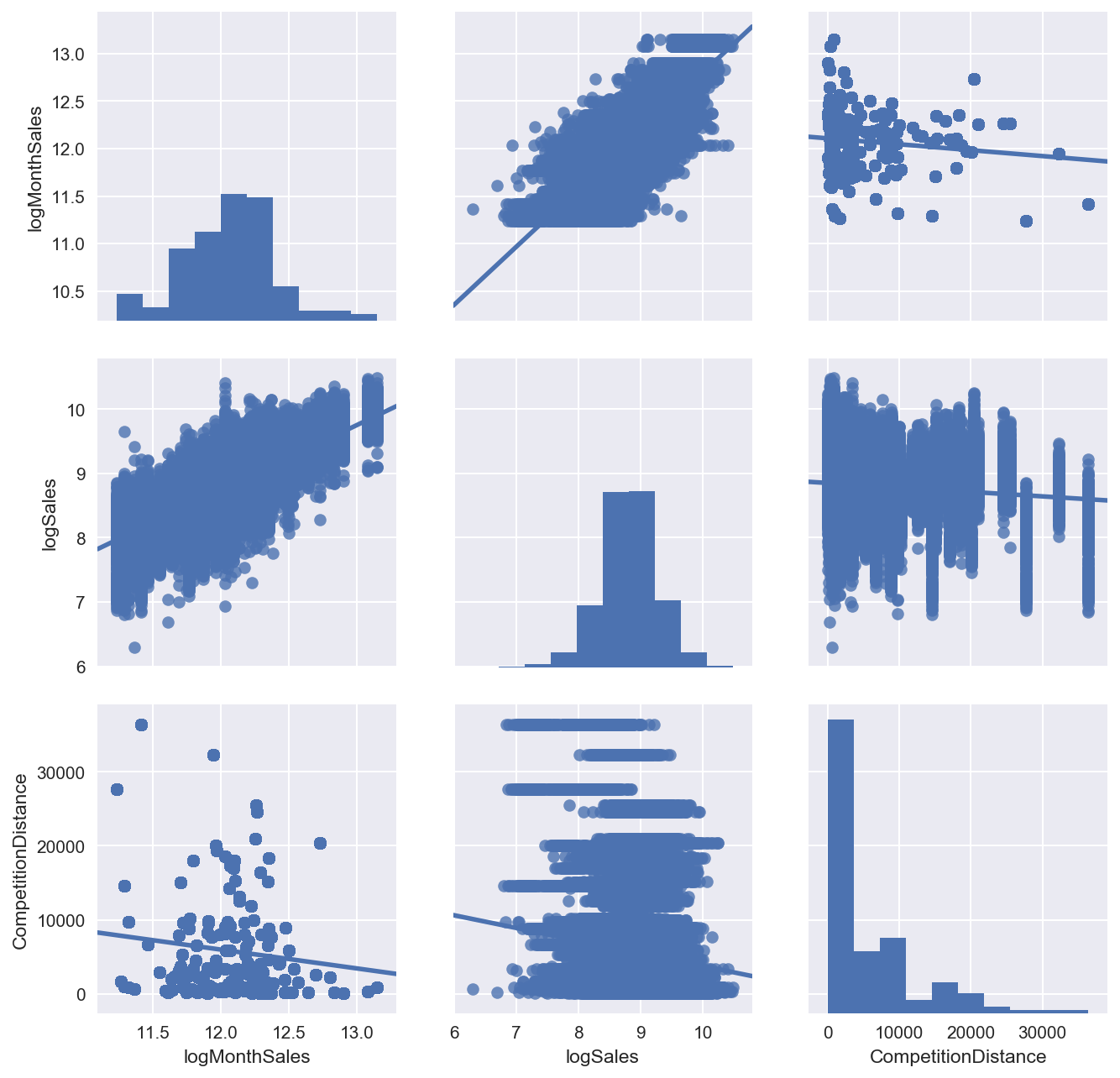}
\caption{Pair plots with log(MonthSales), log(Sales), CompetitionDistance.}
\label{ml_ts_fig4}
\end{figure}

For our analysis, we used store sales historical data from  ``Rossmann Store Sales'' Kaggle competition~\cite{rossmanstorekaggle}. These data describe sales in Rossmann stores.  The calculations were conducted in the Python environment using the main packages \textit{pandas, sklearn, numpy, keras, matplotlib, seaborn}.  To~conduct the analysis, \textit{Jupyter Notebook} was used. 
Figure~\ref{ml_ts_fig1}  shows  typical time series for sales, values~of sales are normalized arbitrary units.  
Firstly,  we conducted the descriptive analytics, which is a study of sales distributions, \mbox{data visualization} with different pairplots. It is helpful in finding 
correlations and  sales drivers on which we focus. Figures~\ref{ml_ts_fig2}--\ref{ml_ts_fig4} show the  results of the exploratory analysis. 
\vspace{-12pt}

\begin{figure}[H]
\center
\includegraphics[width=0.85\linewidth]{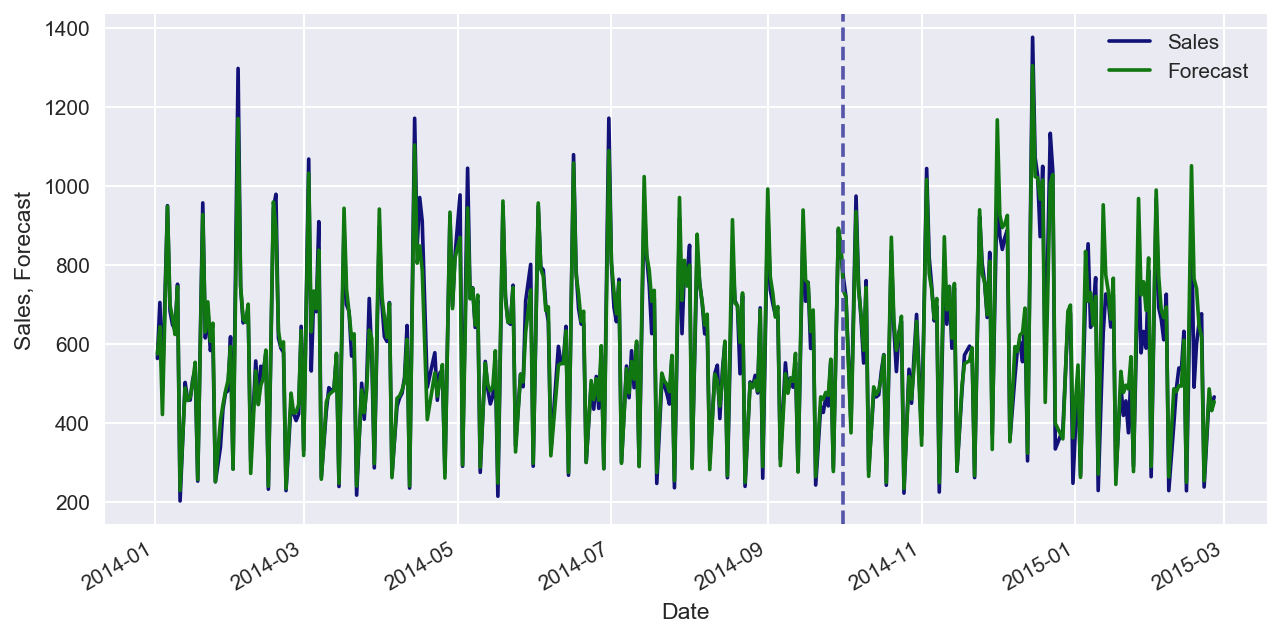}
\caption{Sales forecasting (train set error: 3.9\%, validation set error: 11.6\%).}
\label{ml_ts_fig5}
\end{figure}

\begin{figure}[H]
\center
\includegraphics[width=0.5\linewidth]{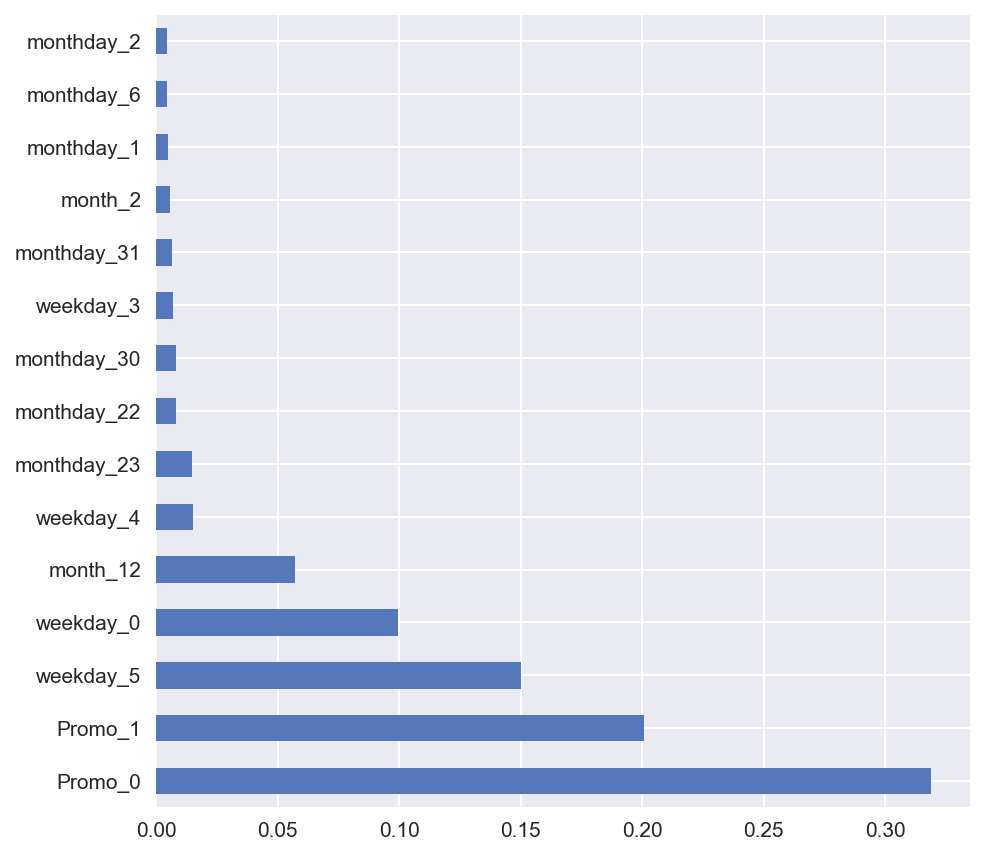}
\caption{Feature importance.}
\label{ml_ts_fig6}
\end{figure}

\begin{figure}[H]
\center
\includegraphics[width=0.75\linewidth]{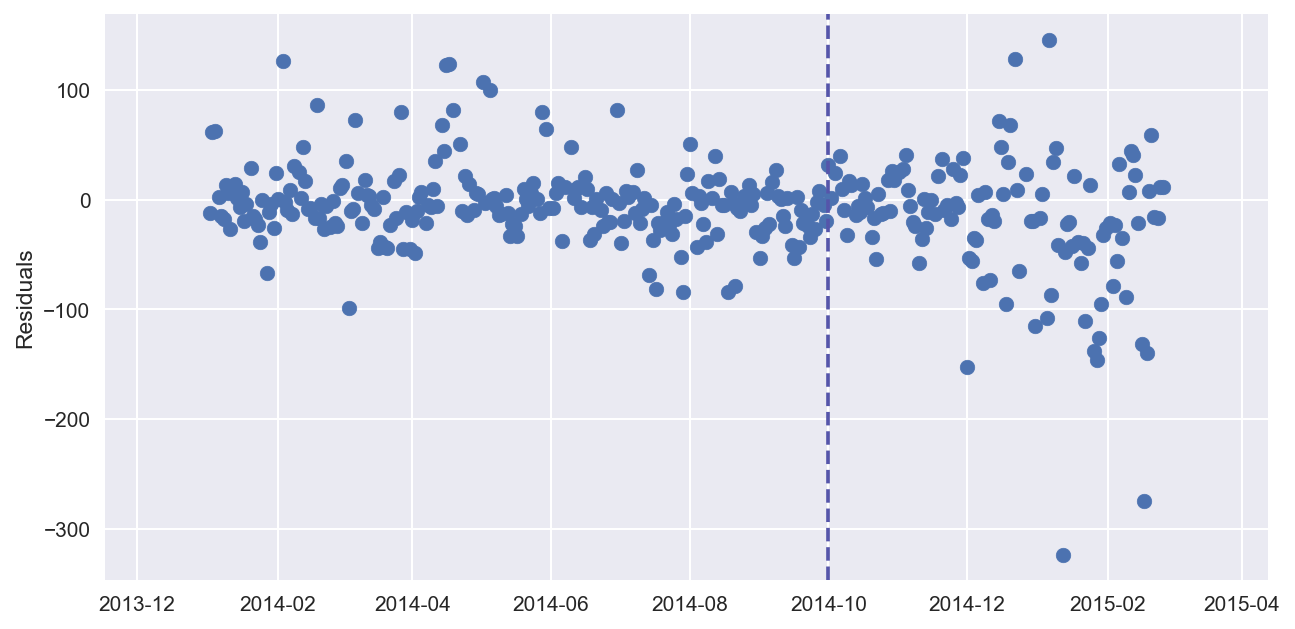}
\caption{Forecast residuals for sales time series.}
\label{ml_ts_fig7}
\end{figure}

\begin{figure}[H]
\center
\includegraphics[width=0.75\linewidth]{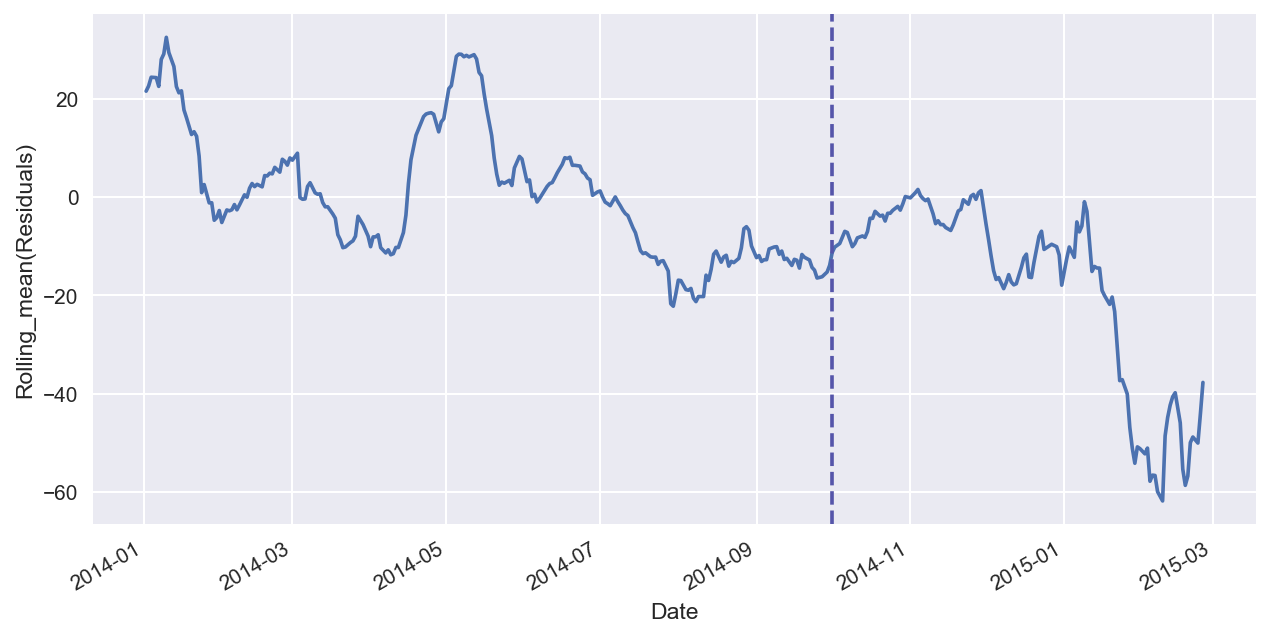}
\caption{Rolling mean of residuals.}
\label{ml_ts_fig8}
\end{figure}

\begin{figure}[H]
\center
\includegraphics[width=0.75\linewidth]{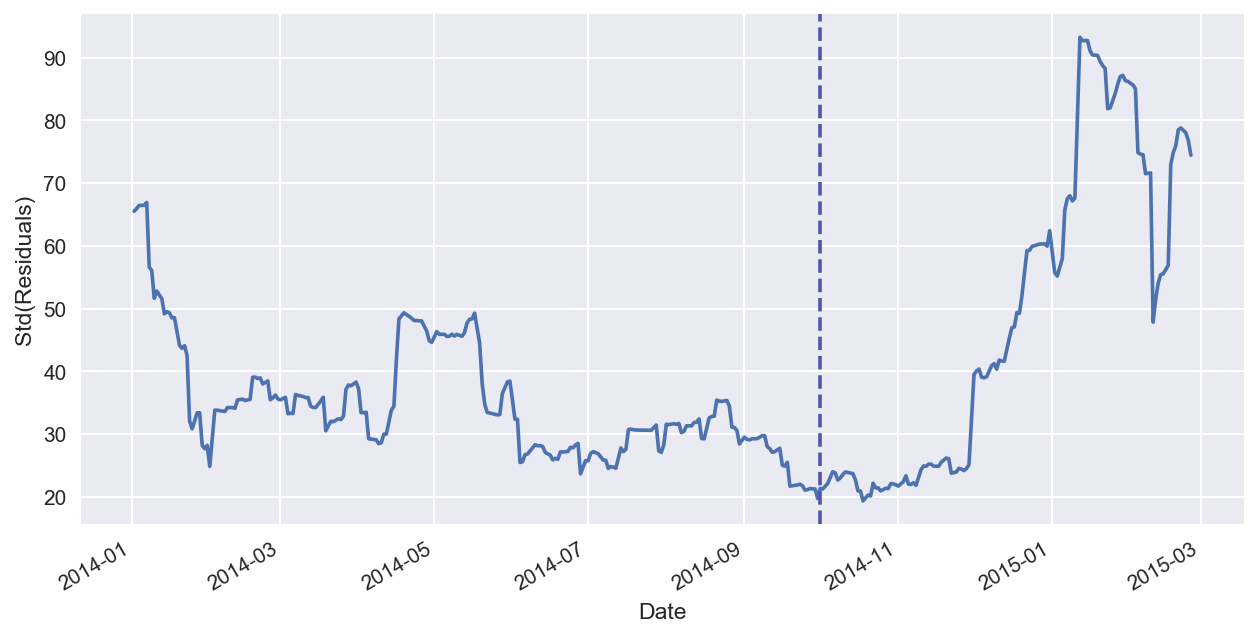}
\caption{Standard deviation of forecast residuals.}
\label{ml_ts_fig9}
\end{figure}

A specific feature of  most machine-learning methods is that they can work with stationary data only. In case of a small trend, we can find bias using linear regression on the validation set. 
Let us consider the supervised machine-learning approach using sales historical time series. For the case study, we used Random Forest  algorithm~\cite{breiman2001random}. As covariates, we used categorical features: promo, day of week, day of month, month. For categorical features, we applied one-hot encoding, when one 
categorical variable was replaced by n binary variables, where n is the amount of unique values of categorical variables.
Figure~\ref{ml_ts_fig5} shows the forecasts of sales time series.  Figure~\ref{ml_ts_fig6} shows the feature importance.
For error estimation, we used a relative mean absolute error (MAE) which is calculated as  $error=MAE/mean(Sales) \cdot 100\%$.
\vspace{-12pt}
Figure~\ref{ml_ts_fig7} shows forecast residuals for sales time series, Figure~\ref{ml_ts_fig8} shows the rolling mean of residuals, 
Figure~\ref{ml_ts_fig9} shows the standard deviation of forecast residuals.
\vspace{-12pt}

In the forecast, we may observe bias on validation set which is a constant (stable) under- or over-valuation of sales when the forecast is going to be 
higher or lower with respect to real values. It~often appears when we apply machine-learning methods to  non-stationary sales.  We can conduct the correction of bias using linear regression on the validation set. We must differentiate the accuracy on a validation set from the accuracy on a training set. On the training set, it can be very high but  on the validation set it is low. 
The accuracy on the validation set is an important indicator for choosing an optimal number of iterations of machine-learning algorithms.

\subsection{Effect of Machine Learning Generalization}
\begin{figure}[H]
\center
\includegraphics[width=0.75\linewidth]{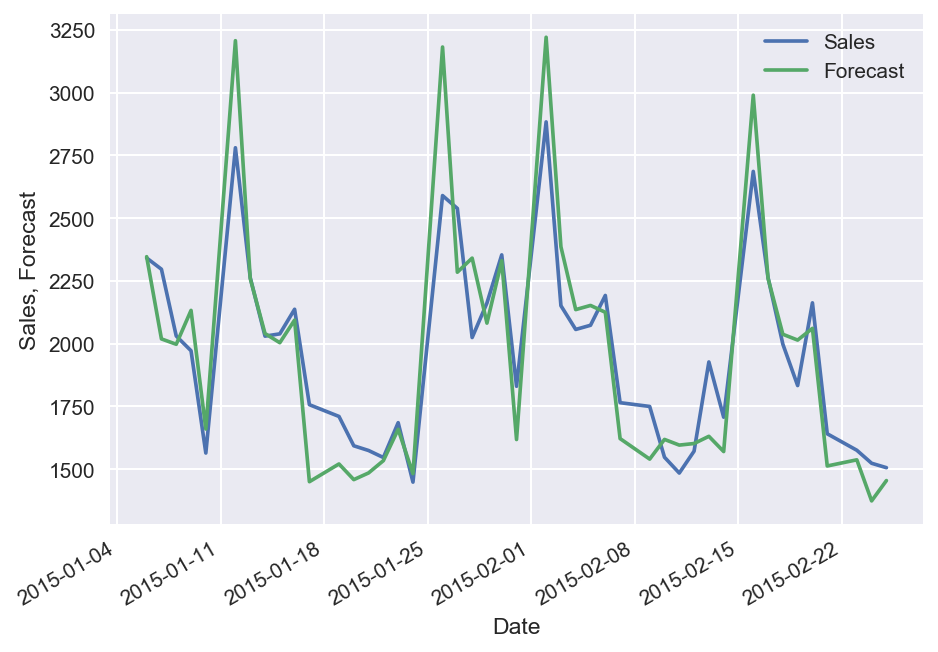}
\caption{Sales forecasting with long time (2 year) historical data, error = 7.1\%.}
\label{ml_ts_fig10}
\end{figure}

\begin{figure}[H]
\center
\includegraphics[width=0.75\linewidth]{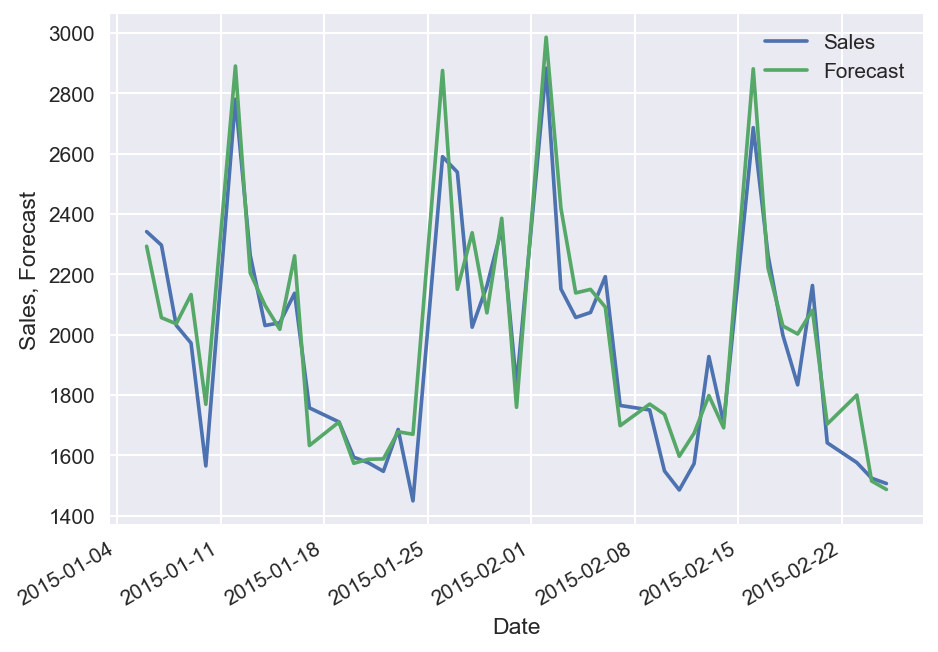}
\caption{Sales forecasting with short time  (3 days), historical data, error = 5.3\%.}
\label{ml_ts_fig11}
\end{figure}

The effect of machine-learning generalization consists in the fact that a regression algorithm captures the patterns which exist in the whole set of stores or products. 
If the sales have expressed patterns, then generalization enables us to get more precise results which  are resistant to sales noise. 
 In the case study of machine-learning generalization, we used the following additional features regarding the previous case study: mean sales value for a  specified time period of historical data, state~and school holiday flags, distance from store to competitor's store, store assortment type.
Figure~\ref{ml_ts_fig10} shows the forecast in the case of  historical data with a long time period (2 years) for a specific store,
Figure~\ref{ml_ts_fig11} shows the forecast in the case of  historical data with a short time period (3 days) for the same specific store.
\vspace{-12pt}
In case of short time period, we can receive even more precise results. 
The effect of machine-learning generalization enables us to make prediction in case of very small number of historical sales data, which is important when we launch a new product or store. 
 If we are going to predict the sales for new products, we can make 
expert correction by multiplying the prediction by a time dependent coefficient to take into account the transient processes,
 e.g., the process of product cannibalization when new products substitute other products. 
\subsection{Stacking of Machine Learning Models}
\begin{figure}[H]
\center
\includegraphics[width=1\linewidth]{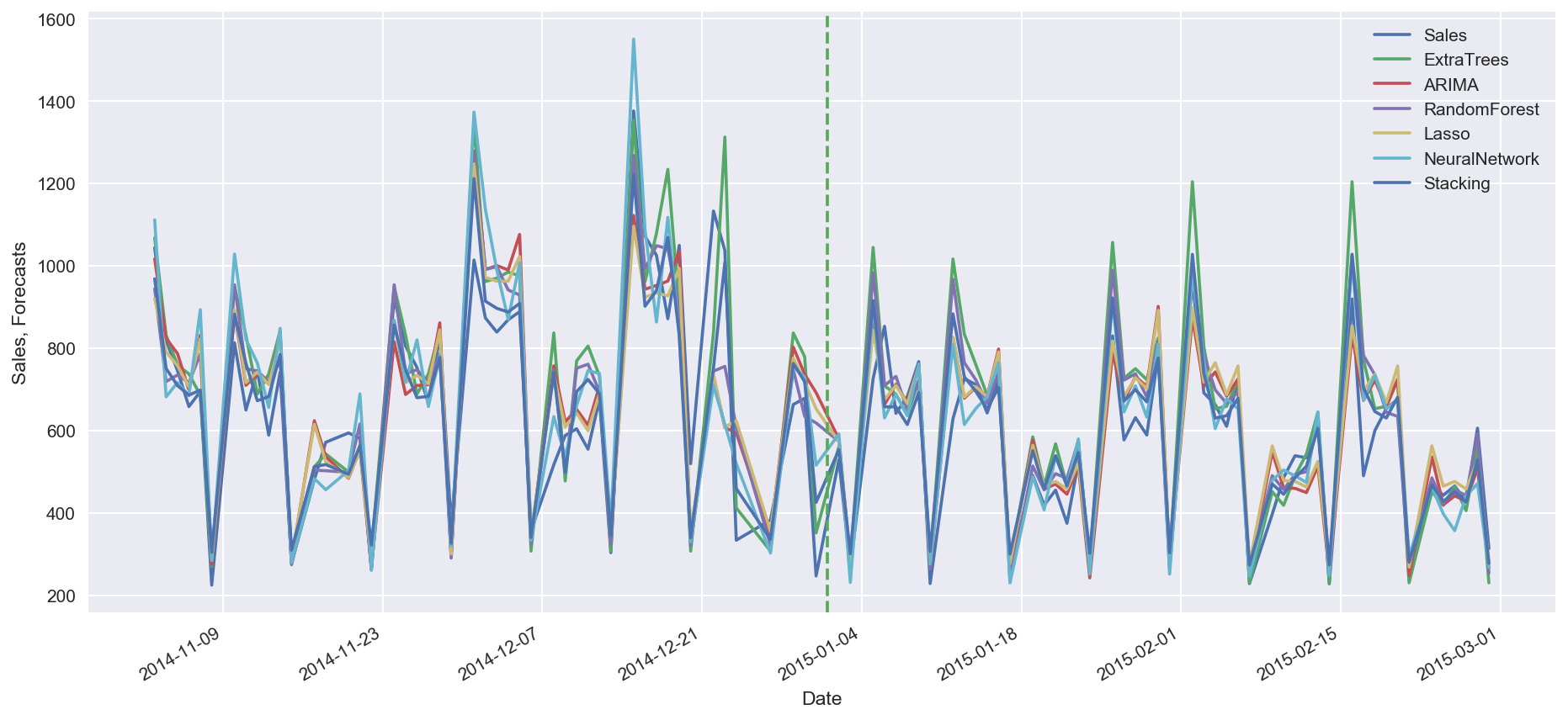}
\caption{Time series forecasting on the validation sets obtained using different models.}
\label{ml_ts_fig12}
\end{figure}

\begin{figure}[H]   \center
\includegraphics[width=0.5\linewidth]{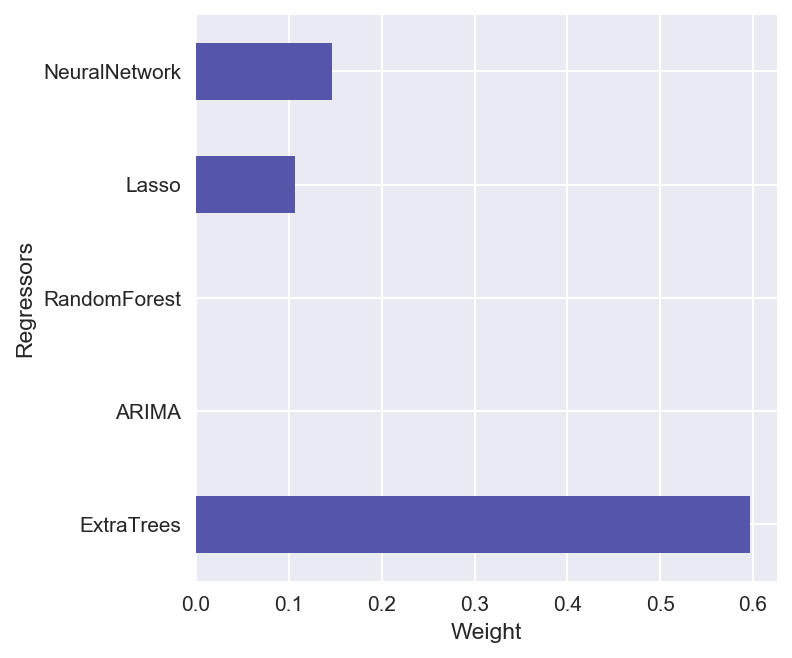}
\caption{Stacking weights for regressors.}
\label{ml_ts_fig13}
\end{figure}

Having different predictive models with different sets of features, it is useful to combine all these results into one.   Let us consider the stacking techniques~\cite{wolpert1992stacked, rokach2010ensemble, sagi2018ensemble, gomes2017survey, dietterich2000ensemble, rokach2005ensemble} for building ensemble of predictive models.  In such an approach, the results of predictions on the validation set are treated as input regressors for the next level models.  As the next level model, we can consider a linear model or another type of a machine-learning algorithm, e.g., Random Forest  or Neural Network. 
  It is important to mention that in case of time series prediction, we cannot use a conventional cross validation approach, we have to split a historical data set on the training set and validation set by using period  splitting, so the training data will lie in the first time period and the validation set in the next one. 
Figure~\ref{ml_ts_fig12} shows the time series forecasts on the validation sets obtained using different models. 
Vertical dotted line on the Figure~\ref{ml_ts_fig12} separates the validation set and out-of-sample set which is not used in 
the model training and validation processes. On the  out-of-sample set, one can calculate stacking errors. 
hl{Predictions on the} validation sets are treated as regressors for the linear model with Lasso regularization. Figure~\ref{ml_ts_fig13} shows the results obtained on the second-level Lasso regression model.
Only three models from the first level (ExtraTree, Lasso, Neural Network) have non-zero coefficients for their results. For other cases of  sales datasets, the results can be different when the other models can play more essential role in the forecasting. 
Table~\ref{tab1} shows the errors on the validation and out-of-sample sets. 
These results show that stacking approach can improve accuracy on the validation and on the out-of-sample sets. 

\begin{figure}[H]
\centerline{\includegraphics[width=0.75\textwidth]{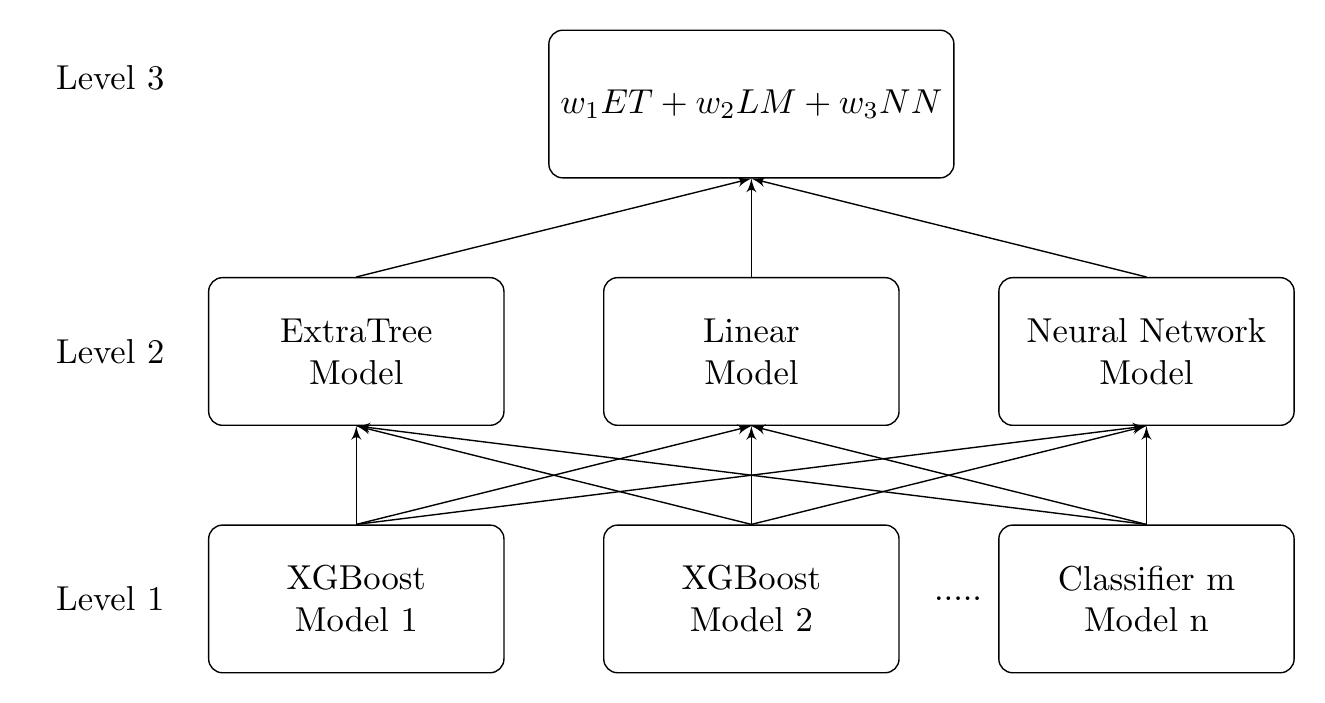}}
\caption{Mulitilevel machine-learning model for sales time series forecasting.}
\label{ml_ts_fig14}
\end{figure}

\FloatBarrier
\begin{table}[H]
\caption{Forecasting errors of different models.}
\centering
\begin{tabular}{l c c}
\hline
\textbf{Model} & \textbf{Validation Error} & \textbf{Out-of-Sample Error} \\
\hline
ExtraTree & 14.6\% & 13.9\% \\
ARIMA & 13.8\% &  11.4\% \\
RandomForest & 13.6\% & 11.9\% \\
Lasso & 13.4\% & 11.5\% \\
\hline
Stacking & 12.6\% & 10.2\% \\
\end{tabular}
\label{tab1}
\end{table}
To get insights and to find new approaches, some companies  propose their analytical problems for data science competitions, e.g., at Kaggle~\cite{kaggle}. 
One of such competitions was  Grupo Bimbo Inventory Demand \cite{compgruppobimbodemand}. 
The challenge of this competition was to predict inventory demand.
  I was a teammate of a great team 'The Slippery Appraisals'  which took the first place on this competition. 
The details of our winner solution are  at~\cite{compgruppobimbsolution}.
Our solution is based on three level model  (Figure~\ref{ml_ts_fig14}). On the first level, we used many single models, most of them 
 were based on XGBoost machine-learning algorithm~\cite{chen2016xgboost}. For the second stacking level, we used two models from Python scikit-learn package - ExtraTree model and  linear model from, as well as  Neural Network model.
 The results from the second level were summed with weights on the third level. We constructed a lot of new features, the most important of them were 
 based on aggregating target variable and its lags with grouping by different factors.   
More details can be found at~\cite{compgruppobimbsolution}. A simple R script with single machine-learning model is  at~\cite{gruppobimboxgboostscript}. 

\FloatBarrier

\subsection{Conclusions}
In our case study, we considered different machine-learning approaches for time series forecasting. 
Sales prediction is rather a regression problem than a time series problem. 
The use of regression approaches for sales forecasting can often give us  better results compared to time series methods. 
One of the main assumptions of regression methods is that the patterns in the historical data will be repeated in future.
 The accuracy on the validation set is an important indicator for choosing an optimal number of iterations of machine-learning algorithms. 
The effect of machine-learning generalization consists in the fact of  capturing the patterns in the whole set of data.
This effect can be used  to make sales prediction when there is a small number of historical data for specific sales time series in the case when a new  
product or store is launched.  In stacking  approach, the results of multiple model predictions on the validation set are treated 
as input regressors for the next level models.  As the next level model, Lasso regression can be used.  Using stacking  
makes it possible to take into account the differences in the results for multiple models with different sets of parameters and 
improve accuracy on the validation and on the out-of-sample data sets.

\section{Bayesian Regression Approach  for  Building And Stacking       Predictive Models}
In this case study,   we consider the use of Bayesian regression for  building time series models and  stacking   different predictive models for time series. Using Bayesian regression for  time series modeling with nonlinear trend was analyzed. 
This  approach makes it possible to estimate an uncertainty of time series prediction and  calculate value at risk  characteristics. 
A hierarchical model for time series using Bayesian regression has been considered. 
In this approach, one set of parameters is the same for all data samples, other parameters can be different for different groups of data samples. Such an approach 
 allows  using this  model in the case of short historical data for specified time series, e.g. in the case of new stores or new products in the sales prediction problem. 
In the study of predictive models stacking, the models  ARIMA, Neural Network, Random Forest, Extra Tree were used for the prediction on the first level of model ensemble. 
On the second  level, time series predictions of these models on the validation set were used for stacking by Bayesian regression. This approach gives distributions for regression coefficients of these models. It makes it possible to estimate  the uncertainty contributed by each model  to stacking result.  
The information about these distributions allows us to select an optimal set of stacking models, taking into account the domain knowledge.  The probabilistic approach for stacking predictive 
models allows us to make risk assessment for the predictions that are important in a decision-making process.

\subsection{Introduction}

We can consider time series forecasting as a regression problem in many cases, 
especially for sales time series. 
 In ~\cite{pavlyshenko2016linear}, we considered linear models, machine learning and probabilistic models for time series modeling. 
In~\cite{pavlyshenko2016machine}, we regarded the logistic regression with Bayesian inference for analysing  manufacturing failures. 
In ~\cite{pavlyshenko2019machine}, we study the use of machine learning models for sales predictive analytics. We researched the main approaches and case studies of implementing machine learning to sales forecasting. 
The effect of machine learning generalization has been studied.  This effect can be used for predicting sales in case of a small number of historical data for specific sales time series in case when a new product or store is launched ~\cite{pavlyshenko2019machine}. A stacking approach for building ensemble of single models using Lasso regression has also been studied.  
The obtained results show that using stacking techniques, we can improve the efficiency of predictive models for sales time series forecasting ~\cite{pavlyshenko2019machine}.

Probabilistic regression models can be based on  Bayesian theorem ~\cite{kruschke2014doing,gelman2013bayesian, carpenter2017stan}. This approach allows us to receive a posterior distribution of model parameters using conditional likelihood and prior distribution. 
Probabilistic approach is more natural for stochastic variables such as sales time series. The difference between Bayesian approach and conventional Ordinary Least Squares (OLS) method is that in the Bayesian approach, uncertainty comes from parameters of model, as opposed to OLS method where the parameters are constant and uncertainty comes from data. In the Bayesian inference, we can use informative prior distributions which can be set up by an expert. So, the result can be considered as a compromise between 
historical data and expert opinion. It is important in the cases when we have a small number of historical data. In the Bayesian model, we can consider the target variable with non Gaussian  distribution, e.g. Student's t-distribution. 
Probabilistic approach enables us an ability to receive probability density function for the target variable. Having such function, we can make risk assessment and calculate value at risk (VaR) which is 5\% quantile. 
For solving Bayesian models, the numerical Monte-Carlo methods are used. Gibbs and Hamiltonian sampling are the popular methods of finding posterior distributions for the parameters of probabilistic model
~\cite{kruschke2014doing,gelman2013bayesian, carpenter2017stan}.
 
 In this work,  we consider the use of Bayesian regression for building time series predictive models and  for stacking time series predictive models on the second level of the predictive model which is the ensemble of the models of the first level.

\subsection{Bayesian  Regression Approach}

Bayesian inference makes it possible to do nonlinear regression.
If we need to fit trend with saturation, we can use logistic curve model for nonlinear trend and find its parameters using Bayesian inference. 
Let us consider the case of nonlinear regression for time series which have the trend with saturation. 
For modeling we consider sales time series. The model can be described as:
\begin{equation}
\begin{split}
& log(Sales) \sim \mathcal{N}(\mu_{Sales}, \, \sigma^2) \\
& \mu_{Sales}=\frac{a}{1+exp(b t +c)} +\beta_{Promo}Promo+ \\
&  \beta_{Time}Time+ \sum_{j} \beta^{wd}_{j}WeekDay_{j}, \\
\end{split}
\label{bayes_ts_eq1}
\end{equation}
where $WeekDay_{j}$ - are binary variables, which is 1 in the case of sales on an appropriate day of week  and 0 otherwise.   
We  can also build hierarchical models using Bayesian inference. 
In this approach, one set of parameters is the same for all data samples, other parameters can be different for different groups of data samples. In the case of sales data, we can consider trends, promo impact and seasonality as the same for all stores. But the impact of a specific store on sales can be described by an intersect parameter, so this coefficient for this variable will be different for different stores. 
The hierarchical model can be described as:
\begin{equation}
\begin{split}
& Sales \sim \mathcal{N}(\mu_{Sales}, \, \sigma^2) \\
& \mu_{Sales}=\alpha(Store)+\beta_{Promo}Promo+ \\
&  \beta_{Time}Time+ \sum_{j} \beta^{wd}_{j}WeekDay_{j}, \\
\end{split}
\label{eq_hierarchical_model_1}
\end{equation}
where intersect parameters $\alpha(Store)$ are different for different stores. 

Predictive models can be combined into ensemble model using stacking approach
~\cite{wolpert1992stacked, rokach2010ensemble, sagi2018ensemble, gomes2017survey, dietterich2000ensemble, rokach2005ensemble}. 
In this approach, prediction results of predictive models on the validation set are 
treated  as covariates for stacking regression. These predictive models are considered as a first level of predictive model ensemble. Stacking model forms the second level of model ensemble.  Using Bayesian inference for stacking regression gives distributions for stacking regression coefficients. It enables us to estimate 
the uncertainty of the first level predictive models.
As predictive models for first level of ensemble, we used the following 
models: ARIMA', 'ExtraTree', 'RandomForest', 'Lasso', 'NeuralNetowrk'. 
The use of these models for stacking by Lasso regression was described in 
~\cite{pavlyshenko2019machine}. 
For stacking, we have chosen the  robust regression with Student's t-distribution for the target variable
as 
 \begin{equation}
y \sim  Student_t (\nu, \mu, \sigma) 
\label{bayesian_ts_eq_1}
\end{equation}
where 
\begin{equation}
 \mu = \alpha + \sum_{i}{\beta_{i}x_{i} }, 
 \label{bayesian_ts_eq_2}
\end{equation}
$\nu$ is a distribution parameter, called as degrees of freedom, 
$i$ is an index of the predictive model in the stacking regression, $i \in$ \{ 'ARIMA', 'ExtraTree', 'RandomForest', 'Lasso', 'NeuralNetowrk' \}.
\subsection{Numerical Modeling}
\begin{figure}
\centering
\includegraphics[width=0.55\linewidth]{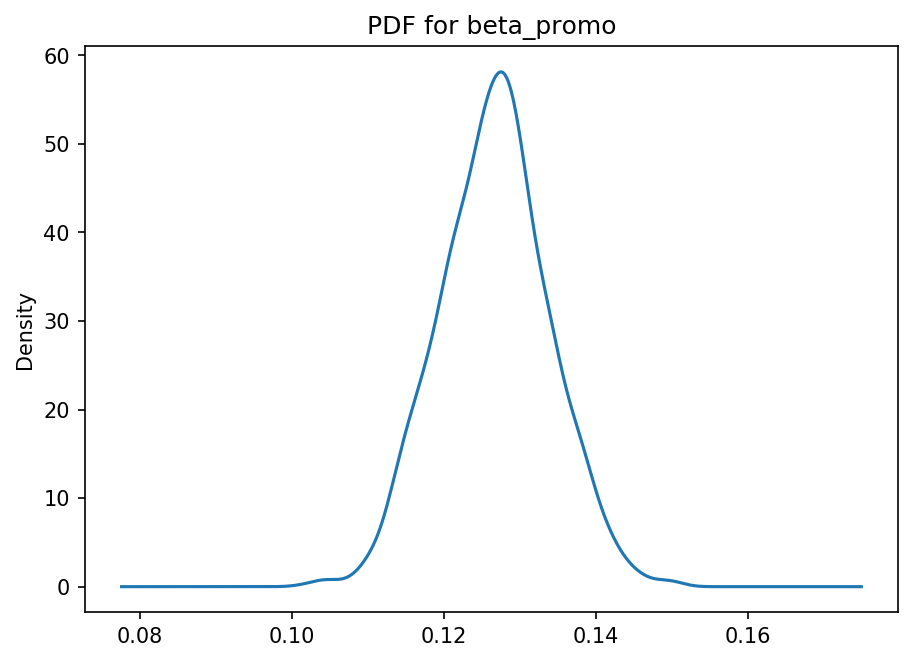}
\caption{Probability density function of regression coefficients for Promo factor}
\label{bayesian_ts_fig2}
\end{figure}

The data for our analysis are based on store sales historical data from  the ``Rossmann Store Sales'' Kaggle competition ~\cite{rossmanstorekaggle}.  For Bayesian regression, we used Stan platform for statistical modeling 
~\cite{carpenter2017stan}. The analysis was conducted in Jupyter Notebook environment  using Python programming language and 
the following  main Python packages \textit{pandas, sklearn, pystan,  numpy, scipy,statsmodels,  keras, matplotlib, seaborn}.  
For numerical analysis we modeled time series with multiplicative trend with saturation. 
Figure~\ref{bayesian_ts_fig1} shows results of this modeling where mean values and Value at Risk (VaR) characteristics are given. VaR was calculated as 5\% percentile. Figure~\ref{bayesian_ts_fig2} shows 
probability density function of regression coefficients for Promo factor. 
Figure~\ref{bayesian_ts_fig3} shows box plots for probability density function of seasonality coefficients. 
\begin{figure}
\centering
\includegraphics[width=0.85\linewidth]{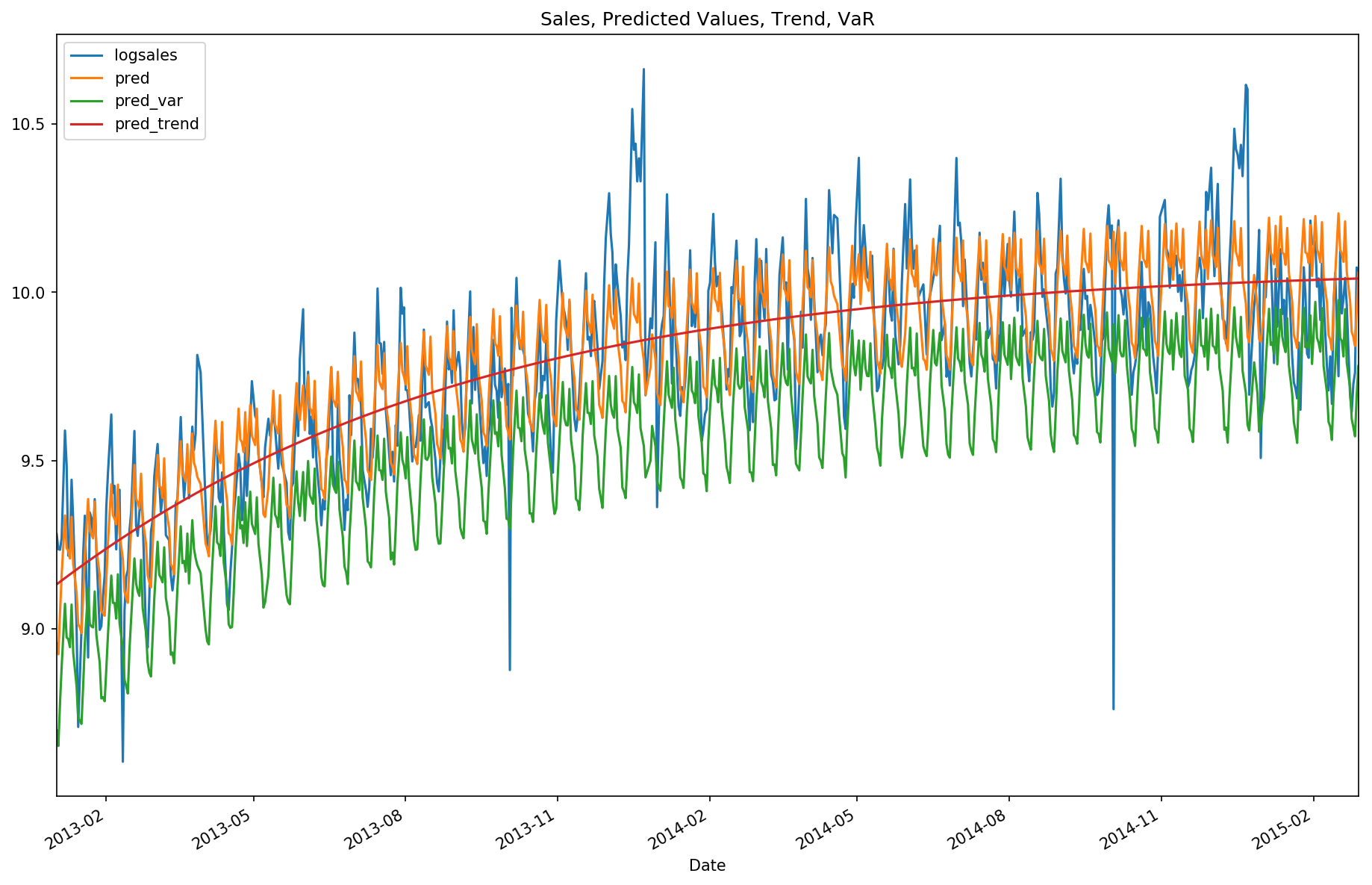}
\caption{Time series and trend forecasting}
\label{bayesian_ts_fig1}
\end{figure}

\begin{figure}
\centering
\includegraphics[width=0.75\linewidth]{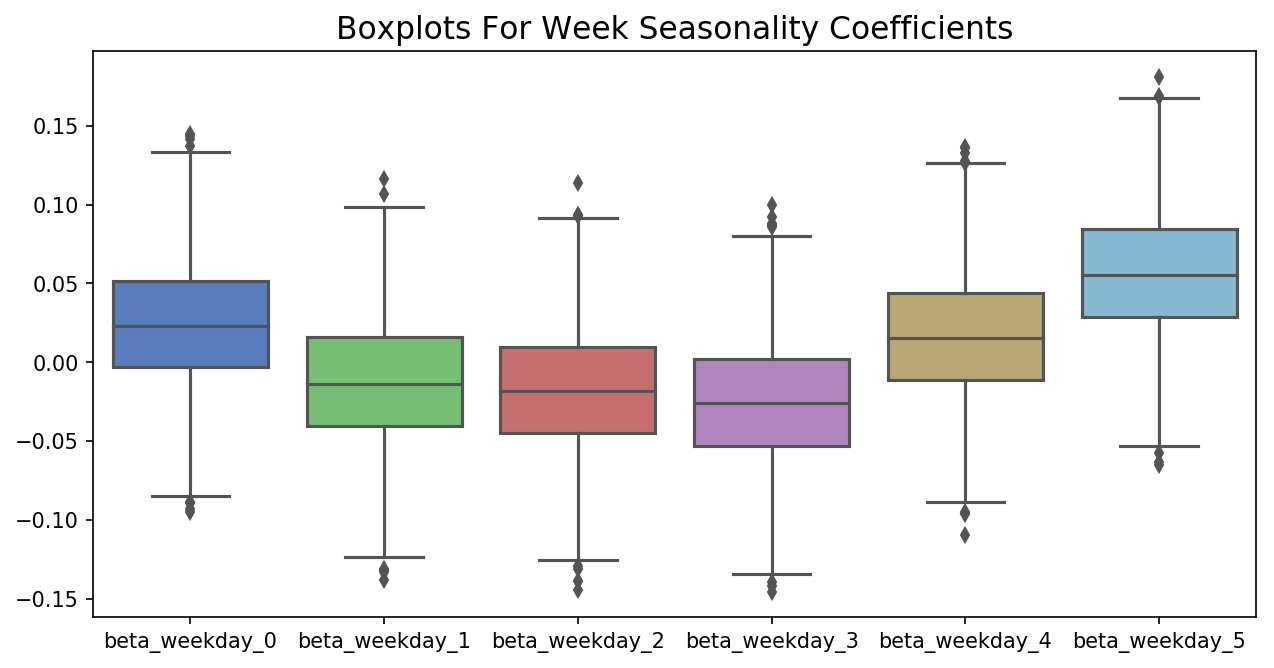}
\caption{Box plots for probability density function of seasonality coefficients}
\label{bayesian_ts_fig3}
\end{figure}
Let us consider Bayersian hierarchical model for time series. Intersect parameters $\alpha(Store)$ in the model (\ref{eq_hierarchical_model_1}) are different for different stores.
We have considered a case  with five different stores. 
Figure~\ref{bayesian_ts_fig4} shows the boxplots for the probability density functions  of intersect parameters. 
The dispersion of these dsitributions describes an uncertainty of influence of specified store on sales. 
 Hierarchical approach makes it possible to use such a model in the case with short historical data for specific stores, e.g. in the case of new stores. 
Figure~\ref{bayesian_ts_fig5}  shows the results of the forecasting on the validation set in the two cases, the fisrt case  is when we use 2 year historical data and the second case is with historical data for 5 days. We can see that such short historical data allow us to estimate sales dynamics correctly. Figure ~\ref{bayesian_ts_fig6}
shows box plots for probability density function of intersect parameters of different time series in case of short time period of historical data for a specified store \textit{Store\_4}. The obtained results show 
 that the dispersion for a specific store with short historical data becomes larger due to uncertainty caused by very short historical data for the specified store.
\begin{figure}
\centering
\includegraphics[width=0.65\linewidth]{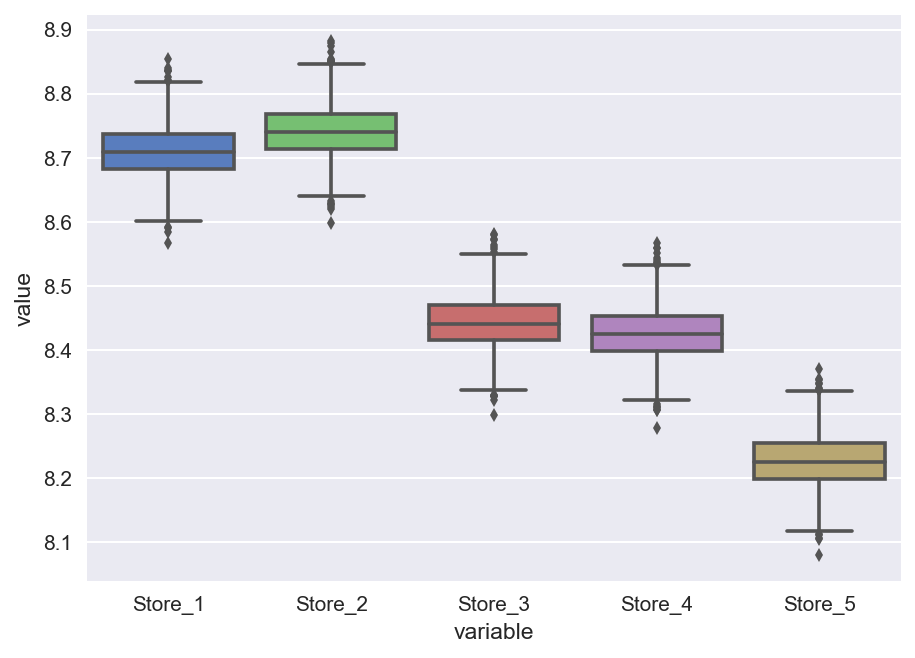}
\caption{Boxplots for  probability density functions  of intersect parameters}
\label{bayesian_ts_fig4}
\end{figure}
\begin{figure}
\centering{\includegraphics[width=0.8\linewidth]{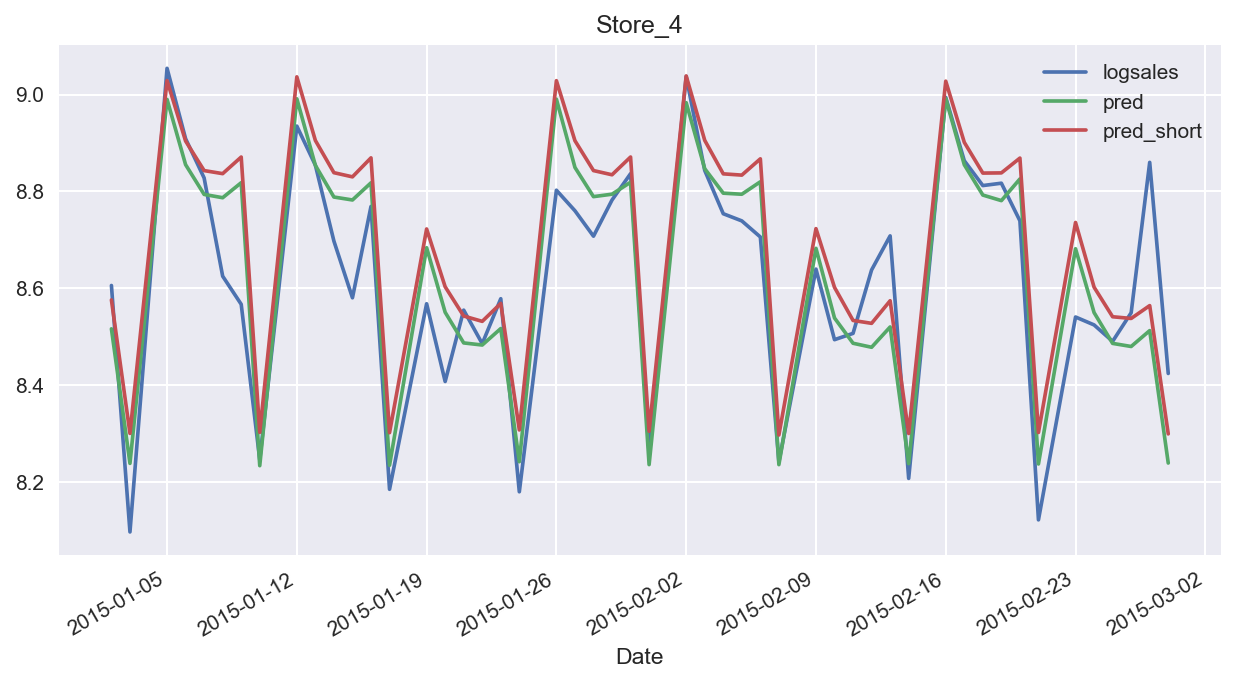}}
\caption{ Forecasting on the validation set for specified time series in cases with different size of historical data}
\label{bayesian_ts_fig5}
\end{figure}
\begin{figure}
\centering
\includegraphics[width=0.65\linewidth]{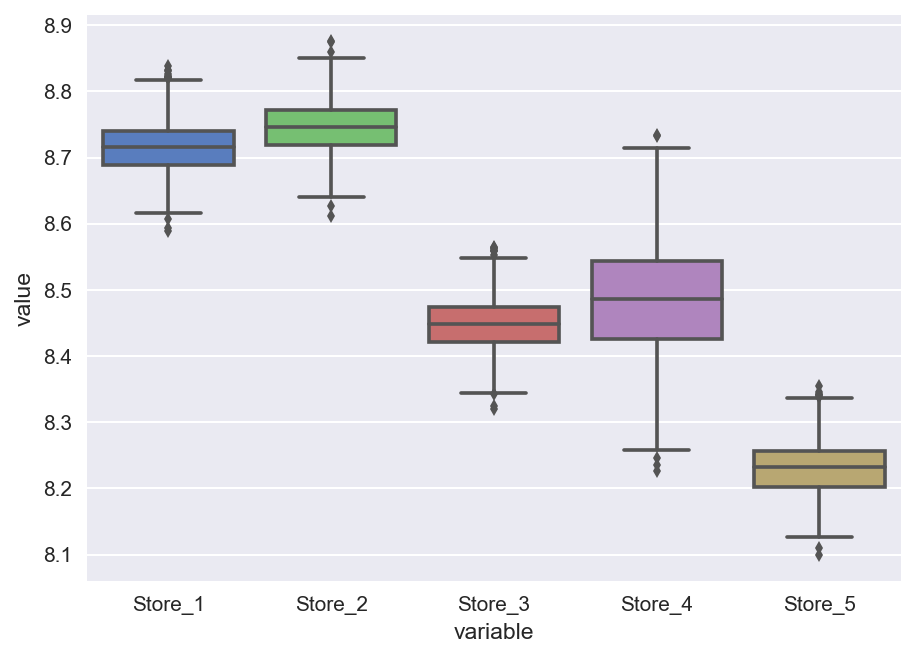}
\caption{Box plots for probability density function of intersect parameters of diferent time series in case of short time period of historical data for a specified store \textit{Store\_4}. }
\label{bayesian_ts_fig6}
\end{figure}

Let us consider the results of Bayesian regression approach for stacking predictive models.
 We trained different predictive models and made the predictions on the validation set. 
 The  ARIMA model was evaluated using \textit{statsmodels} package, Neural Network was evaluated using \textit{keras} package, Random Forest and  Extra Tree was evaluated using \textit{sklearn} package. In these calculations, we used the approaches described in 
 ~\cite{pavlyshenko2019machine}.
Figure ~\ref{bayesian_ts_fig1} shows the time series forecasts on the validation sets obtained using different models. 

\begin{figure}
\centering
\includegraphics[width=0.85\linewidth]{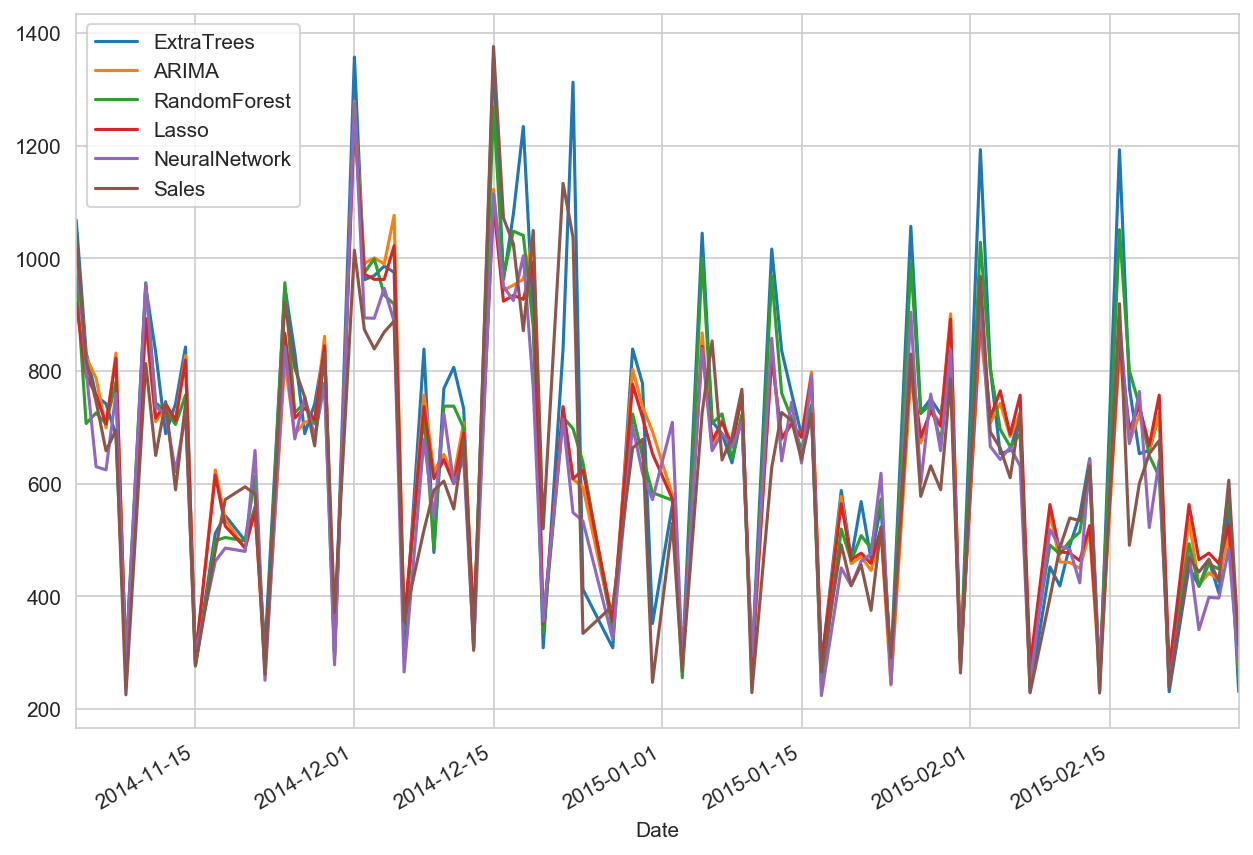}
\caption{Forecasting of different models on the validation set}
\label{bayesian_ts_fig1}
\end{figure}

The results of prediction of these models on the validation sets are considered as
the  covariates for the regression on the second stacking level of the  ensemble of models.
For stacking predictive models we split the validation set on the training and testing sets.
For stacking regression, we normalized the covariates and target variable using 
z-scores:
\begin{equation}
z_{i}= \frac{x_{i}-\mu_{i}}{\sigma_{i}},
\end{equation}
where $\mu_{i}$ is the  mean value, $\sigma_{i}$ is the standard deviation. 
The prior distributions  for parameters $\alpha, \beta, \sigma $ in the  Bayesian  regression model 
(\ref{bayesian_ts_eq_1})-(\ref{bayesian_ts_eq_2})  are considered as Gaussian with mean values equal to 0, and standard deviation equal to 1. 
We split the validation set on the training and testing sets by time factor.
The parameters for prior distributions can be adjusted 
using prediction scores on testing sets or using expert opinions in the case of small 
data amount. 
To estimate uncertainty of regression coefficients, 
we used  the coefficient of variation
which is defined as a ratio between the 
standard deviation and mean value for model coefficient distributions:
\begin{equation}
v_{i}= \frac{\sigma_{i}}{\mu_{i}},
\end{equation}
where $v_{i}$ is the coefficient of variation, $\sigma_{i}$ is a standard deviation, $\mu_{i}$ is the mean value for the distribution of the regression coefficient of the model $i$.
 Taking into account that $\mu_{i}$ can be negative, we will analyze the absolute value of the coefficient of variation $\vert v_{i} \vert$.
 For the results evaluations, we used a relative mean absolute error (RMAE) and root mean square error (RMSE).  Relative mean absolute error (RMAE) was considered as a ratio between the mean absolute error (MAE) and mean values of target variable:
  \begin{equation}
RMAE=\frac{E ( \vert y_{pred} - y \vert  )} {E( y )}  100\%
\end{equation}
 Root mean square error (RMSE) was considered as:
  \begin{equation}
RMSE=\sqrt{{\frac{\sum_{i}^{n}(y_{pred}-y)^2}{n}}}
\end{equation}

 The data with predictions of different models  on  the validation set were split on the training set (48 samples) and testing set (50 samples) by date. 
We used the robust regression with Student's t-distribution for the target variable. 
As a result of calculations, we received the following scores: RMAE(train)=12.4\%, RMAE(test)=9.8\%, RMSE(train)=113.7, RMSE(test)= 74.7.
Figure ~\ref{bayesian_ts_fig2}  shows mean values time series for real and forecasted sales on the validation and testing sets.
The vertical dotted line  separates the training and testing sets.
Figure ~\ref{bayesian_ts_fig3}  shows the  probability density function (PDF) for the intersect parameter. One can observe 
a positive bias of this (PDF). It is caused by the fact that we applied machine learning  algorithms to  nonstationary time series. If a nonstationary trend is small, it can be compensated on  the validation set using stacking regression.  

\begin{figure}
\center
\includegraphics[width=0.75\linewidth]{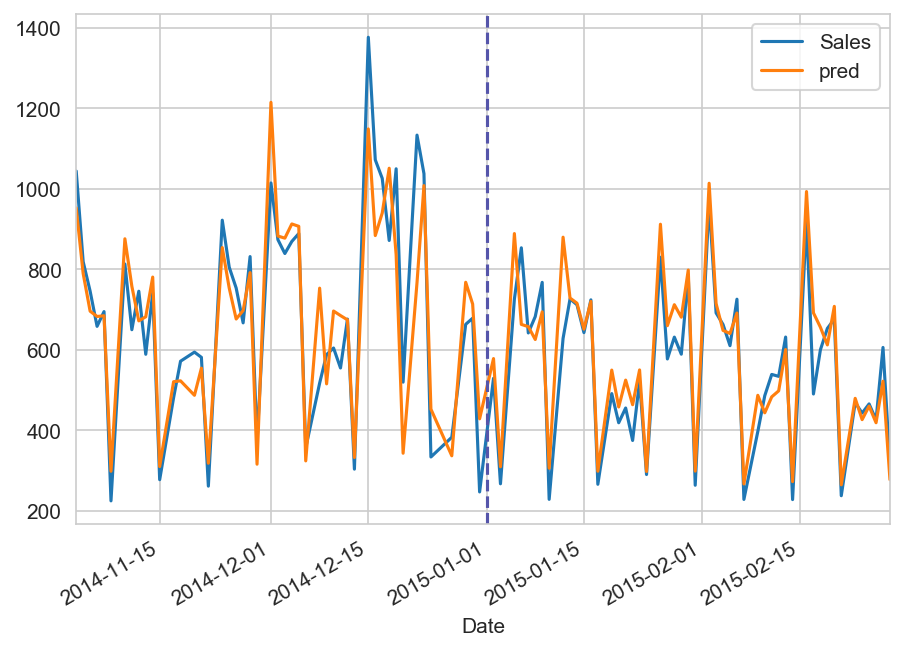}
\caption{Time series of mean values for real and forecasted sales on validation and testing sets}
\label{fbayesian_ts_ig2}
\end{figure}
\begin{figure}
\centering
\includegraphics[width=0.65\linewidth]{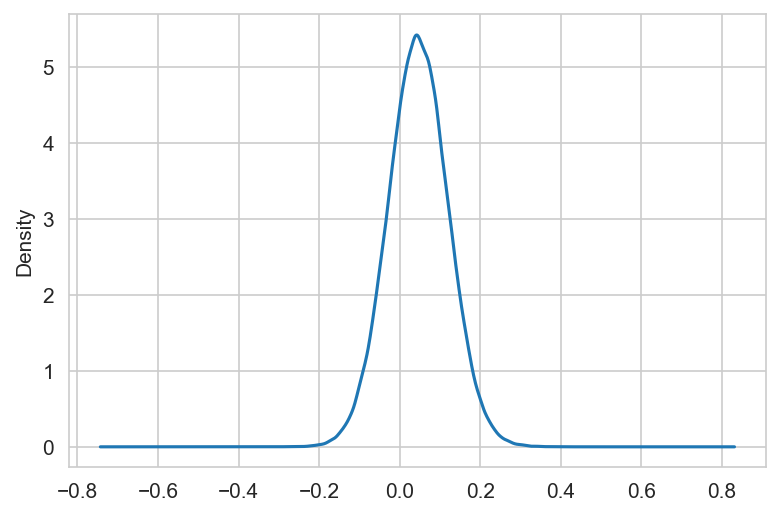}
\caption{The PDF for intersect parameter of stacking regression}
\label{bayesian_ts_fig3}
\end{figure}
Figure ~\ref{bayesian_ts_fig4}  shows the box plots for the PDF of  coefficients of models.
Figure ~\ref{bayesian_ts_fig5}  shows the coefficient of variation for the PDF of  regression coefficients of models. 
\begin{figure}
\centering
\includegraphics[width=0.75\linewidth]{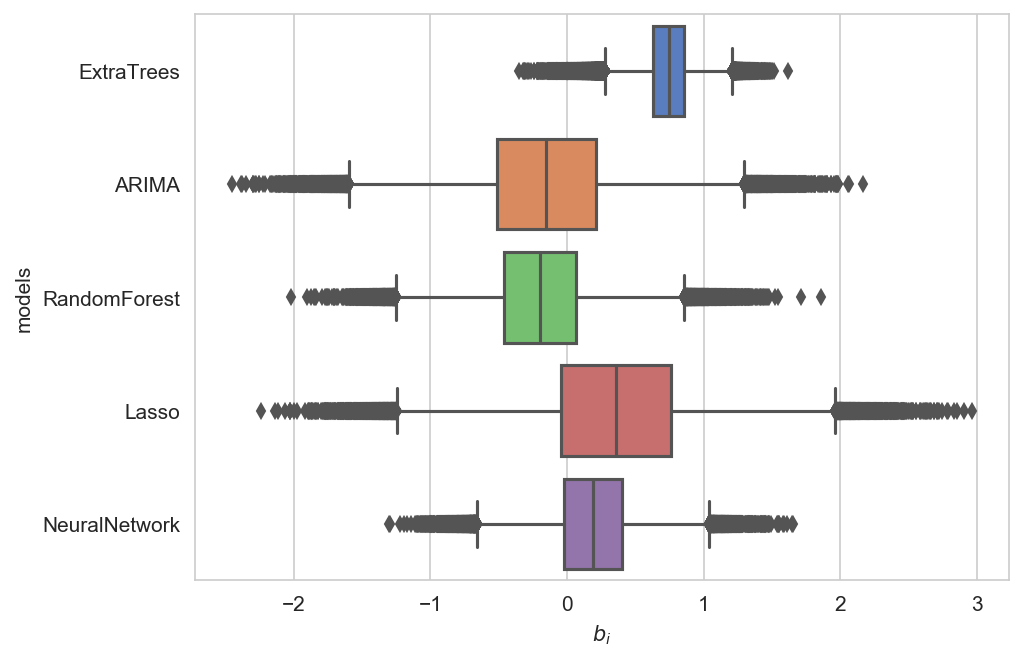}
\caption{Box plots for the PDF of  regression coefficients of models}
\label{bayesian_ts_fig4}
\end{figure}
\begin{figure}
\centering
\includegraphics[width=0.75\linewidth]{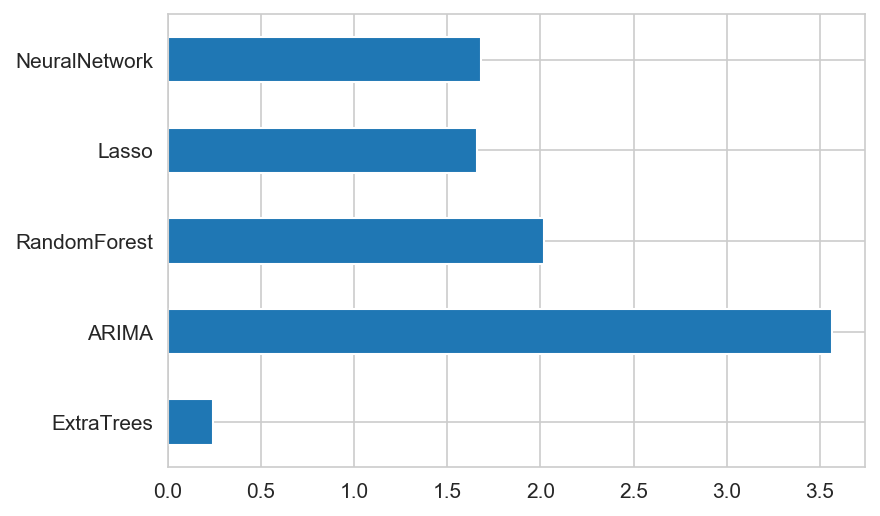}
\caption{Absolute values of the coefficient of variation for the PDF of  regression coefficients of models}
\label{bayesian_ts_fig5}
\end{figure}

We considered the case with the restraints to  regression coefficient of models that they should be positive.
We received the similar results:
RMAE(train)=12.9\%, RMAE(test)=9.7\%, RMSE(train)=117.3, RMSE(test)=76.1.
Figure ~\ref{bayesian_ts_fig6}  shows the box plots for the PDF of model regression  coefficients for this case.

\begin{figure}
\centering
\includegraphics[width=0.75\linewidth]{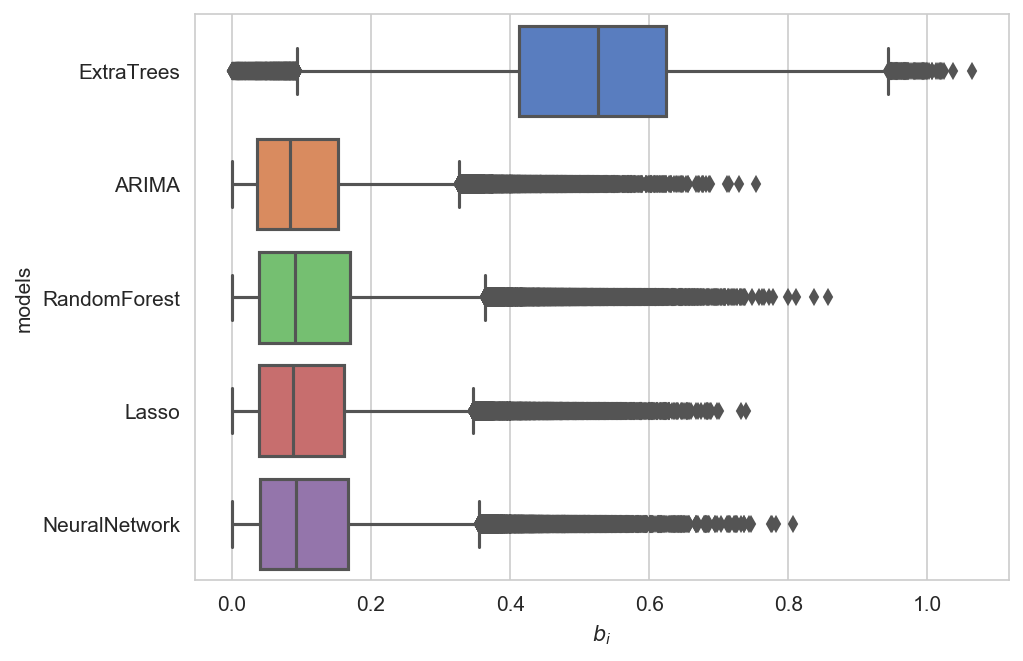}
\caption{Box plots for the PDF of  regression coefficients of models}
\label{bayesian_ts_fig6}
\end{figure}

\begin{figure}
\centering
\includegraphics[width=0.75\linewidth]{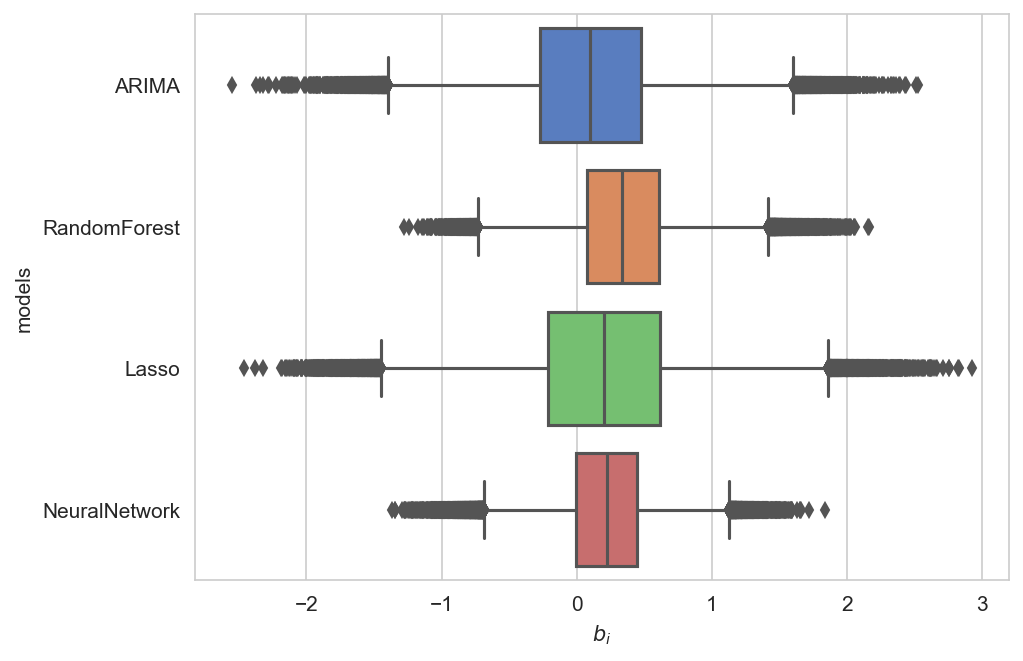}
\caption{Box plots for the PDF of  regression coefficients of models}
\label{bayesian_ts_fig7}
\end{figure}

\begin{figure}
\centering
\includegraphics[width=0.75\linewidth]{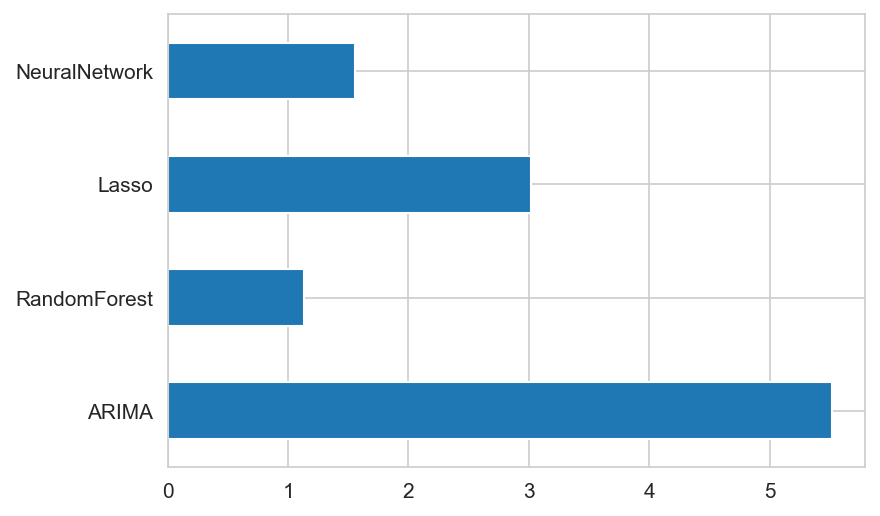}
\caption{Absolute values of the coefficient of variation for the PDF of models regression coefficients}
\label{bayesian_ts_fig8}
\end{figure}

\begin{figure}
\centering
\includegraphics[width=0.75\linewidth]{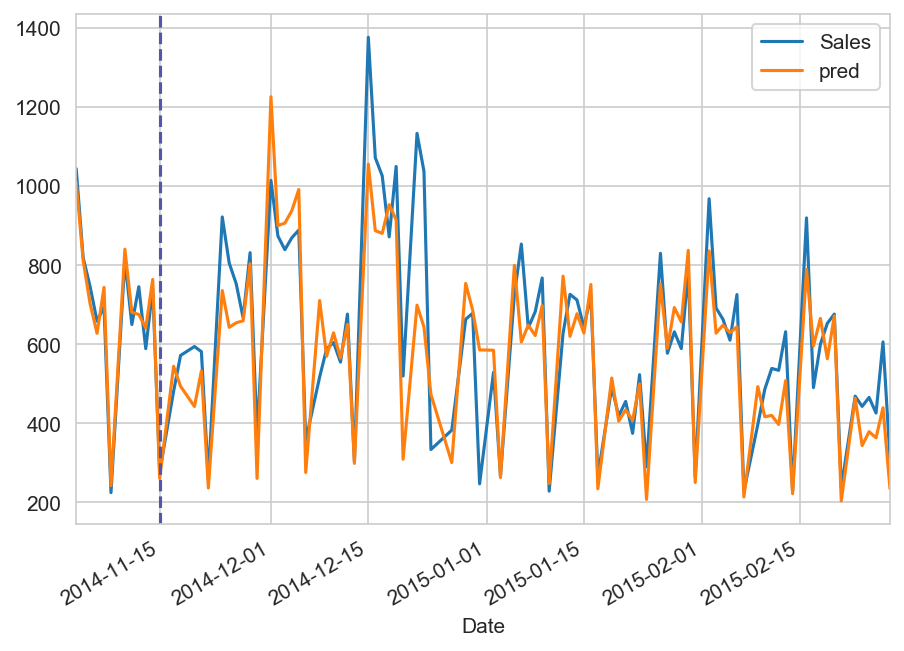}
\caption{Mean values time series for real and forecasted sales on the validation and testing sets in the case of small training set}
\label{bayesian_ts_fig9}
\end{figure}

\begin{figure}
\centering
\includegraphics[width=0.75\linewidth]{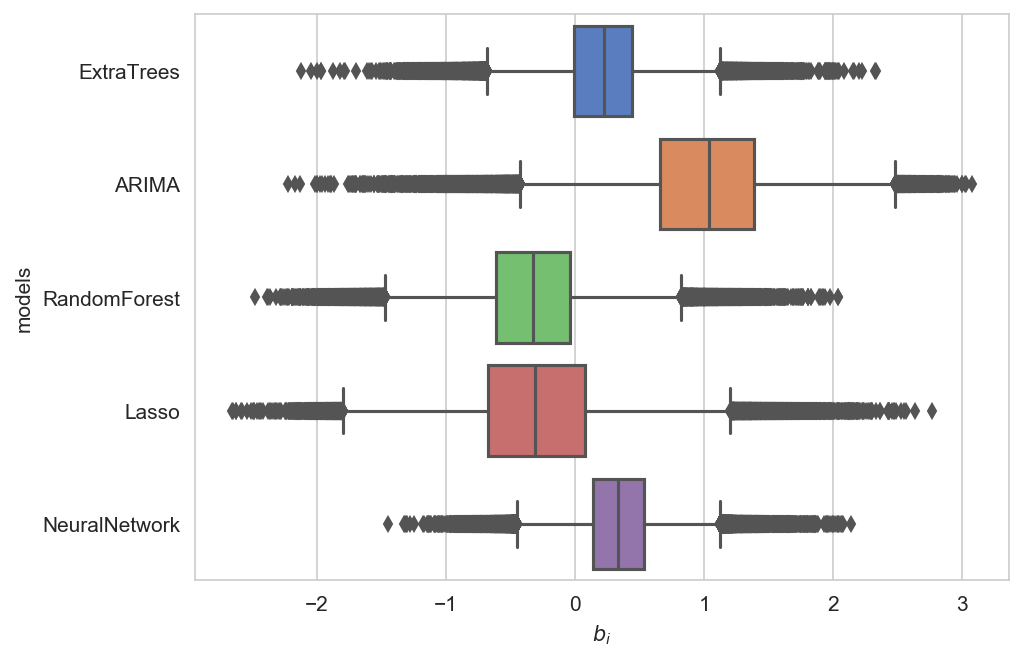}
\caption{Box plots for the PDF of regression coefficients of models in the case of small training set}
\label{bayesian_ts_fig10}
\end{figure}

\begin{figure}
\centering
\includegraphics[width=0.75\linewidth]{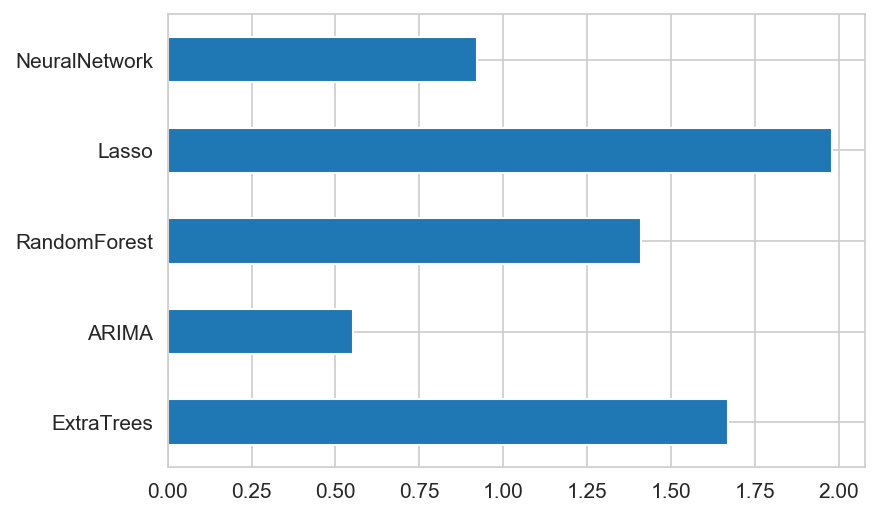}
\caption{Absolute values of the coefficient of variation for the PDF of  regression coefficients of models in the case of small training set}
\label{bayesian_ts_fig11}
\end{figure}

  All models have a similar mean value and variation coefficients. 
We can observe that errors characteristics RMAE and RMSE  on the validation set can be similar with respect to these errors on the training set. It tells us about the fact that Bayesian regression does not overfit on the training set comparing to the machine learning algorithms which can  demonstrate essential overfitting on  training sets, especially in the cases 
 of small amount of training data. 
We have chosen the best ExtraTree stacking model    and conducted Bayesian regression  with this one model only. We received the following scores:
RMAE(train)=12.9\%, RMAE(test)=11.1\%, RMSE(train)=117.1, RMSE(test)=84.7.
We also tried  to exclude the best model ExtraTree from the stacking regression and 
conducted Bayesian regression  with the rest  of models without ExtraTree. In this case
we received  the following scores:
RMAE(train)=14.1\%, RMAE(test)=10.2\%, RMSE(train)=139.1, RMSE(test)=75.3.
Figure~\ref{bayesian_ts_fig7}  shows the box plots for the PDF of model regression coefficients,
 figure~\ref{bayesian_ts_fig8}  shows the coefficient of variation for the PDF of regression coefficients of models for this case study.   We received worse results on the testing set. At the same time these models have the similar influence  and thus they  can potentially provide more stable results in the future  due to possible  changing of the quality of features. Noisy models can decrease accuracy on large training data sets, at the same time they contribute to sufficient results in the case of small data sets. 
We considered the case with a small number of training data, 12 samples. To get stable results, we fixed the $\nu$ parameter of Student's t-distribution in Bayesian regression model (\ref{bayesian_ts_eq_1})-(\ref{bayesian_ts_eq_2}) equal to 10.  
 We received  the following scores: RMAE(train)=5.0\%, RMAE(test)=14.2\%, RMSE(train)=37.5, 
 RMSE(test)=121.3.
Figure ~\ref{bayesian_ts_fig9}  shows mean values time series for real and forecasted sales on the validation and testing sets.  
Figure ~\ref{bayesian_ts_fig10}  shows the box plots for the PDF of regression coefficients of models.
Figure ~\ref{bayesian_ts_fig11}  shows the coefficient of variation for the PDF of  regression coefficients of models. In this case, we can see  that an other model starts playing an important role comparing with the previous cases and ExtraTree model does not dominate. 
The obtained results show that optimizing  informative prior distributions 
of stacking model parameters can improve the scores of prediction results on the testing set.

\FloatBarrier
\subsection{Conclusions}
Bayesian regression approach for time series modeling  with nonlinear trend was analyzed. 
Such  approach allows us to estimate an uncertainty of time series prediction and  calculate value at risk (VaR) characteristics. 
Hierarchical model for time series using Bayesian regression has been considered. 
In this approach, one set of parameters is the same for all data samples, other parameters can be different for different groups of data samples. This approach 
 makes it possible to use such a model in the case with short historical data for specified time series, e.g. in the case of new stores or new products in the sales prediction problem.
 The use of Bayesian inference for time series predictive models stacking  has been analyzed.
A two-level ensemble of the predictive models for time series was considered.  
The models  ARIMA, Neural Network, Random Forest, Extra Tree were used for the prediction on the first level of ensemble of models. 
On the second stacking level, time series predictions of these models on the validation set were conducted by Bayesian regression. 
Such an approach gives distributions for regression coefficients of these models. It makes it possible to estimate the uncertainty  contributed by  each model to the stacking result.  
The information about these distributions allows us to select an optimal set of stacking models, taking into account domain knowledge.  Probabilistic approach for stacking predictive 
models allows us to make risk assessment for the predictions that is important in a decision-making process. 
Noisy models can decrease accuracy on large training data sets, at the same time they contribute to sufficient results in the case of small data sets. 
Using Bayesian inference for stacking  regression can be useful in cases of small datasets and help experts to select a set of models for stacking as well as make assessments of different kinds of risks. 
Choosing the final models for stacking is up to an expert  who takes into account different factors such as  uncertainty of each model on the stacking regression level,
   amount of training and testing data, the stability of models.
   In Bayesian regression,  we can receive a quantitative measure for the uncertainty that can be a very  useful information for experts in model selection and stacking. 
  An  expert can also set up informative prior distributions for  stacking regression  coefficients of models,
   taking into account the domain knowledge information. 
 So, Bayesian approach for stacking regression can give us the information about 
uncertainty of predictive models. Using this information and domain knowledge, 
an expert can select models to get stable stacking ensemble of predictive models.   

\section{Forecasting of  Non-Stationary Sales Time Series Using Deep Learning}
In this case study,   we consider   the deep learning approach for forecasting non-stationary time series with using time trend correction  in a neural network model.  
Along with the layers for predicting sales values, the neural network model includes a subnetwork block for the prediction weight for a time trend term which is added to a predicted sales value. The time trend term is considered as a product of the predicted weight value and normalized time value.  The results show that 
 the forecasting accuracy can be essentially improved for non-stationary sales with time trends using the trend correction block in the deep learning model.  

\subsection{Introduction}
Sales and demand forecasting are widely being used in business analytics~\cite{mentzer2004sales,efendigil2009decision,zhang2004neural}. 
Sales can be treated as time series.
Different time series approaches are described in~\cite{chatfield2000time,brockwell2002introduction,box2015time,doganis2006time,hyndman2018forecasting,
tsay2005analysis,wei2006time,cerqueira2018arbitrage,hyndman2007automatic,papacharalampous2017comparison,tyralis2017variable,
tyralis2018large, papacharalampous2018predictability, taieb2012review, graefe2014combining}. 
Machine learning is widely used  for forecasting different kinds of time series along with classical statistical methods like ARIMA,  Holt-Winters,  etc.  
 Moderm deep learning algorithms DeepAR~\cite{salinas2020deepar, alexandrov2020gluonts}, \mbox{N-BEATS}~\cite{oreshkin2019n}, Temporal Fusion Transformers~\cite{lim2019temporal} show state-of-the-art results for time series forecasting.  
 Sales prediction is more a regression problem than a time series problem. 
The use of regression approaches for sales forecasting can often give us  better results compared to time series methods. 
Machine-learning algorithms make it possible to find patterns in the time series.
Some of the most popular ones are  tree-based machine-learning algorithms~\cite{james2013introduction}, e.g., Random Forest~\cite{breiman2001random}, Gradiend Boosting Machine~\cite{friedman2001greedy, friedman2002stochastic}.  
 The important time series features for their successful forecasting  are 
  their stationarity and sufficiently long time of historical observation to be able to capture intrinsic time series patterns. 
 One of the main assumptions of regression methods is that the patterns in the past data will be repeated in future.  
 There are some limitations of time series approaches for sales forecasting.  Let us consider some of them. 
We need to have historical data for a long time period to capture seasonality. 
However, often we do not have historical data for  a target variable, for example in case when a new product is launched. 
At the same time,  we have sales time series for a similar product and we can expect that our new product will have  a similar  sales pattern. 
 Sales data can have  a lot of outliers and missing data. We must clean those outliers and interpolate data before using a time series approach.
 We need to take into account a lot of exogenous factors which impact on sales.   
 On the other hand,  sales time series  have their own specifics,  e.g. their dynamics is caused by rather exogenous factors than intrinsic patterns, they are often highly non-stationary,  we frequently face with short time sales observations, e.g.  in the cases when new products or stores are just launched. 
 Often non-stationarity is caused by a time trend.  Sales trends can be different for different stores, e.g. some stores can have an ascending trend,  while others have a descending one. 
  Applying machine learning regression to non-stationary data, bias in sales prediction  on validation dataset can appear. This bias can be corrected by additional linear regression on a validation dataset when a covariate stands for predicted sales and a target variable stands for  real sales~\cite{pavlyshenko2019machine}. 

 In~\cite{pavlyshenko2016linear}, we studied linear models, machine learning, and probabilistic models for time series modeling. 
For probabilistic modeling, we considered the use of copulas and Bayesian inference approaches. 
In~\cite{pavlyshenko2018using}, we studied stacking approaches for time series forecasting and logistic regression with highly imbalanced data. 
In~\cite{pavlyshenko2019machine}, we study the usage of machine-learning models for sales time series forecasting.  
In~\cite{pavlyshenko2020salests}, we analyse sales time series using Q-learning from the perspective of the  dynamic price and supply optimization.
In~\cite{pavlyshenko2020bayesian},  we study the Bayesian approach for stacking machine learning predictive models for sales time series forecasting. 

 In this work, we consider the deep learning approach for forecasting non-stationary time series with the trend using a trend correction block in the deep learning model.   
Along with the layers for predicting sales values, the model includes a  subnetwork block for the prediction weight for a trend term which is added to the predicted sales value.  The trend term is considered as a product of the predicted weight value and normalized time value. 
  
\subsection{Sales time series with time trend}

\begin{figure}[H]
\center
 \includegraphics[width=1\linewidth]{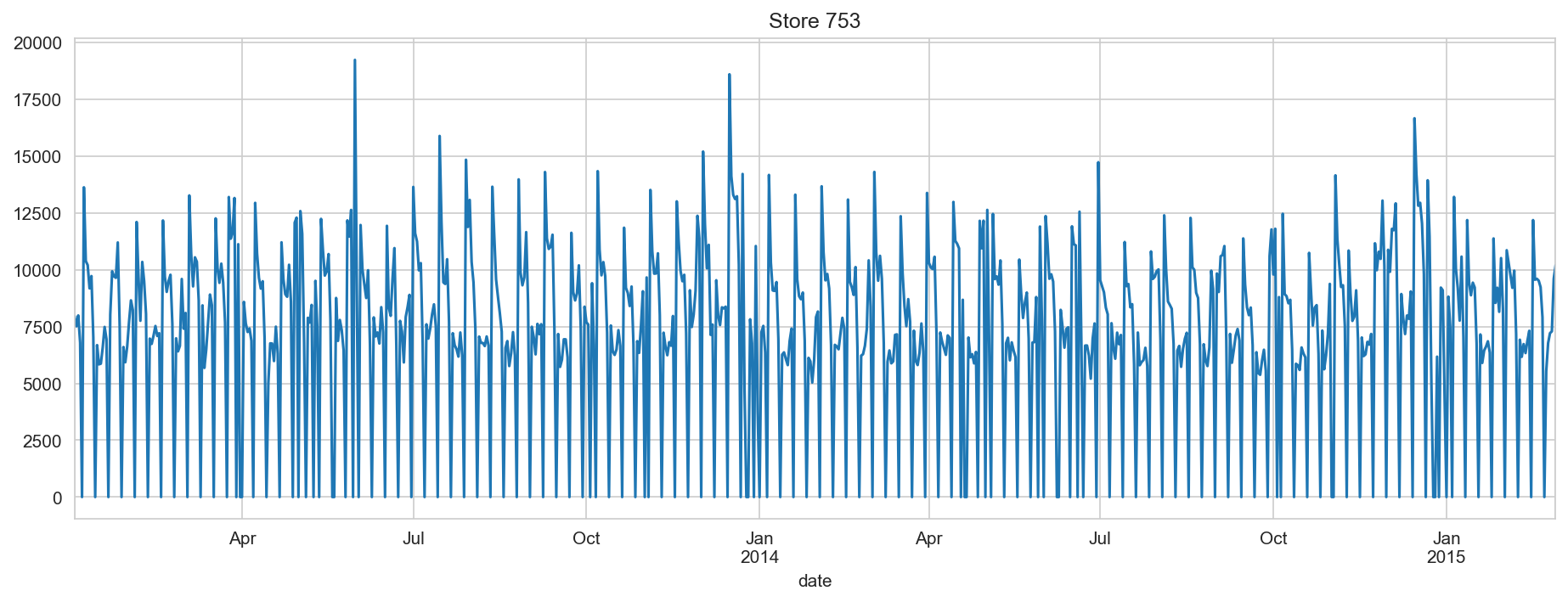}
 \caption{Store sales time series}
 \label{img1}
 \end{figure}

\begin{figure}[H]
\center
 \includegraphics[width=1\linewidth]{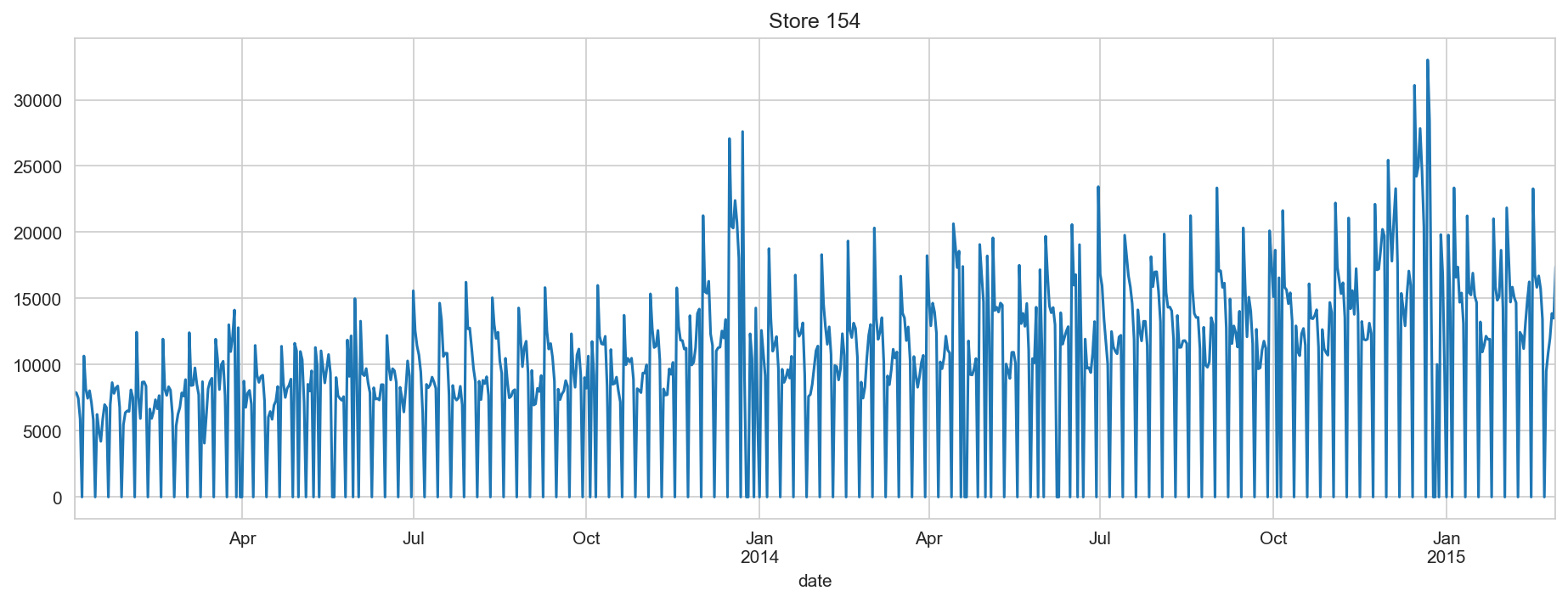}
 \caption{Store sales time series}
 \label{img2}
 \end{figure}

\begin{figure}[H]
\center
 \includegraphics[width=1\linewidth]{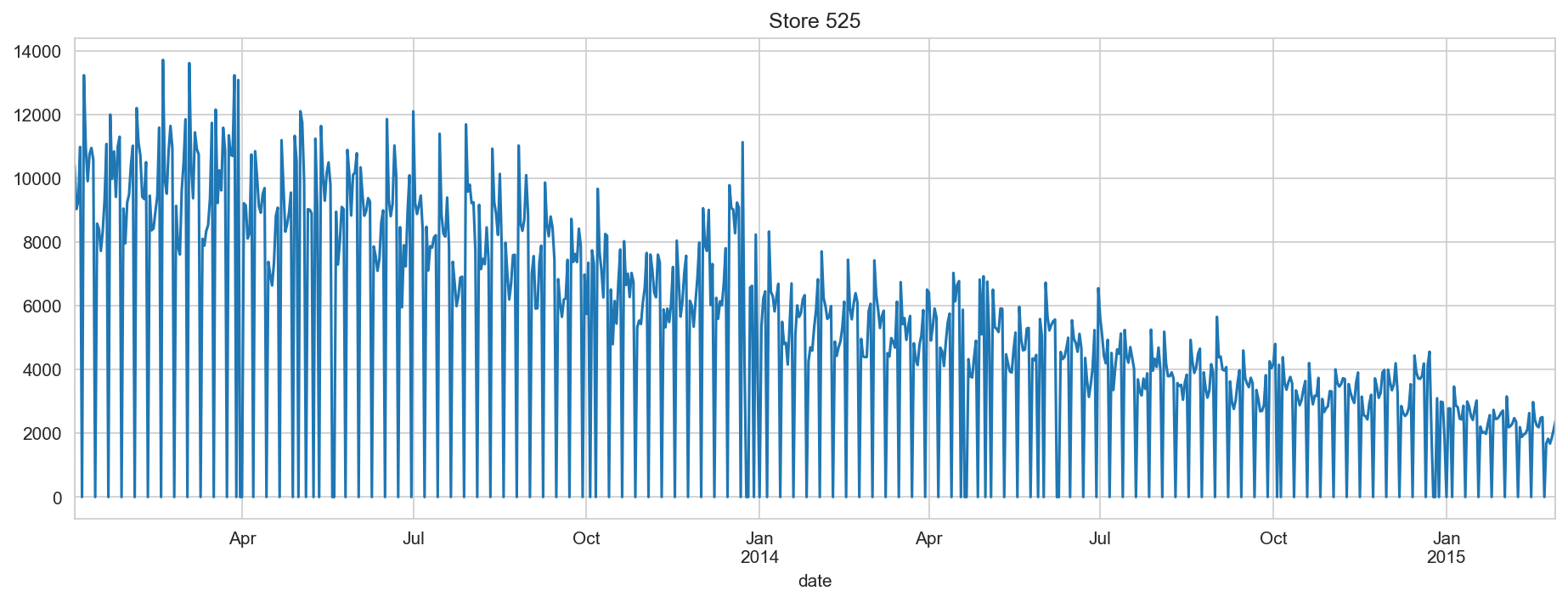}
 \caption{Store sales time series}
 \label{img3}
 \end{figure}

\begin{figure}[H]
\center
 \includegraphics[width=0.55\linewidth]{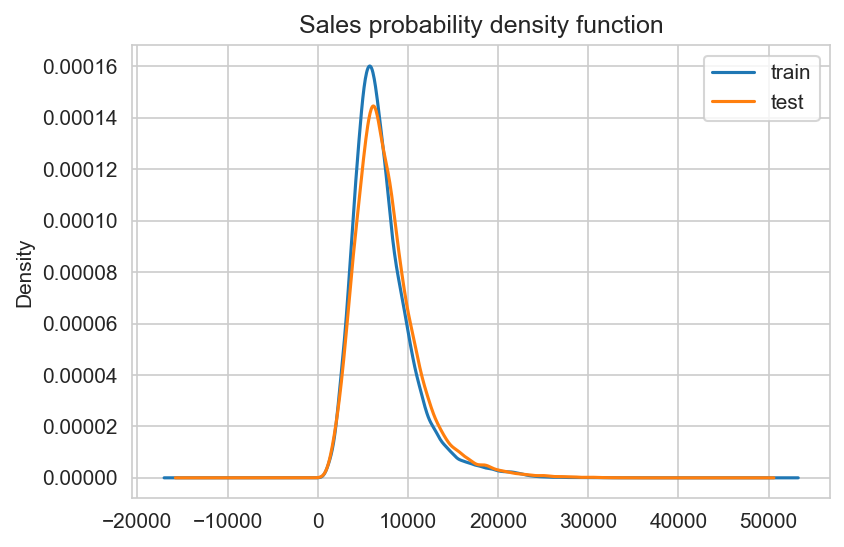}
 \caption{Probability density function for sales in all  stores}
 \label{img7}
 \end{figure}

\begin{figure}[H]
\center
 \includegraphics[width=0.55\linewidth]{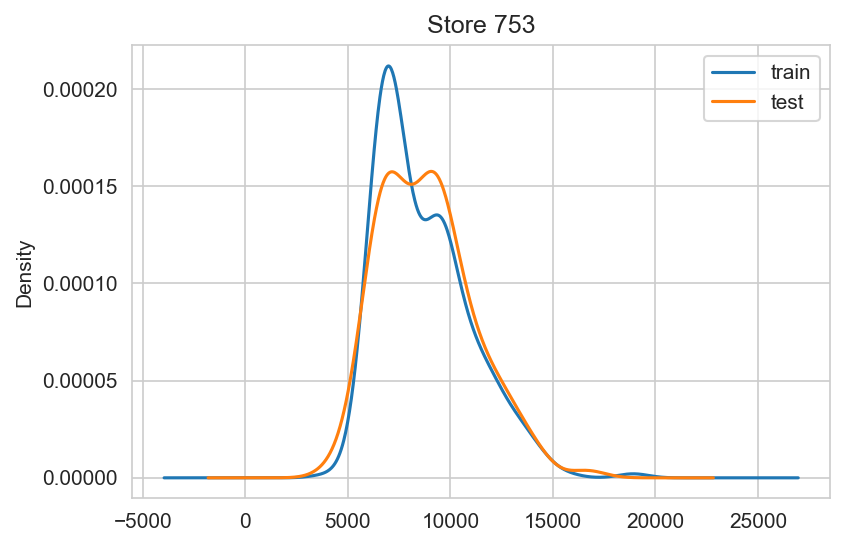}
 \caption{Probability density function for sales in specified  store}
 \label{img8}
 \end{figure}

\begin{figure}[H]
\center
 \includegraphics[width=0.55\linewidth]{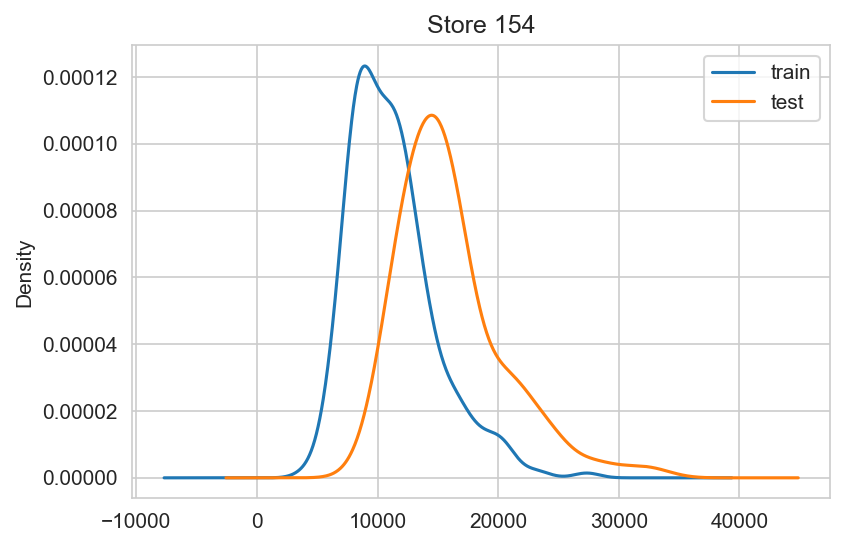}
 \caption{Probability density function for sales in specified  store}
 \label{img9}
 \end{figure}

\begin{figure}[H]
\center
 \includegraphics[width=0.55\linewidth]{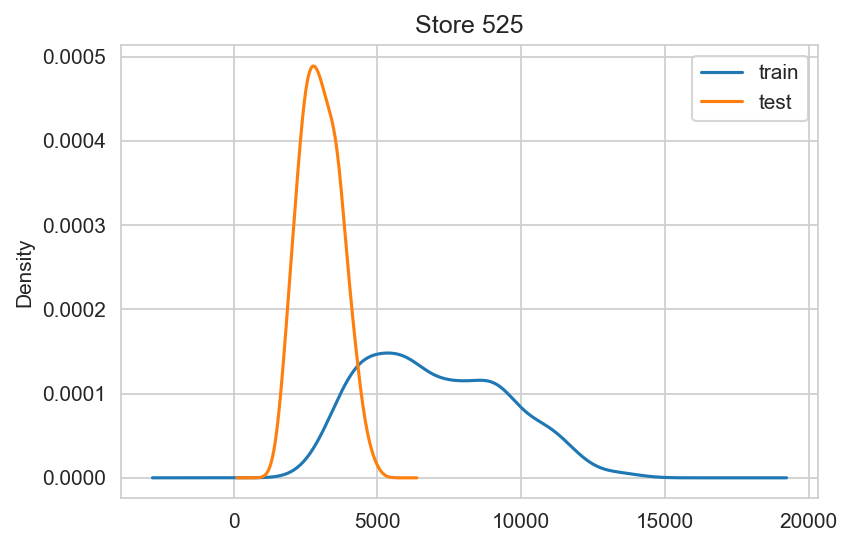}
 \caption{Probability density function for sales in specified  store}
 \label{img10}
 \end{figure}

  For the study, we have used the sales data which are based on the dataset from the 'Rossman Store Sales' Kaggle competition~\cite{rossmanstorekaggle}. 
  These data represent daily sales aggregated on the granularity level of customers and stores.  As the main features, \textit{ 'month','weekday',  'trendtype', 'Store', 'Customers',  'StoreType', 'StateHoliday','SchoolHoliday','CompetitionDistance'} were considered. 
  The \textit{'trendtype'} feature can be used in the case if the sales trend has different behavior types  on different time periods.  
  To study non-stationarity, arbitrary time trends were artificially added to the data grouped by stores.  
  The calculations were conducted in the Python environment using the main packages \textit{pandas, sklearn, numpy, keras, matplotlib, seaborn}.  To~conduct the analysis, \textit{Jupyter Notebook} was used. 
  Figures~\ref{img1}-\ref{img3} show the arbitrary examples of 
  aggregated sales time series with different  time trends for different stores. 
  Both training and validation datasets were received by splitting the dataset by date, so the validation time period is next with respect to the training period.  
Figure~\ref{img7} shows the probability density function (PDF) for all stores for training and validation datasets.  
  Figures~\ref{img8}--\ref{img10} show the probability density function for sales in data  samples, of specified stores which correspond to the stores time series shown in 
   Figures~\ref{img1}--\ref{img3}.
For some stores,  sales  PDF are different for training and validation datasets due to non-stationarity caused by different sales trends for different stores. 

\subsection{Deep learning model with time trend correction}
\begin{figure}[H]
\center
 \includegraphics[width=0.7\linewidth]{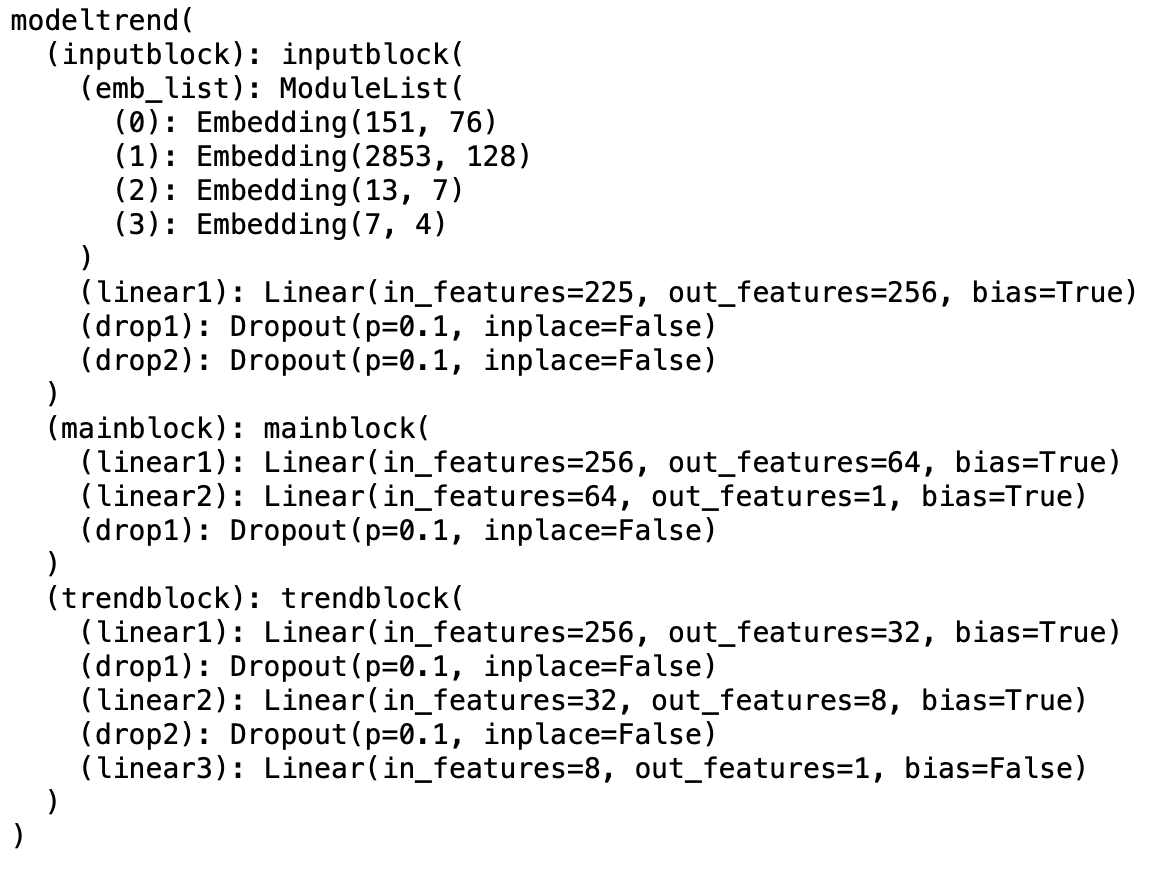}
 \caption{ Parameters of neural network layers}
 \label{model1}
 \end{figure}

\begin{figure}[H]
\center
 \includegraphics[width=0.8\linewidth]{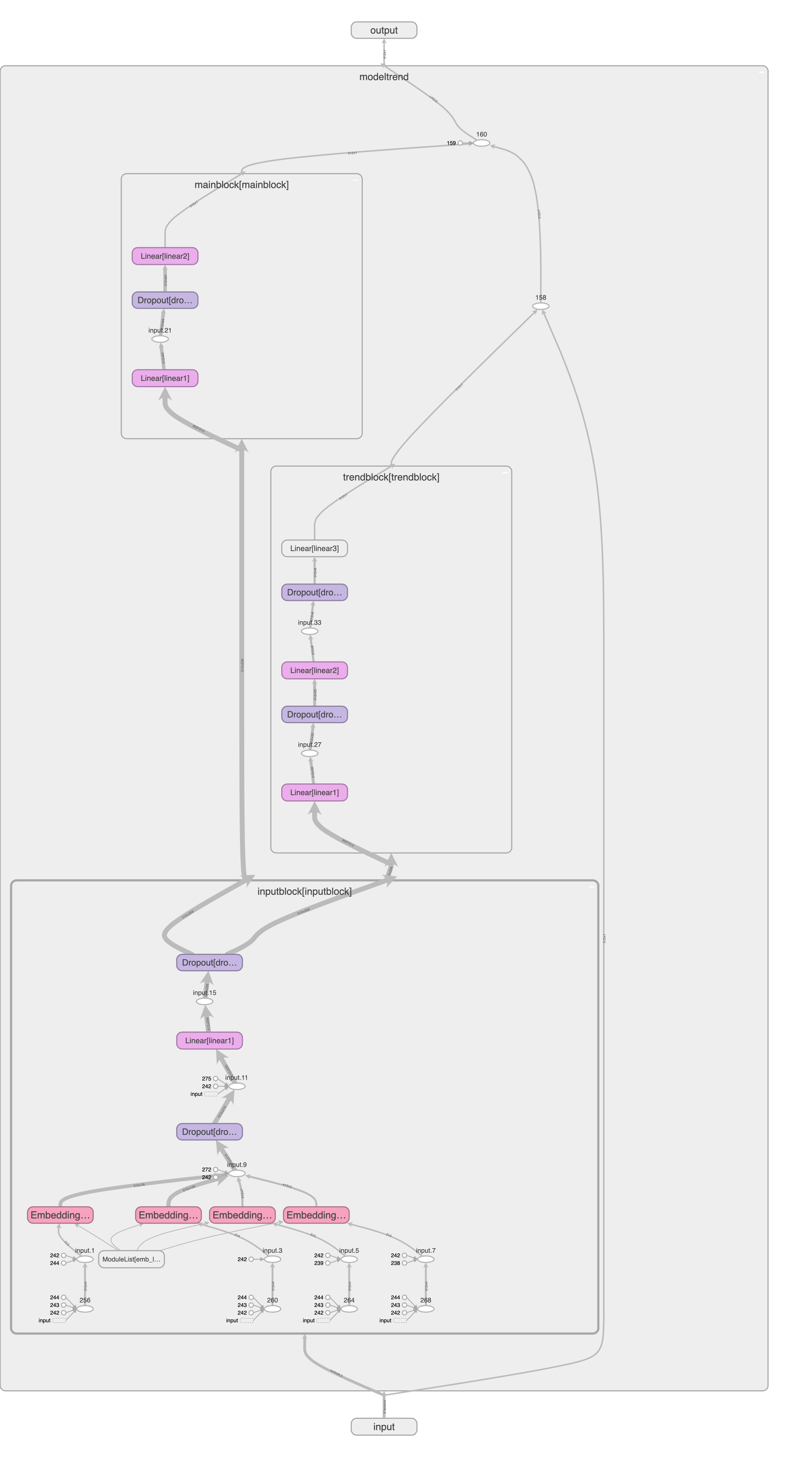}
 \caption{Neural network structure}
 \label{model2}
 \end{figure}

Let us consider  including a correction trend bock into the deep learning model. 
Along with the layers for predicting sales values, the model  will include a subnetwork block for the prediction weight for the time trend term which is added to the predicted sales value. The time trend term is considered as a product of the predicted weight value and normalized time value. 
The predicted sales values and the time trend term are combined in the loss function. As a result,  one can receive  an optimized trend correction for non-stationary sales for different groups of data with different trends.  
For modeling and deep learning case study,  the Pytorch deep learning library~\cite{paszke2017automatic, paszke2019pytorch}  was used. 
Categorical variables \textit{Store, Customer} with a large number of unique values were coded using embedding layers separately for each variable,  categorical variable with a small number of unique values \textit{StoreType,  Assortment} were represented using one-hot encoding. 
Figure~\ref{model1} shows the  parameters of neural network layers. 
Figure~\ref{model2} shows the neural network structure.

\begin{figure}[H]
\center
 \includegraphics[width=0.55\linewidth]{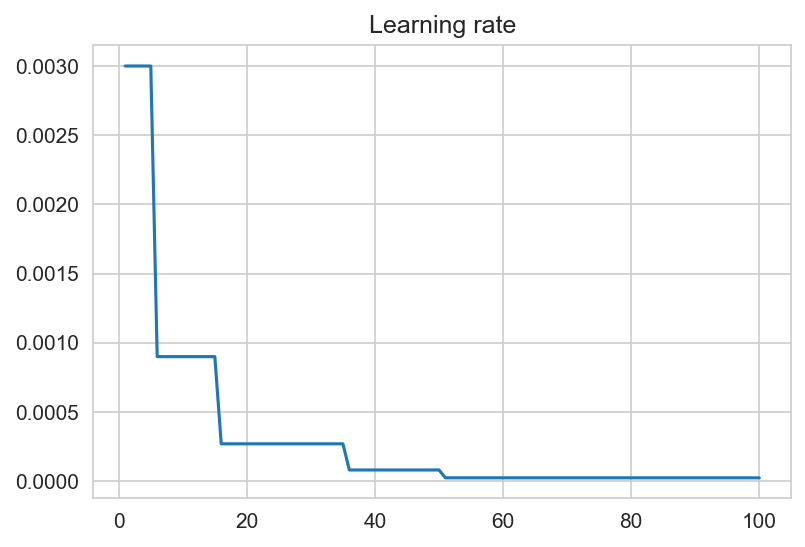}
 \caption{Learning rate changes with epochs}
 \label{img14}
 \end{figure}

\begin{figure}[H]
\center
 \includegraphics[width=0.55\linewidth]{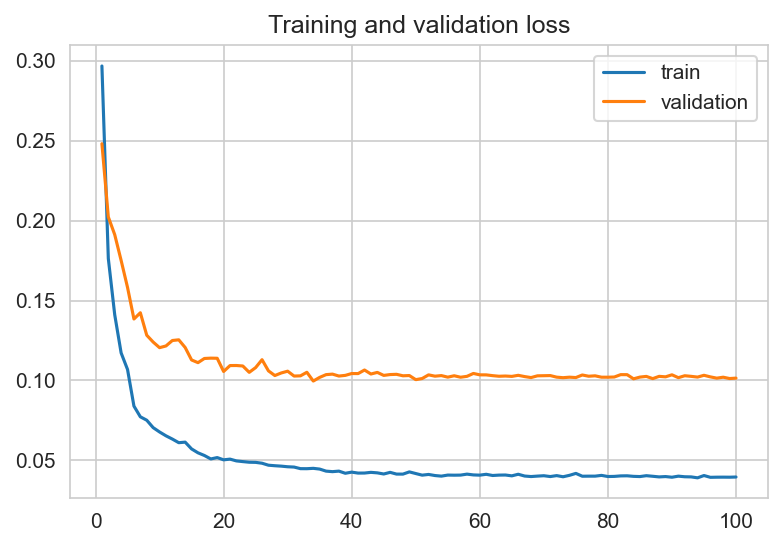}
 \caption{Train and validation loss  for model without trend correction block}
 \label{img15}
 \end{figure}

\begin{figure}[H]
\center
 \includegraphics[width=0.55\linewidth]{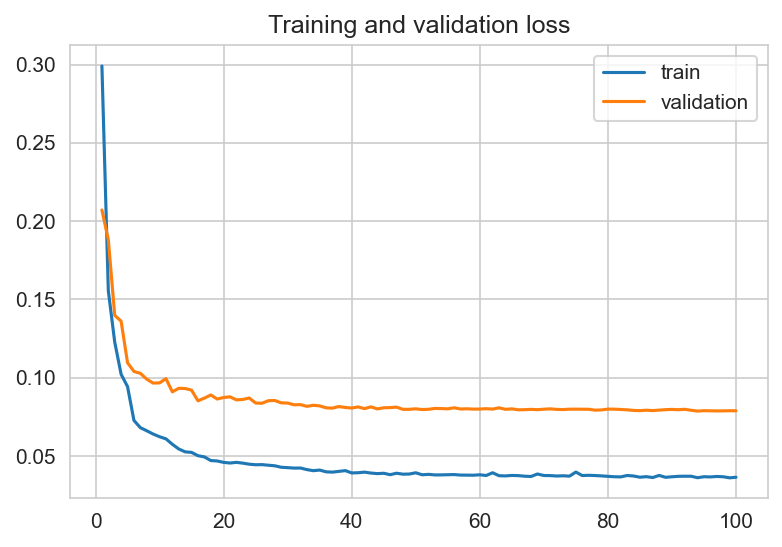}
 \caption{Train and validation loss for model with trend correction block}
 \label{img17}
 \end{figure}

\begin{figure}[H]
\center
 \includegraphics[width=0.65\linewidth]{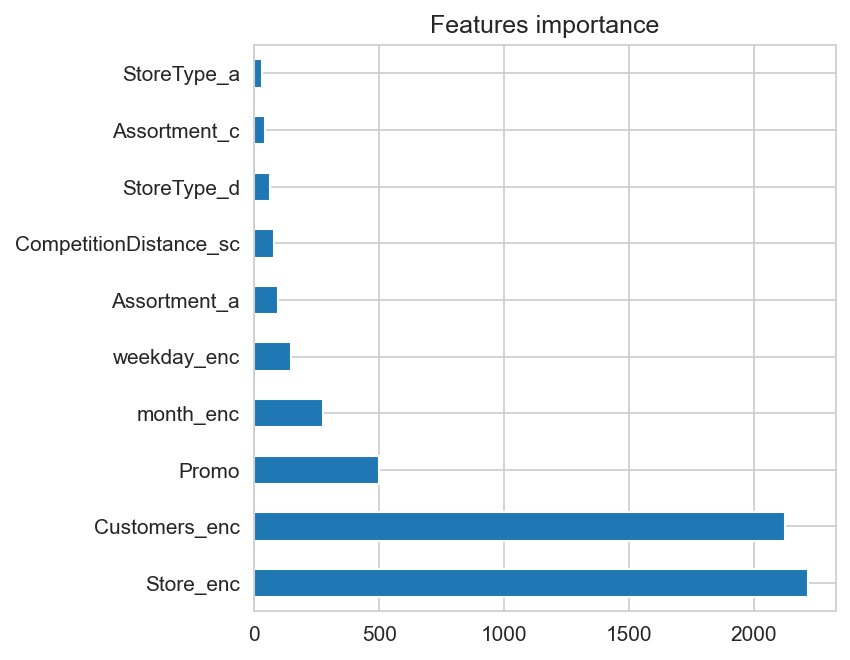}
 \caption{Features importance}
 \label{img18}
 \end{figure}

 The tensors of the output of embedding layers and numerical input values are concatenated 
and connected with the linear layer with \textit{ReLU} activation that forms an input block.  The output of this input block  is directed to the main block which consists of fully connected linear layers with \textit{ReLU} activation and dropout layers for predicting sales values using the output linear layer.  The output of the input block  is also directed to the trend correction block which consists of fully connected layer with ReLU activation,  dropout layer and output linear layer for the prediction of the trend weight.  The time trend term which  is a normalized time value multiplied by the trend weight is added to 
the predicted sales values.  For the comparison, we have considered two cases of the model with and without trend correction block.  Let us consider the results of model training and evaluation.  
Figure~\ref{img14} shows learning rate changes with epochs. 
Figure~\ref{img15} shows train and validation loss  for the model without trend correction block,  
Figure~\ref{img17} shows these losses  for the model with the trend correction block.
Figure~\ref{img18} shows the features importance which has been received  using the permutation approach. 
 
\begin{figure}[H]
\center
 \includegraphics[width=0.65\linewidth]{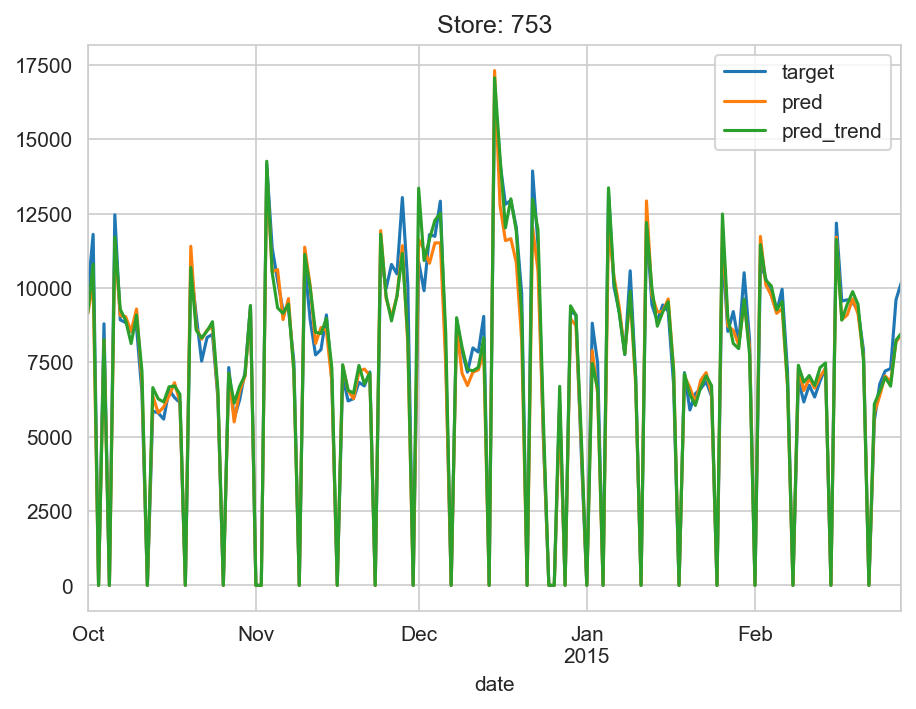}
 \caption{Store sales time series on validation dataset,  predicted  using model without (\textit{pred}) and with  (\textit{pred\_trend}) trend correction block ($RMSE=706, RMSE_{trend}=646$) }
 \label{img22}
 \end{figure}

\begin{figure}[H]
\center
 \includegraphics[width=0.65\linewidth]{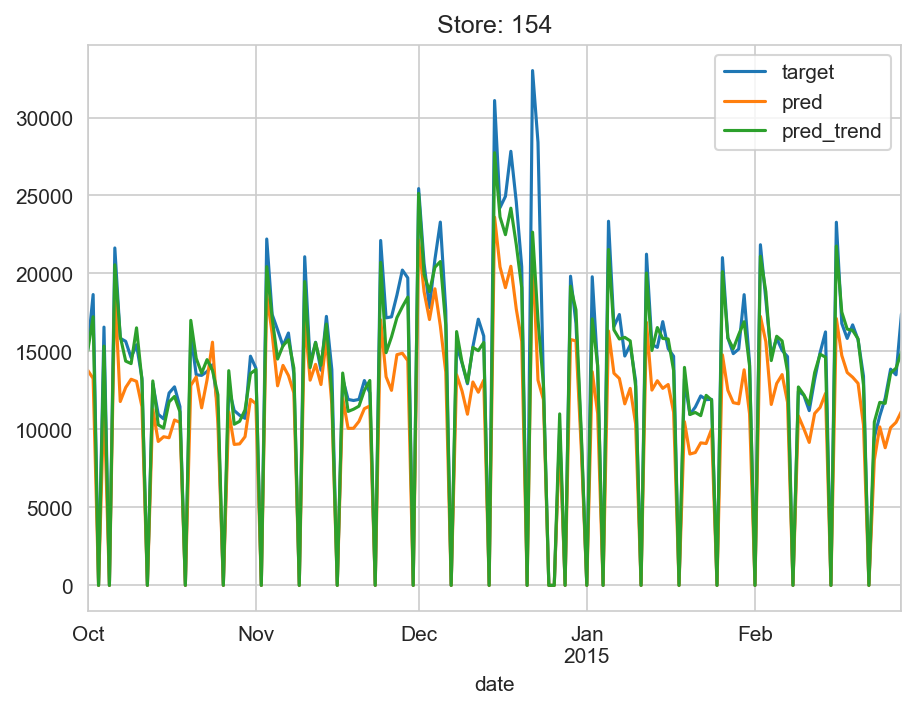}
 \caption{Store sales time series on validation dataset,  predicted  using model without (\textit{pred}) and with  (\textit{pred\_trend}) trend correction block ($RMSE=3699, RMSE_{trend}=1753$) }
 \label{img23}
 \end{figure}

\begin{figure}[H]
\center
 \includegraphics[width=0.65\linewidth]{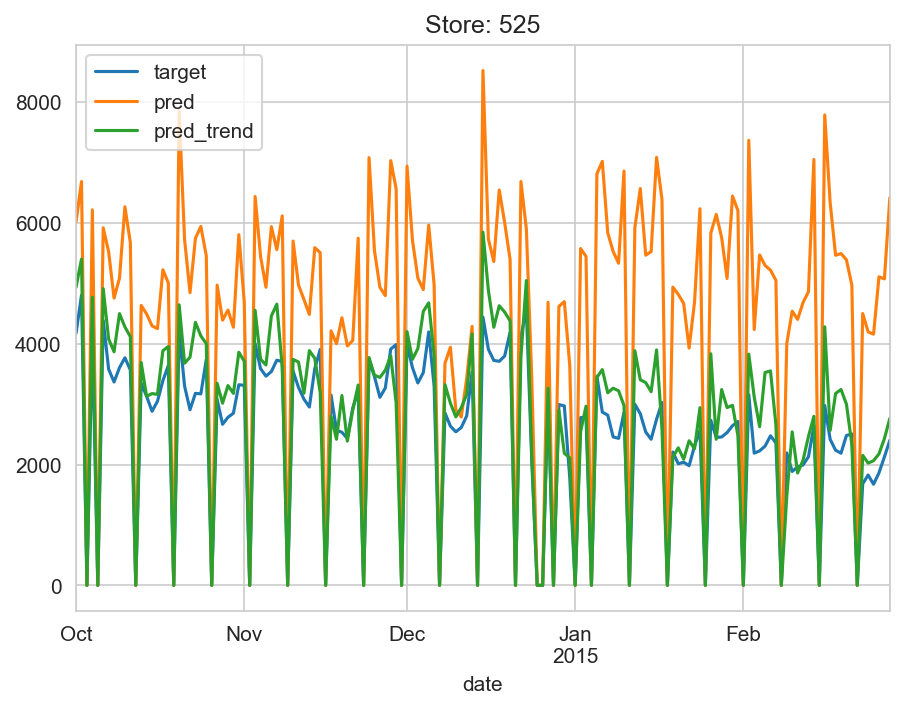}
 \caption{Store sales time series on validation dataset,  predicted  using model without (\textit{pred}) and with  (\textit{pred\_trend}) trend correction block ($RMSE=2515, RMSE_{trend}=572$) }
 \label{img24}
 \end{figure}

 Numerical features and target variables before feeding into the neural network were normalized by extracting their mean values from them and dividing them by their standard deviation. 
 The forecasting score over all stores is 
 $RMSE=1076$, for the model without the trend correction block and  $RMSE_{trend}=943$ for the model with the trend correction block.
 We can see a small improvement in the accuracy on the validation set for the model  with the trend correction block.  
 Figures~\ref{img22}-\ref{img24} show aggregated store sales time series on the validation dataset,  which was predicted  using the model without and with  the  trend correction block.  These results of predicted sales correspond to time series which are  shown  in Figures~\ref{img1}-\ref{img3}.   One can see that  for some stores sales with time trend,  the forecasting accuracy  can be essentially improved using the trend correction block.  
 
\subsection{Conclusions}
Applying of  machine learning for non-stationary sales time series with time trend can cause a forecasting bias.  
The approach with the trend correction block in the deep learning model for sales forecasting has been considered.  The model predicts sales values simultaneously with the weight for the time trend term.  The time trend term is considered as a product of the predicted weight value and normalized time value and then added to predicted sales values.  As a result, an optimized weight for the time trend for different groups of sales data can be received, e.g.  for each store with  the intrinsic time trend, the optimized weight for the time trend can be found. 
The results show that 
 the forecasting accuracy can be essentially improved for non-stationary sales with time trends using the trend correction block in the deep learning model.

\section{Sales Time Series Analytics Using Deep Q-Learning}

In this case study,  we consider the use of deep Q-learning models  in the problems of sales time series 
analytics. In contrast to supervised machine learning which is a kind of passive learning using historical data, 
Q-learning is a kind of active learning with goal to maximize a reward by optimal sequence of actions. 
Model free Q-learning approach for optimal pricing strategies and supply-demand problems was considered in the work. The main idea of the study is to  show that using deep Q-learning approach in time series analytics, the sequence of actions can be optimized by maximizing the reward function when the environment for learning agent interaction can be modeled using the parametric model and in the case of using the model which is based on the historical data. 
In the pricing optimizing case study environment was modeled using sales 
dependence on extras price and randomly simulated demand.
  In the pricing optimizing case study, the environment was modeled using sales dependence on 
  extra price  and randomly simulated demand.  
 In the supply-demand case study, it was proposed to use historical demand time series 
 for environment modeling, agent states were represented by promo actions, previous  demand values and weekly seasonality features. 
 Obtained results show that using deep Q-learning, we can optimize the decision making process for price optimization and supply-demand problems.  Environment modeling using parametric models and historical data can be used for the cold start of learning agent. On the next steps, after the cold start, the trained agent can be used in real business environment.   
 
\subsection{Introduction}
Sales time series analytics is an important part of modern business intelligence. 
We can mention classical and popular  time series models - Holt-Winters, ARIMA, SARIMA, SARIMAX, GARCH, etc. Different time series models and  approaches can be found in 
~\cite{box2015time, doganis2006time,hyndman2018forecasting,tsay2005analysis}.
In ~\cite{pavlyshenko2019machine} we 
 studied the use of machine-learning models for sales predictive analytics. We considered the main approaches and case studies of using machine learning for sales forecasting, effect of machine-learning generalization. In this paper, we also considered a stacking approach for building regression ensemble of single models. 
In~\cite{pavlyshenko2016linear}, we studied linear models, machine learning and probabilistic models for time series modeling. 
For probabilistic modeling, we considered the use of copulas and Bayesian inference approaches. 
In~\cite{pavlyshenko2016machine}, we studied the logistic regression in the problem of  detecting manufacturing failures. 
For the logistic regression, we considered a generalized linear model, machine learning  and Bayesian models. 
In~\cite{pavlyshenko2018using}, we studied stacking approaches for time series forecasting and logistic regression with highly imbalanced data. 
The use of regression approaches for sales forecasting can often give us better results compared to time series methods. One of the main assumptions of regression methods is that the patterns in the historical data will be repeated in future. 
Supervised machine learning can be considered as a kind of passive learning using historical observations. 
Time series forecasting using Machine Learning gives us insights and allows us to make right business decisions.
Reinforcement learning allows us to find sequences of optimized actions directly without historical data. 
In this approach, we have an environment and a learning agent which interacts with environment.
As a result of each interaction, learning agent receives reward. The goal of Reinforcement 
Learning  is to find such sequence of actions which will maximize an average cumulative  reward on the episodes of agent-environment interactions. There are policy based and policy free approaches. Policy can be described by parametrized distribution function for states and actions. 
The parameters of these distributions can be found using policy gradient approach where on each iteration, we calculate the gradient of objective function. 
Policy free approach can be Q-learning which is based on Bellman equation ~\cite{sutton1998introduction, mnih2015human, mnih2013playing} .
On each iteration, we upgrade the Q-table where rows represent states and columns represent actions. 
In the case of continuous action, space Q-table can be approximated by Neural Network using DQN approach ~\cite{mnih2015human, mnih2013playing}.
\subsection{Related Work}
The main principles of reinforcement learning can be found at
~ \cite{sutton1998introduction}. In ~\cite{mnih2015human, mnih2013playing} deep Q-learning approaches were studied. 
In ~ \cite{rana2014real}, real-time dynamic pricing in a non stationary
environment has been considered. In the article, the problem of establishing a pricing policy that maximizes the revenue for selling a given inventory by a fixed deadline has been considered.
In ~ \cite{maestre2018reinforcement} reinforcement learning for fair dynamic pricing has been considered.
In ~ \cite{vengerov2007gradient} , 
 Reinforcement Learning algorithm that can tune parameters of a seller’s dynamic pricing policy in a gradient direction  even when the seller's environment is not fully observable was considered.
Different approaches for dynamic pricing is considred in ~ \cite{den2015dynamic}. In the paper ~\cite{kim2005adaptive} adaptive inventory-control models for a supply chain consisting of one supplier and multiple retailers were proposed. The paper ~\cite{ raju2003reinforcement} describes 
 the use of reinforcement learning techniques in the problem of determining dynamic prices in an electronic retail market. The papers ~\cite{huang2018financial, jiang2017deep} consider the use of reinforcement learning for financial  analytics.  
The paper ~\cite{liu2005neural}  formulates an autonomous data-driven approach to identify a parsimonious structure for the NN so as to reduce the prediction error and enhance the modeling accuracy. The Reinforcement Learning based Dimension and Delay Estimator (RLDDE) was proposed. The paper ~\cite{jiang2017deep} presents a model-free Reinforcement Learning framework to provide a deep machine learning solution to the portfolio management problem. In ~\cite{dulac2015deep}, 
Deep Reinforcement Learning in Large Discrete Action Spaces was considered.

\subsection{Deep Q-Learning Approaches}
The goal of Q-learning is to maximize cumulative future reward ~\cite{sutton1998introduction,mnih2015human, mnih2013playing}. 
To training the Q-learning network, the gradient descent algorithm is often used. 
To eliminate influence between sequence data and non-stationary  distribution,
the replay mechanism ~\cite{lin1993reinforcement} can be used.
This approach consists in random sampling of previous data which represent 
states and actions.  It makes it possible to average distributions of data which describe previous agent's behaviour. The goal for the agent lies in choosing a sequence action strategy which maximizes future rewards ~\cite{mnih2013playing}. 
An optimal action-value function can be considered as:
\begin{equation}
Q^*(s,a)=\mathbb{E}_{s' \sim \varepsilon}\left[ r+ \gamma \underset{a'}{\max}Q(s',a') 
\mid s,a \right] 
\end{equation} 
where $r$ is a reward, $s$ is a state, $a$ is an action, $s', a'$ are possible states and actions on the next time step.
To estimate the function $  Q^*(s,a) $ approximately, Bellman equation can be used in the iteration process:
 \begin{equation}
Q_{i+1}(s,a)=\mathbb{E}\left[ r+ \gamma \underset{a'}{\max}Q_{i}(s',a') 
\mid s,a \right] 
\end{equation} 
The main problem of such an approach is that there is no generalization of revealed patterns in the agent-environment interaction due to the fact that 
 $  Q(s,a) $ is estimated on each separate step. 
 To improve generalization, one can use an approximation function for 
 $  Q^*(s,a) $.  Neural network can be used for this purpose. 
 Parameters of such deep Q-network can be found with gradient methods minimizing 
 the loss function
 \begin{align}
 &L_i(\theta_i)=\mathbb{E}_{s,a \sim \rho}\left[ (y_i-Q(s,a;\theta_i))^2 \right] \\
 &y_i=\mathbb{E}_{s' \sim \varepsilon}\left[ r+ \gamma \underset{a'}{\max}Q(s',a',\theta_{i-1}) 
\mid s,a \right] 
\end{align} 
where $  \rho $ - is the behaviour distribution, $\theta$ is the weights of Q-network.

Experience replay  approach ~\cite{lin1993reinforcement}  is effectively used in deep Q-network ~\cite{mnih2013playing,mnih2015human}.
In this approach, agent actions and states are stored in the replay memory at each time step as tuples 
$e_t=(s_t,a_t,r_t,s_{t+1})$. Tuples $e_t$ are stored in the data set $D_t=\lbrace e_1,...e_t \rbrace$.
On each step for Q-learning updates, the algorithm gets samples $e_t$ from the replay memory by uniform random sampling $e_t \sim U(D)$ ~\cite{mnih2013playing}. 
  On the next experience replay step, Q-learning updates  
the weights $\theta$. On the next step, the  agent selects an optimal action, using $ \varepsilon $-greedy policy ~\cite{mnih2013playing}.   
Data mini batches are formed on each iteration for updating the weights for 
Q-network.
Mini batches are chosen randomly. 
Such an approach provides generalization of approximation given  data for agent-environment interaction. 
Due to the experience replay approach, behavioural distribution is averaged 
on many previous states and actions of learning agent that provides the convergence of iteration process. 
One of widely used approaches for Q-network consists in treating agent states as input parameters when outputs are $Q$-values for each separate agent action ~\cite{mnih2013playing}. 
In such an approach,   $Q$-value for each action is calculated during one Q-network  step forward. 

In this work, we consider two cases of using Q-learning in sales time series analytics. 
One case is an optimal pricing strategy, the second case is supply and demand problems which appear in retail domain areas. In our numerical experiments, we used the algorithms which were based on model architecture described in ~\cite{mnih2015human, mnih2013playing} and the approaches for Q-learning agent implementations from ~\cite{githubrep1,githubrep2,githubrep3}. For environment modeling,
we used the parametric models in the case of price optimization and historical time series of demand in the case of supply-demand problem.  
For the analysis of supply-demand problem, we used the store sales historical data from  'Rossmann Store Sales' Kaggle competition ~\cite{rossmanstorekaggle}. These data describe sales in Rossmann stores.  The calculations were conducted in the Python environment using the main packages \textit{pandas, sklearn, numpy, keras, matplotlib, seaborn}.  To conduct the analysis, \textit{Jupyter Notebook} was used.

\subsection{Optimal  Pricing Strategy using Q-Learning}
\begin{figure}
\centerline{\includegraphics[width=0.5\textwidth]{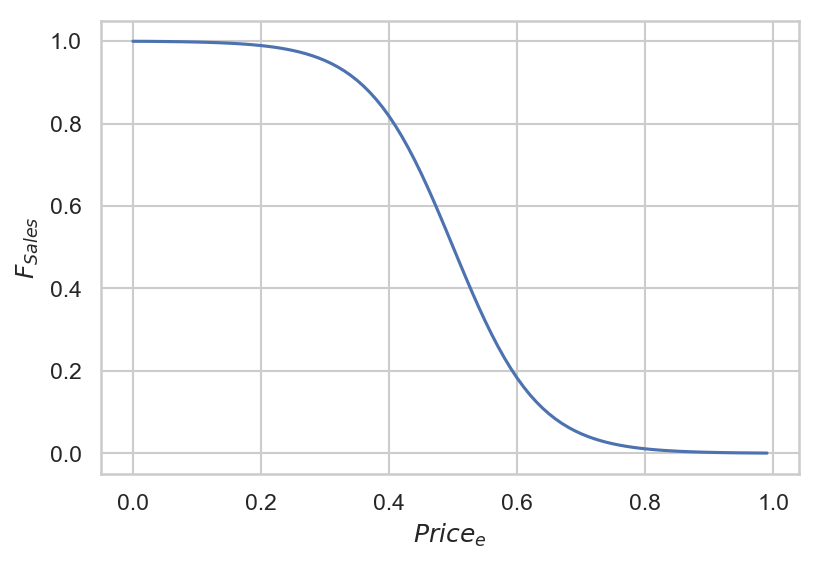}}
\caption{Sales vs extra prices}
\label{ql_fig1}
\end{figure}

Let us consider a simple case for the pricing strategy.  
The question is what strategy can be applied to maximize profit for some time period. 
Generally, the strategy can consist of many factors and means. In the simplest model, it consists of discrete extra price values which are the fraction of cost price. Dependence between sales and extra price can be considered as
 \begin{equation}
 F_{Sales}=\frac{a}{(1+b \cdot exp(c \cdot( Price_m\cdot(1+Price_{e})-d)))},
 \label{opt_price_f1}
 \end{equation}
 where $Price_m$ is an marginal price, $Price_{e}$ is an  extra price, $a,b,c,d$ are the  parameters for 
 function $F_{Sales}$. 
 The function $F_{Sales}$ ~(\ref{opt_price_f1}) describes relative decreasing of sales with an increasing extra price. Extra price is considered in relative parts of marginal price.   
 The reward function for Q-learning can be considered as 
 \begin{equation}
 Reward=Demand \cdot  F_{Sales} \cdot Price_e
  \label {opt_price_f2}
 \end{equation}
We created a simple model with normalized variables. The Figure~\ref{ql_fig1}  shows sales vs extra prices ($Price_m$=1, a=1, b=1, c=15, d=1.5).
We can observe that for high extra price we get small sales. The real parameters for the logistic curve can be found by the gradient method using historical data. 
\begin{figure}
\centerline{\includegraphics[width=0.5\textwidth]{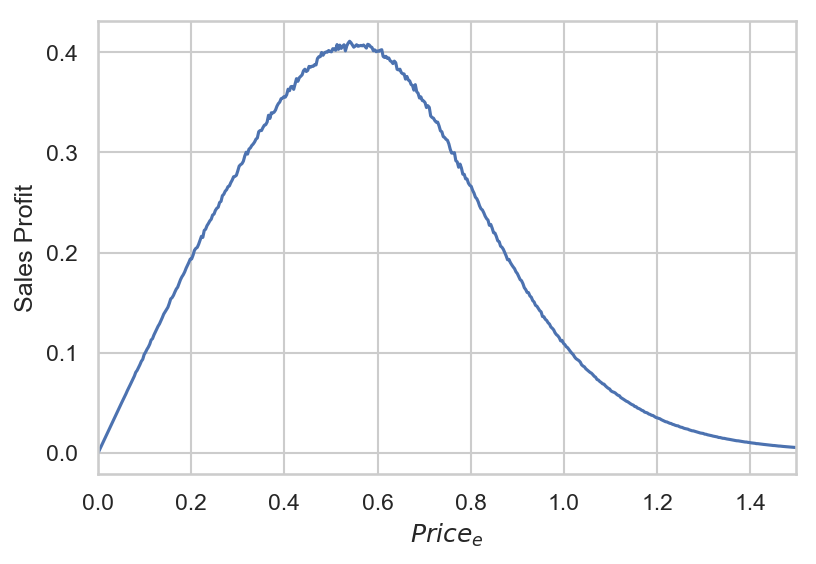}}
\caption{Profit versus extra prices}
\label{ql_fig2}
\end{figure}
\begin{figure}
\centerline{\includegraphics[width=0.7\textwidth]{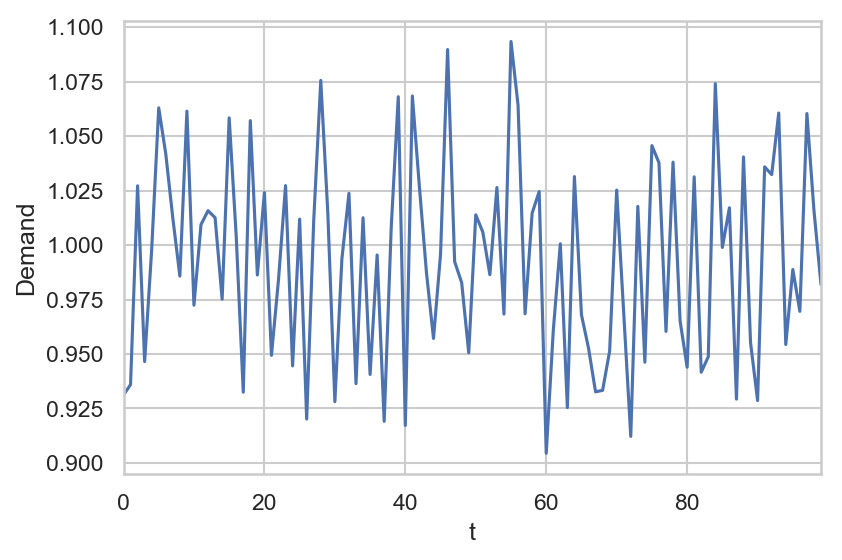}}
\caption{Simulated demand time series.}
\label{ql_fig3}
\end{figure}
 The Figure  ~\ref{ql_fig2}  shows profit versus extra prices.   One can see that there is no profit with low and high extra price.
 
The challenge is to find the optimal extra price. We can tabulate the objective function using simulated demand time series and find  the optimal value for extra price.
But in real cases, profit versus extra prices can have a complicated functional dependence, including the dependence on many qualitative factors which are included into complex multilevel 
pricing strategy. To find which pricing strategy is optimal, we can use a Q-learning approach. 
This simple model can be used for the cold start of learning agent before the interaction with real business environment.

Let us consider the parameters for numerical modeling.
For the Q-values approximation we used 
2-layers feed forward neural network with 32 neurons in each 
layer. Output neural network dimension is equal to the number of possible actions, which was chosen equal to  8.  For the actions, we took the list of the  following  values for normalized extra price [0, 0.15, 0.25, 0.5, 0.75, 0.85, 1, 1.5].
The extra price is considered in relative parts of the marginal price.  
As time steps we consider days.  
The number of time steps in the each episode equal to 7, the batch size was 32,
the number of learning iterations was 50, the optimizer is Adam, the learning rate for neural network was 0.001, the epsilon decay was 0.97.  For approximation of the function (~\ref{opt_price_f1}), we used the following parameters: $Price_m$=1, a=1, b=1, c=7, d=1.7

\begin{figure}
\centerline{\includegraphics[width=0.5\textwidth]{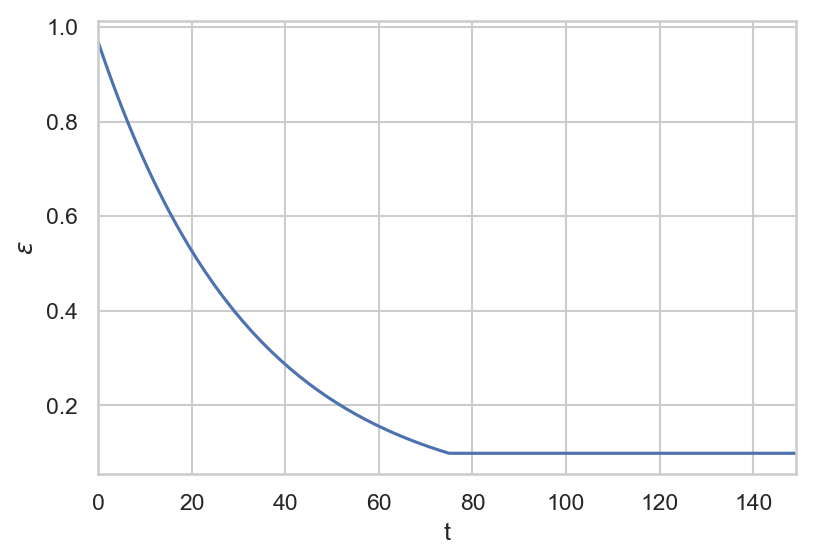}}
\caption{Epsilon vs time dependence}
\label{ql_fig4}
\end{figure}

\begin{figure}
\centerline{\includegraphics[width=0.5\textwidth]{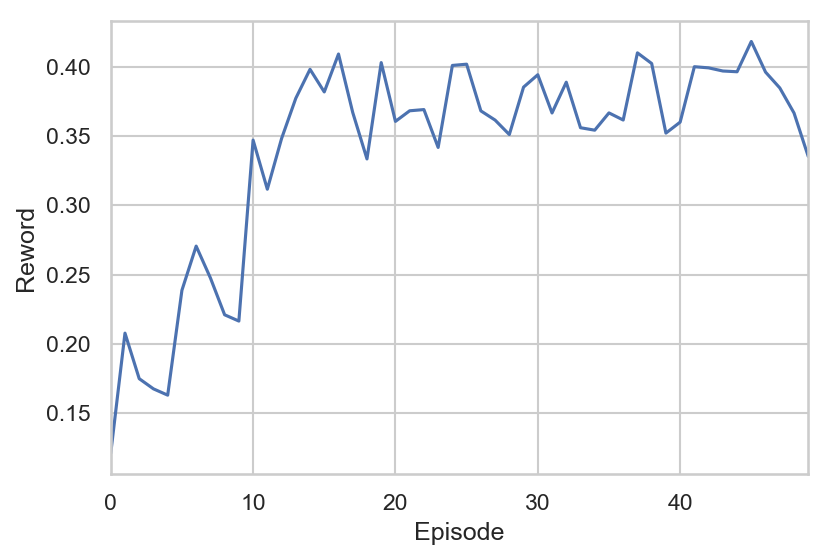}}
\caption{Mean reward over episodes}
\label{ql_fig5}
\end{figure}

\begin{figure}
\centerline{\includegraphics[width=0.5\textwidth]{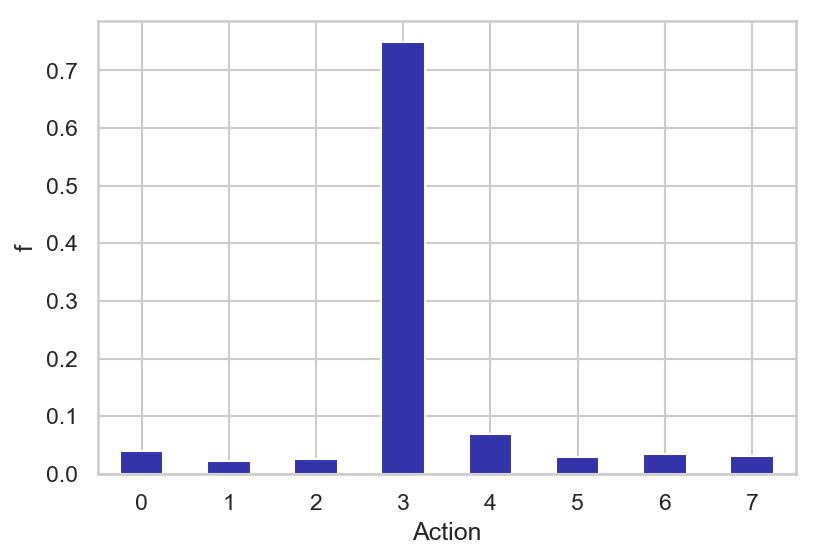}}
\caption{Frequencies of actions}
\label{ql_fig6}
\end{figure}

\begin{figure}
\centerline{\includegraphics[width=0.5\textwidth]{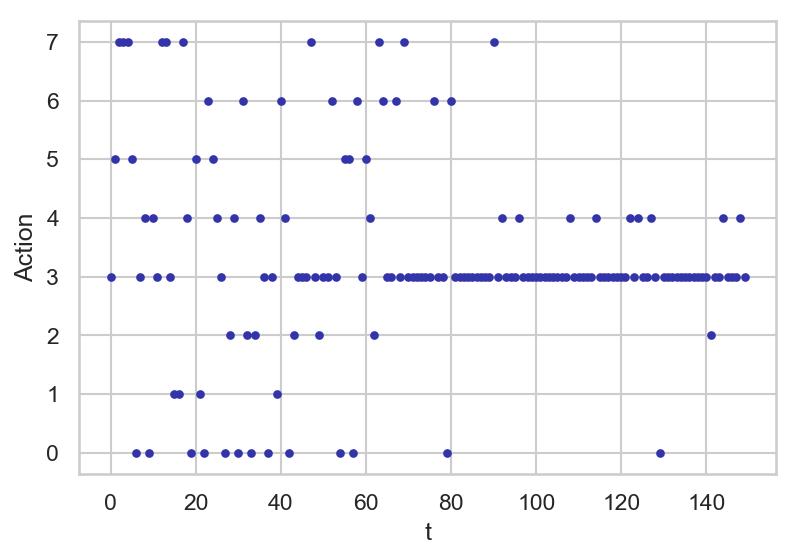}}
\caption{ Actions vs time}
\label{ql_fig7}
\end{figure}

In this case study, we used DQN approach with epsilon-greedy exploration-exploitation trade off. 
Epsilon describes the probability of random action. On each iteration, epsilon decreases. 
The Figure ~\ref{ql_fig4} shows epsilon vs time dependence. For the numerical experiment,  we simulated demand with random uniform distribution of relative units. As a  result of numerical experiment we received trained DQN model which can output optimal actions that maximize future cumulative reward. The Figure ~\ref{ql_fig3}  shows simulated demand time series. The Figure ~\ref{ql_fig5}  shows mean reward over the episodes.
One can observe that the reward  goes up with iterations. It means that the learning agent improves the way it interacts with the environment. 
The Figure ~\ref{ql_fig6} shows the frequencies of actions and we can see one dominated action which corresponds to the optimal pricing strategy for this simple model.
The Figure ~\ref{ql_fig7} shows the actions with time and we can see a big dispersion of actions at the beginning  of interaction process due to the domination of exploration type of interactions. 
We can observe that on the next steps after exploration time period, one action dominates which can be considered as a found optimal pricing strategy.

\subsection{Supply and Demand Case Study using Q-Learning}
\begin{figure}
\centerline{\includegraphics[width=0.5\textwidth]{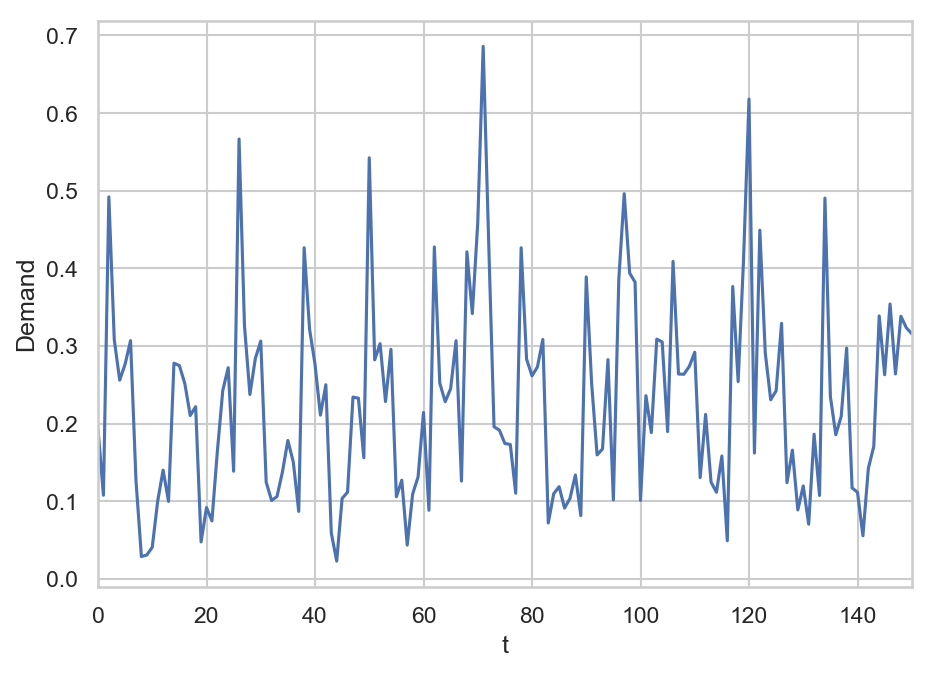}}
\caption{Simulated demand time series}
\label{ql_fig8}
\end{figure}

\begin{figure}
\centerline{\includegraphics[width=0.5\textwidth]{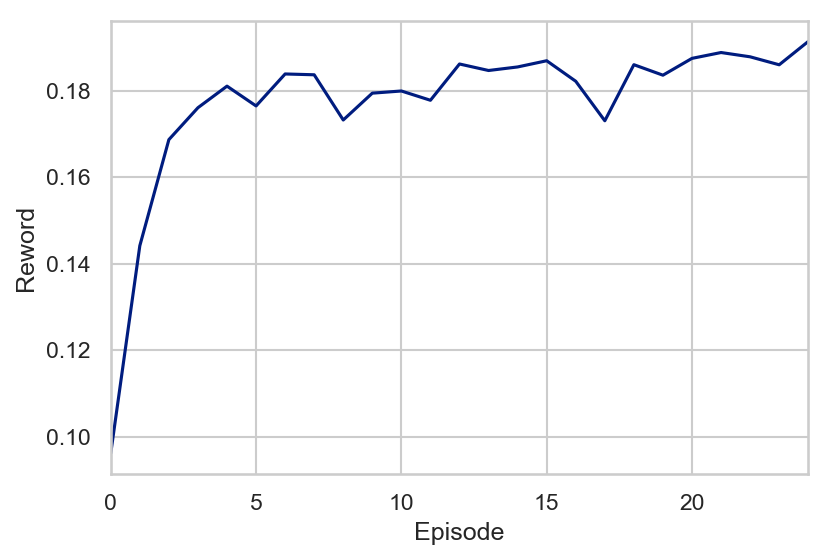}}
\caption{Mean reward over episodes}
\label{ql_fig9}
\end{figure}

Let us consider a case study of using Q-learning for supply and demand problems. 
We can also use historical data for the initial start of Q-learning algorithm along with parametric modeling of the environment. Such an approach makes it possible to conduct the cold start for Q-learning agents. In this case study, we took normalized demand time series with seasonality and promo action factor.
The challenge is to find optimal discrete actions in the supply-demand problem. Products can be supplied by batches with discrete amount. In the model, we took into account the  expenses for product processing which are related to  logistic, storing and other expenses.  
The reward on each step can be considered as 
\begin{equation}
 Reward=SalesProfit -  ProcessCost
\end{equation}
In this case study, we have more features for state representation.
They are  the product demand the day before, expected promo action, day of week. 
Let us consider the parameters for modeling supply-demand problem.
For modeling environment, we used the historical time series for demand. 
For each episode, the demand for 150 days was taken. 
This time series was simulated given the sales time series from   'Rossmann Store Sales' Kaggle competition ~\cite{rossmanstorekaggle}.
 To eliminate overfitting, random lag for the first point of demand time series was applied. The lag was calculated by uniform random distribution for integer values ranging from 0 to 25.
 For each episode, we used  the  time period of 150 days, the number of actions was 7. We used the feed forward neural network with 2 layers with 64 neurons in each layer. The batch size was 32, learning rate was 0.001, epsilon decay was 0.995, gamma coefficient  for Bellman equation was 0.3. To calculate the reward, we used the following parameters: the value of price profit was 1, pack unit was 0.05,  price support was 0.5. 
For state features we used a promo binary feature, 
previous day sales, weekly seasonality features which were 
7 binary features for each week day which were obtained by one hot encoding 
of day of  week categorical feature. 
We also set up generating stop episode flag when the reward became lower than    specified reward. 
Applying this rule accelerates the process of Q-learning, since in the cases 
of low current reward values learning process were stopped and new learning 
iteration was started.   
These actions were considered as discrete values for supply. For numerical modeling, 7 actions were chosen  with the following values [ 0,  2,  4,  6,  8, 10, 12], which are a discrete number of packs. The amount of simulated product in the pack was 0.025 of relative units. 
 The Figure ~\ref{ql_fig8} shows simulated demand time series for arbitrary chosen episode.
 For numerical experiment, demand is simulated in some relative units. 
The Figure ~\ref{ql_fig9} shows calculated mean reward over episodes.

\begin{figure}
\centerline{\includegraphics[width=0.5\textwidth]{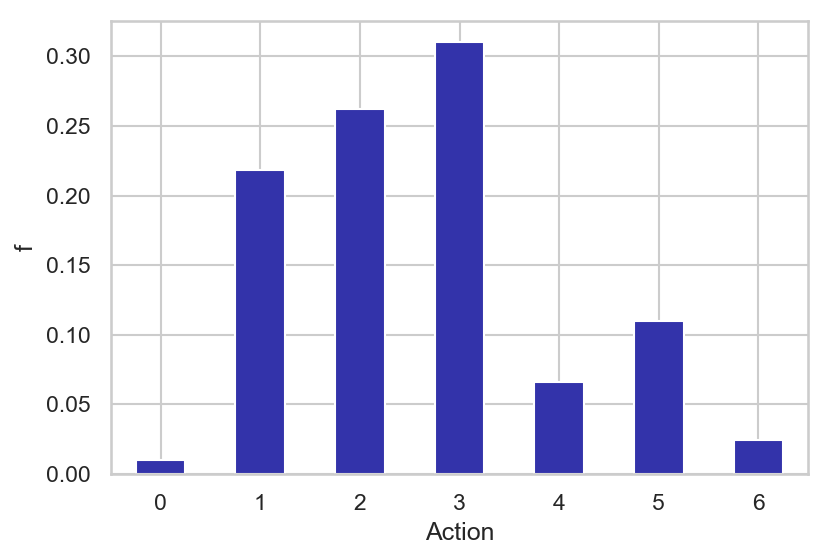}}
\caption{Frequencies of actions for supply-demand problem}
\label{ql_fig10}
\end{figure}

\begin{figure}
\centerline{\includegraphics[width=0.7\textwidth]{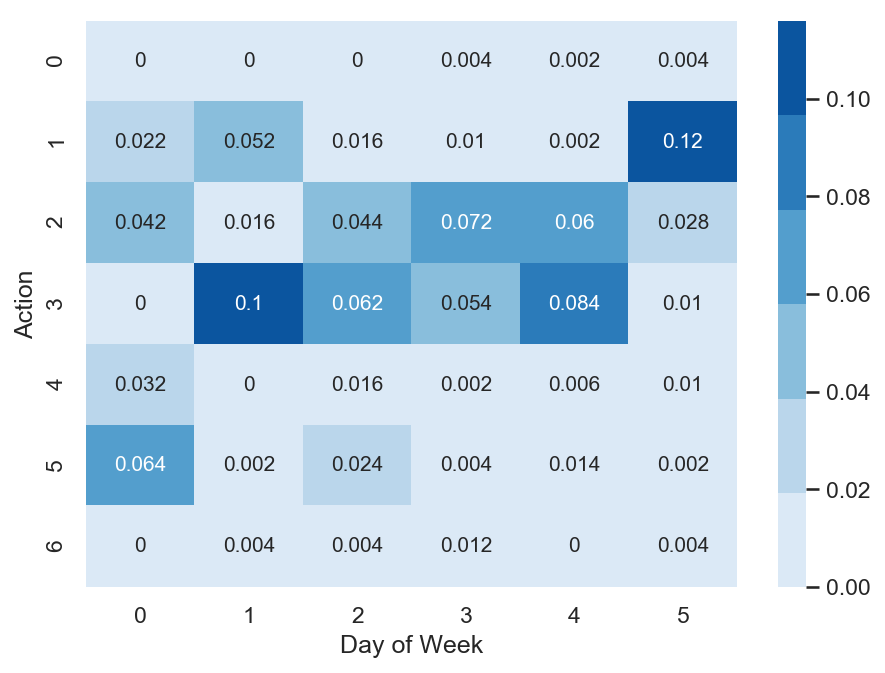}}
\caption{ Heatmap for action frequencies vs weekday}
\label{ql_fig11}
\end{figure}

We can observe that the learning agent optimizes action sequences over the episodes. In Figure ~\ref{ql_fig10}, we can see that several actions dominate in comparison with the previous case study for price strategy optimization where only one 
action dominated.
Here we have more features for state representation, so the learning agent chose different optimal actions for different states. 
The Figure ~\ref{ql_fig11} shows  the heatmap for action frequencies against weekday. 
\begin{figure}
\centerline{\includegraphics[width=0.7\textwidth]{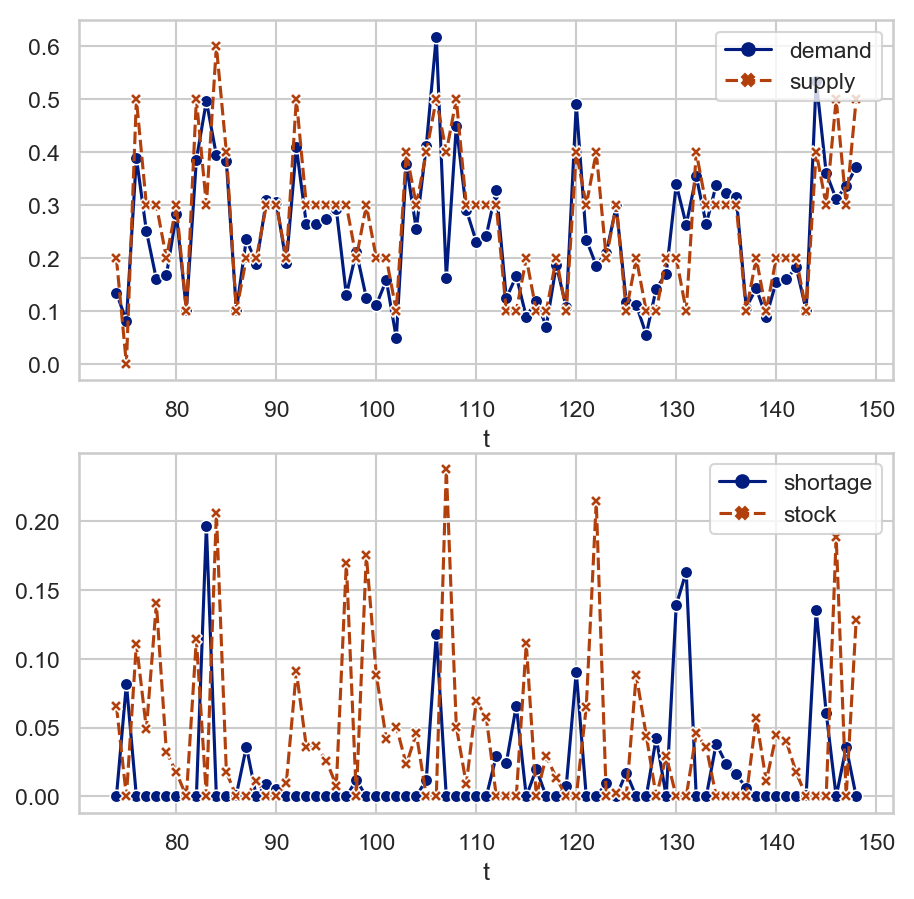}}
\caption{Time series of demand, supply, stock and shortage }
\label{ql_fig12}
\end{figure}
We can see that for different days we have different dominated actions which also depend on promo action which  can take place in different days. The Figure ~\ref{ql_fig12} shows  the time series of demand, supply, stock and shortage.
Dynamics of supply, shortage and stock depend on reward function which can be formed by by profit, expenses on logistic,  product processing and lost profit.  

\subsection{Conclusions}
The article describes the use of deep learning models for Q-learning in the problems of sales time series 
analytics. In contrast to supervised machine learning which is a kind of passive learning using historical data, 
Q-learning is a kind of active learning with goal to maximize the reward by optimal sequence of actions. 
Model free Q-learning approach for optimal pricing strategies and supply-demand problems was considered in the work. It was shown that using deep Q-learning 
approach, the sequence of actions can be optimized by maximizing the reward function. 
  In the pricing optimizing case study, the environment was modeled using sales dependence on 
  extra price  and randomly simulated demand.  
 In the supply-demand case study, it was proposed to use historical demand time series 
 for environment modeling, agent states were represented by promo actions, previous  demand values and weekly seasonality features. 
 Obtained results show that using deep Q-learning, we can optimize the decision making process for price optimization and supply-demand problems.  Environment modeling using parametric models and historical data can be used for the cold start of learning agent. On the next steps, after the cold start, the trained agent can be used in real business environment.   
 So, using Q-learning we can build a decision-making algorithm, which proposes qualitative decisions. This algorithm can start 
with the data simulated by parametric expert model or with the model based on historical data and then it can work with real business environment and do adaptation to the  changes in business processes. 
In more complicated cases, we can take into account the price for spoilt products in case we deal with perishable products and other expenses.

\section{Bitcoin Price Predictive Modeling  Using Expert Correction}
In this case study,  we consider the linear model for Bitcoin price which includes regression features based on Bitcoin currency statistics, 
mining processes, Google search trends, Wikipedia pages visits. 
The pattern  of deviation of regression model prediction  from real prices is simpler comparing to price time series. It is  assumed
that this pattern can be predicted by an experienced expert. In such a way, using the combination of the regression model and 
expert correction, one can receive better results than with either regression model or expert opinion only. 
It is shown that Bayesian approach makes it possible to  utilize the probabilistic approach using distributions with fat tails and take into account the outliers in Bitcoin 
price time series.

\subsection{Introduction}
One of the main goals in the Bitcoin analytics is price forecasting. There are many factors which influence the price dynamics. 
The most important factors are: interaction between supply and demand, attractiveness for investors, financial and macroeconomic indicators, technical indicators such as difficulty, 
the number of  blocks created recently, etc. A very important impact on the cryptocurrency price has trends in social networks and search engines. 
Using these factors, one can create a regression model with good fitting of bitcoin price on the historical data. 
Paper \cite{kristoufek2013bitcoin} shows that views of Wikipedia Bitcoin related pages correlate with Bitcoin price movements. It reflects
the potential investors' interests in  cryptocurrency. Google trends of search Bitcoin related keywords show different effects, including investors' interests, 
speculators' activities, etc.   
In ~\cite{bouoiyour2015bitcoin}, Bitcoin price was analyzed.
The paper ~\cite{bouoiyour2016drives} studies different drivers of Bitcoin price.
In ~\cite{matta2015bitcoin}, a significant  correlation  between Bitcoin price and social and web search media trends was shown.
In ~\cite{dyhrberg2016bitcoin},  the analysis shows that bitcoin has many similarities to both gold and the dollar. 
In ~\cite{shah2014bayesian},  the use of Bayesian regression for Bitcoin price analytics was  studied.  
The impact of news on investors' behavior  is studied in ~\cite{barber2007all}. 
Bitcoin economy, behavior and mining are considered in ~\cite{grinberg2012bitcoin, kroll2013economics}. 
Bitcoin market behaviour, especially  price dynamics is the subject of different studies ( \cite{ciaian2016economics, kristoufek2013bitcoin}).  
Different factors affecting bitcoin price are analyzed in the \cite{ciaian2016economics}. 
The specific feature of Bitcoin is that this cryptocurrency is neither issued nor controlled by financial or political institutions such as Central Bank, government, etc. 
 Bitcoin is being mined without economic underlying factors.  
In ~\cite{ciaian2016economics}, the economics of BitCoin Price Formation has been analyzed.   
Price dynamics is mostly affected by speculative behaviour  of investors. One of the main bitcoin drivers is the news in the Internet. 
According to efficient market theory, the stock and financial markets are not predictable, since all available information 
is already reflected in stock prices. But nowdays the dominance of efficient market theory is not so obvious.  Some influential scientists
argue that market can be partially predictable  (~\cite{malkiel2003efficient}). Behind modern market prediction, there are behavioral and psychological theories.
Some economists believe that historical prices, news, social network activities contain patterns that make it possible to partially predict financial market. 
Such theories and approaches are considered in the survey ~\cite{malkiel2003efficient}. 

In this work, we consider an approach for building regression predictive model for bitcoin price using expert correction by adding a correction term.
It is assumed that an experienced expert can make model correction relying on his or her experience. 
\subsection{Bitcoin Price Modeling}
\begin{figure}
\centerline{\includegraphics[width=1\textwidth]{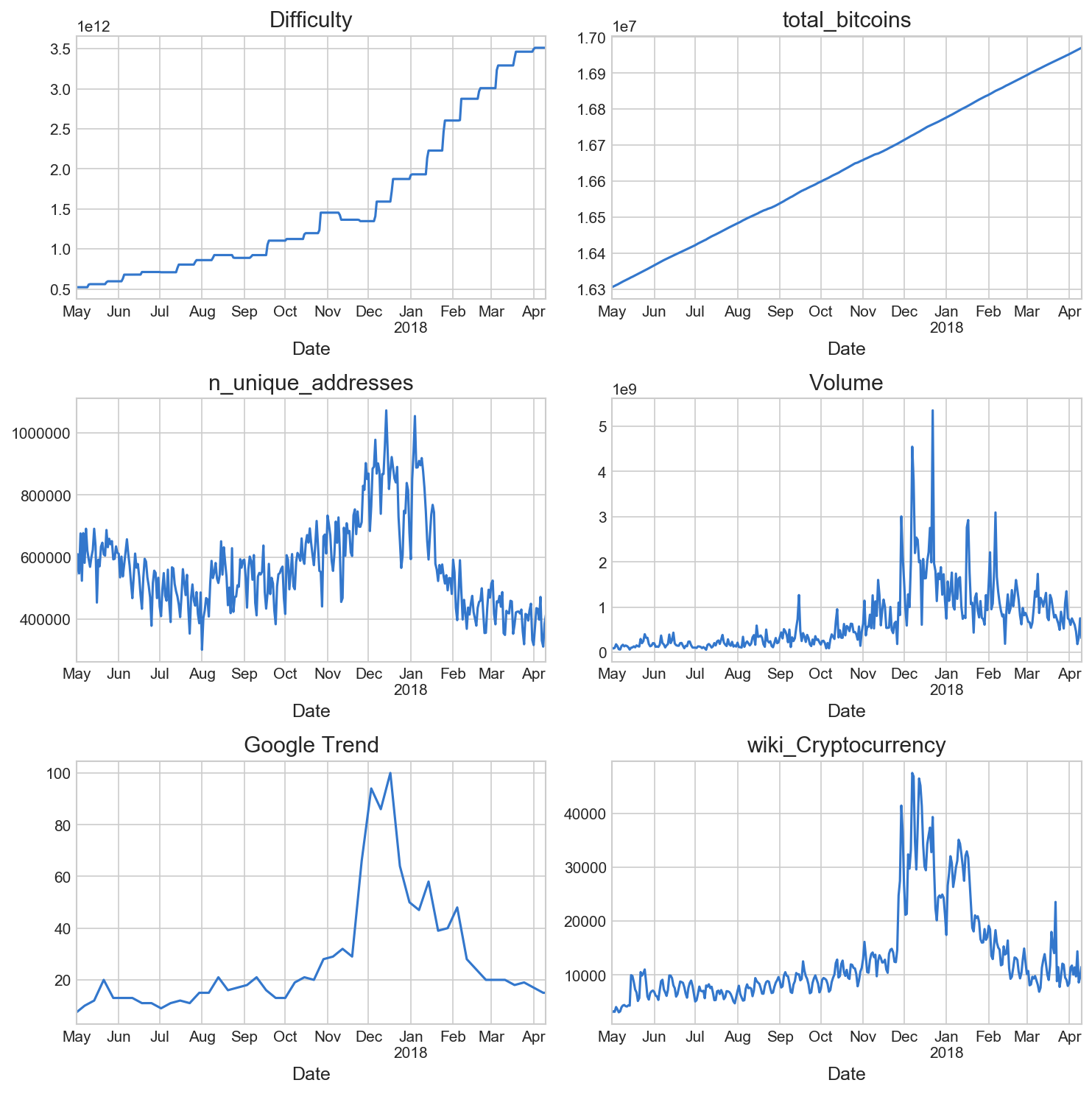}}
\caption{Time series of features}
\label{btc_fig1}
\end{figure}

\begin{figure}
\centerline{\includegraphics[width=1\textwidth]{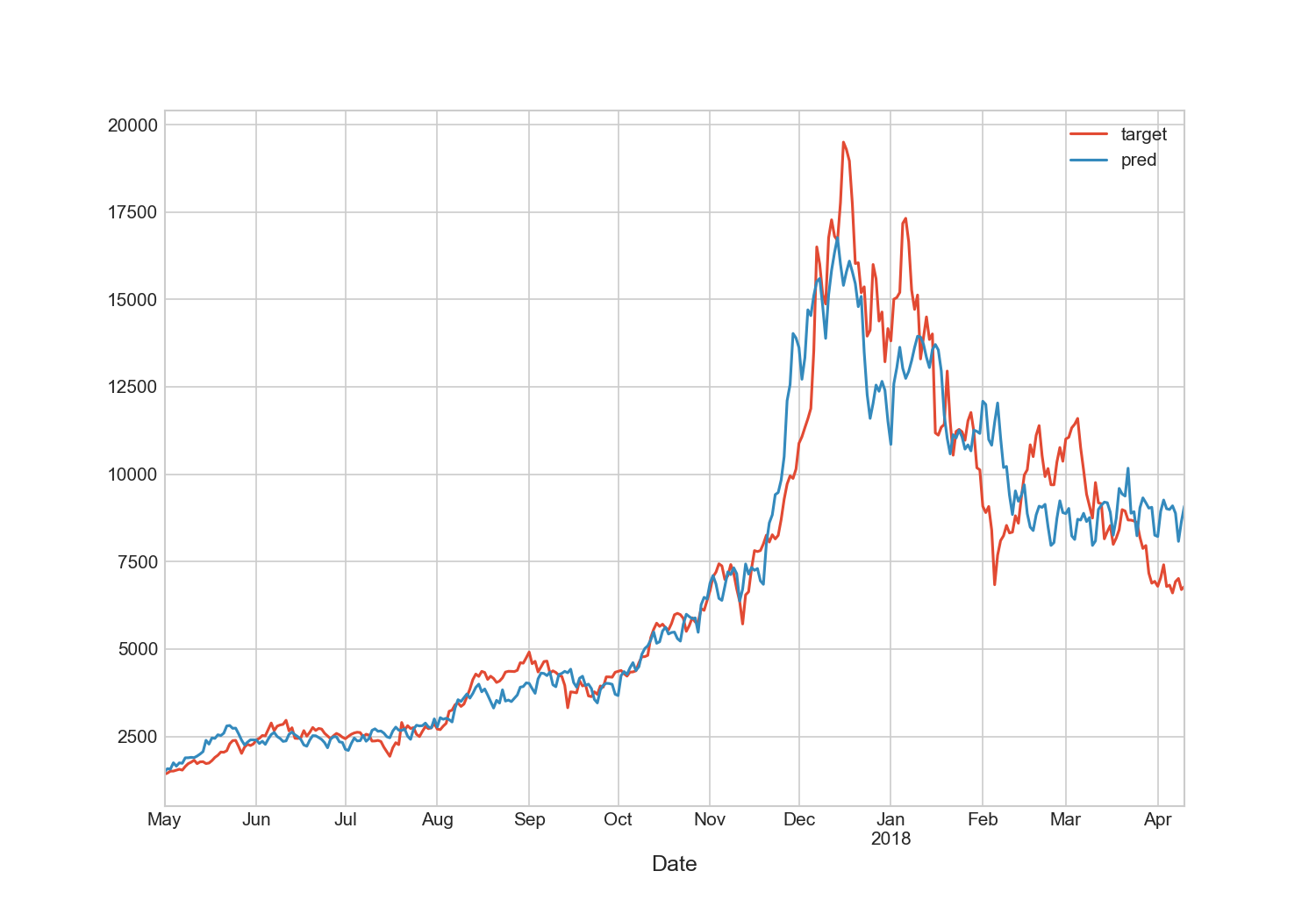} }
\caption{Bitcoin price dynamics and prediction}
\label{pred1}
\end{figure}

\begin{figure}
\centerline{\includegraphics[width=0.75\textwidth]{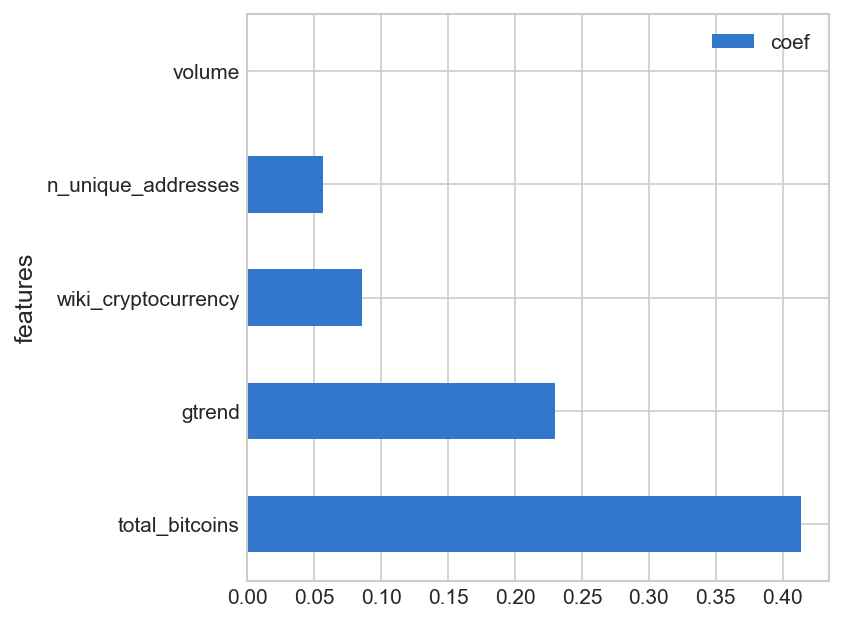}}
\caption{Features regression coefficients}
\label{coef1}
\end{figure}

Let us consider a regression model for Bitcoin price. 
As the regressors in our model, we used historical data which describe Bitcoin currency statistics, 
mining processes, Google search trends, Wikipedia pages visits. 
As the currency statistics,  we took 
$total\_bitcoins$ - the total number of bitcoins that have already been mined,
$price$ - average USD market price across major bitcoin exchanges,
$volume$ -the total USD value of trading volume on major bitcoin exchanges.
As mining information, we took 
$difficulty$, which is a relative measure of how difficult it is to find a new block.
As network activity, we took $n\_unique\_addresses$  which is the total number of unique addresses used on the Bitcoin blockchain.
Time series for mentioned above variables was taken from Bitcoin.info site. 
As the factors of price formation, we also considered Google trend for 'bitcoin'  keyword and  number of visits of  Wikipedia page about 'cryptocurrency'.
Time series for chosen features are shown on the Fig.~\ref{btc_fig1} .
A target variable $price$ and all the regressors are considered in the logarithmic scale, so  $price'=ln(price+1)$, $x_{i}'=ln(x_{i}+1)$ 
After logarithmic transformation all the regressors were normalized by extracting mean values and then dividing by standard deviation. 
We can write the regression model  as:
\begin{equation}
\label{e1}
\begin{split}
&price'=\alpha+\beta_{gtrend}\cdot gtrend'+ \\
&\beta_{wiki\_cryptocurrency}\cdot wiki\_cryptocurrency'+ \\
&\beta_{difficulty}\cdot difficulty + \\
&\beta_{n\_unique\_addresses}\cdot n\_unique\_addresses' + \\
&\beta_{total\_bitcoins}\cdot total\_bitcoins' +\\
& \beta_{volume}\cdot volume' 
\end{split}
\end{equation}
For the numerical modeling, we used Python with the packages \textit{pandas, numpy, scipy, matplotlib, seaborn , sklearn, quandl, pystan} . 
To get time series of Wikipedia pages visits, we used Python package \textit{mwviews}.
To find coefficients $\alpha, \beta_i$, we used linear regression with Lasso regularization from python package scikit-learn. 
Fig.~\ref{pred1} shows real dynamics of Bitcoin price  and the price predicted by  regression model (\ref{e1}). 
Fig.~\ref{coef1} shows obtained linear regression coefficients for features. 
The results show the importance of Google trends for Bitcoin keyword searches, and the importance of views of Wikipedia pages about the cryptocurrency. 
For this analyzed model,  we received the error value RMSE=1277.8. 

\subsection{Modeling of expert correction}
\begin{figure}
\centerline{\includegraphics[width=0.75\textwidth]{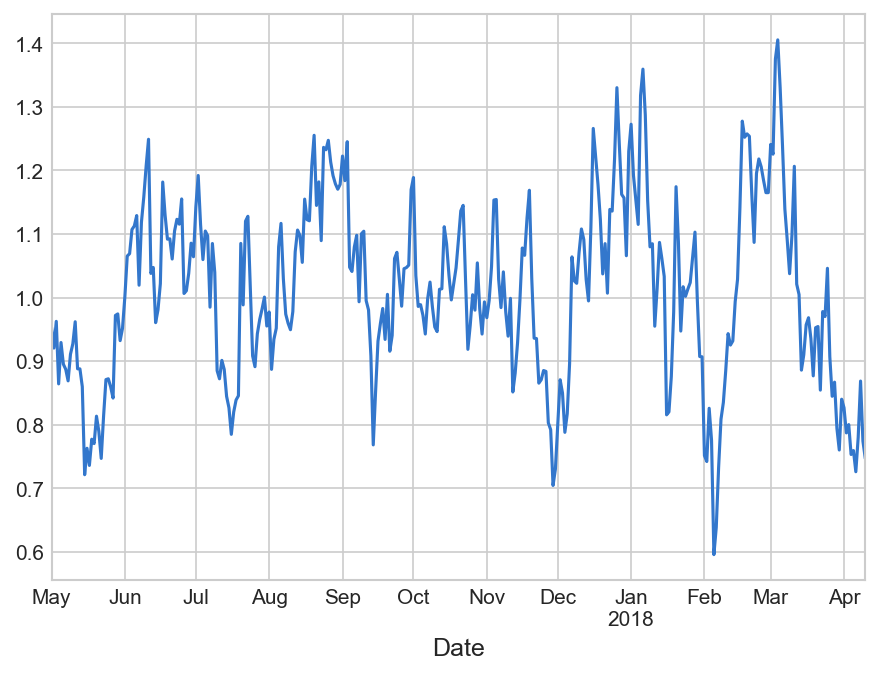}}
\caption{The ratio of real Bitcoin price to predicted price}
\label{targetpred}
\end{figure}

\begin{figure}
\centerline{\includegraphics[width=0.75\textwidth]{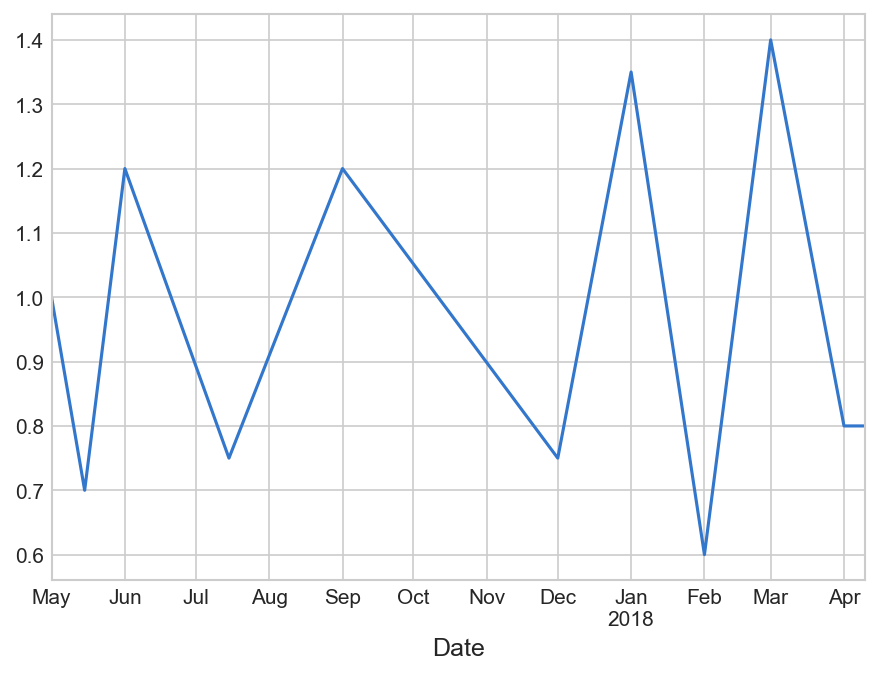}}
\caption{Expert correction term time series}
\label{expcorr}
\end{figure}

\begin{figure}
\centerline{\includegraphics[width=0.75\textwidth]{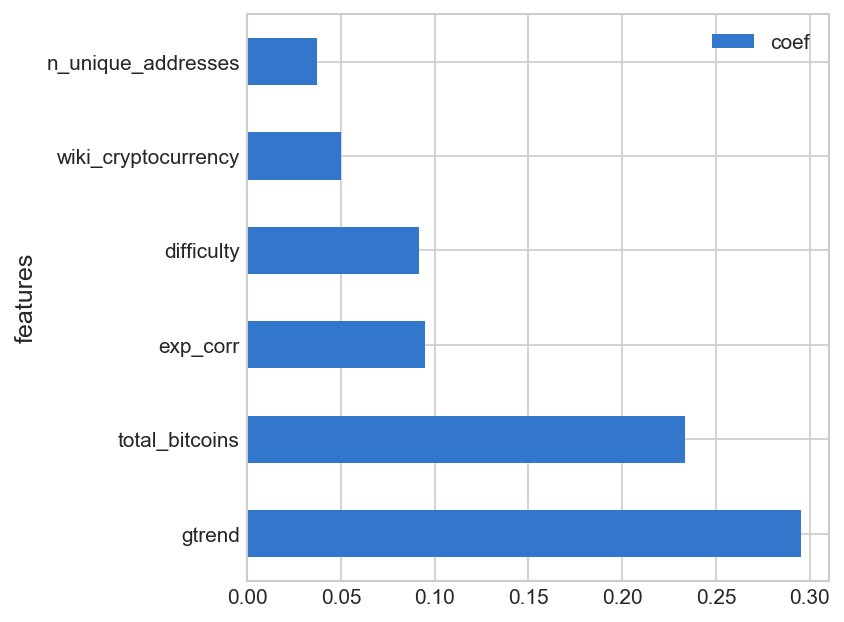}}
\caption{Feature coefficients for regression model with expert correction}
\label{featcoefcorr}
\end{figure}

 Fig.~\ref{pred1} shows that in some time periods, predicted price is higher comparing to real price, in some time periods it is lower. 
 Fig.~\ref{targetpred} shows the ratio of real and predicted price. We can see that this ratio has periodic oscillations which describe the deviation of predictions from real values. 
It can be explained by existence of some factors which impact price but they are not included into the model (\ref{e1}). 
It can be the factors of complex behavior of investors. Suppose that some experienced expert understands such type of behavior. As a result, he or she can explain the deviation dynamics 
of regression model (\ref{e1}) forecasting from real values. Expert correction can be involved into the model by an additional term in the regression model  (\ref{e1}).
This term describes the dynamics of model deviation. To describe this term, the expert needs to define the local extremum on deviation time series which are pivot points for deviation trends. 
We suppose that the expert can define such points correctly relying on his or her personal experience. The time series for possible expert correction is shown on Fig. \ref{expcorr}.
\begin{figure}
\centerline{\includegraphics[width=1\textwidth]{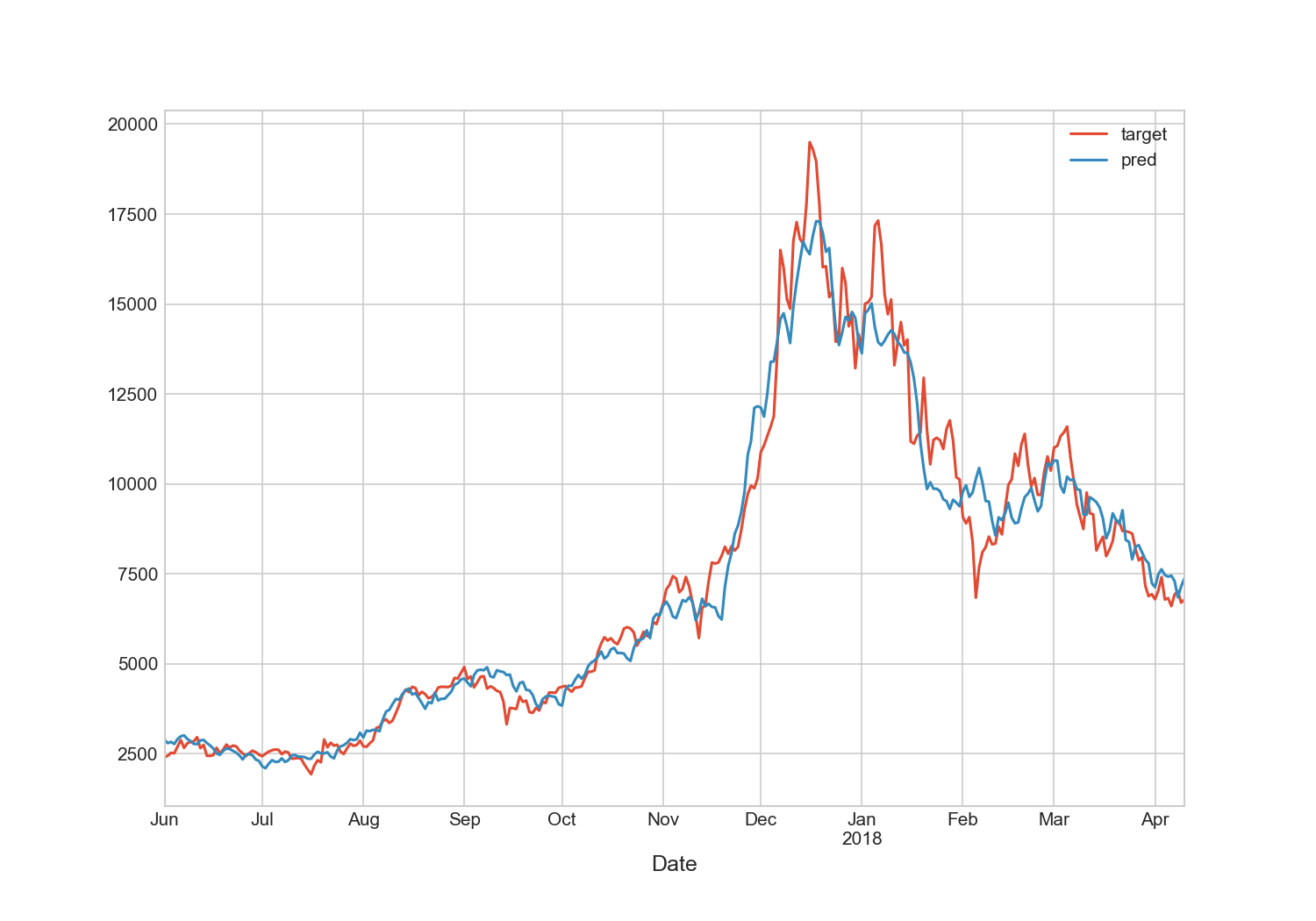}}
\caption{Bitcoin price dynamics \& prediction with expert correction}
\label{predcorr1}
\end{figure}
The results of the calculation of regression model  (\ref{e1}) with added such expert correction term are shown on the Fig.~\ref{predcorr1}.
For this case, we received the error value RMSE=856.4.  Fig.~\ref{featcoefcorr} shows the regression coefficient in case of expert correction term included into regression model.
Obtained results show that adding the expert term improves regression model performance. 
\subsection{Bayesian regression model}
\begin{figure}
\centerline{\includegraphics[width=1\textwidth]{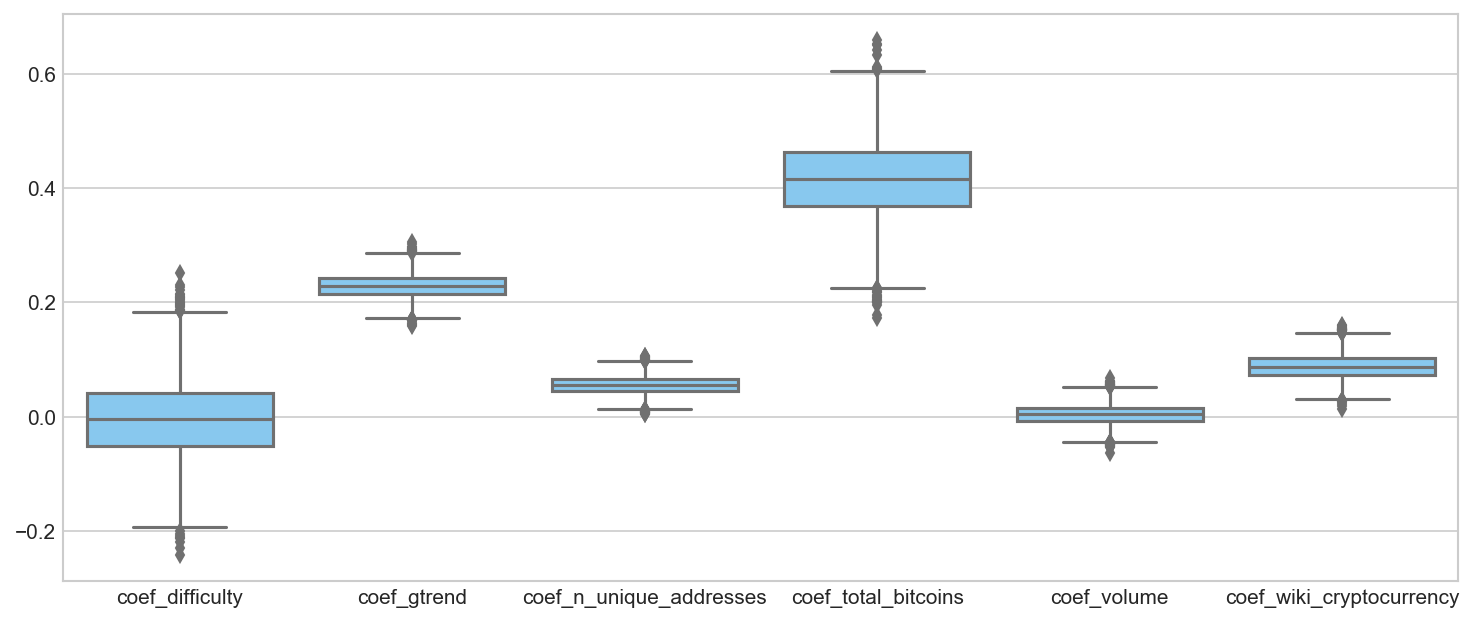}}
\caption{Boxplot for regression model coefficients}
\label{boxplot1}
\end{figure}

\begin{figure}
\centerline{\includegraphics[width=1\textwidth]{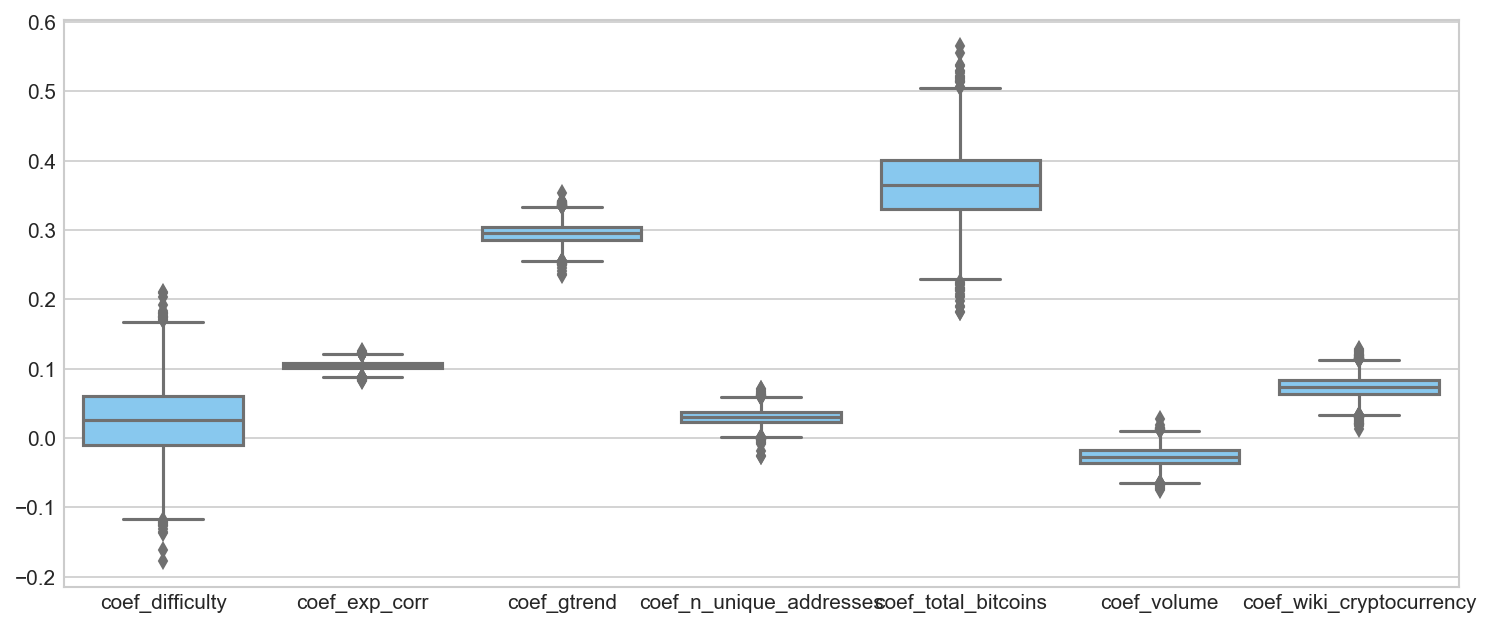}}
\caption{Boxplot for coefficients of regression model  with  expert correction term}
\label{boxplot1corr}
\end{figure}

For probabilistic approach, which makes it possible to get risk assessments, one can use Bayesian inference approach. 
Bayesian regression is a method which gives advantages when we need to take into account non Gaussian statistics for 
the variables under study (~\cite{kruschke2014doing}). 
For Bayesian modeling, we used Stan software with \textit{pystan} python package.
To do the model robust to outliers, we can consider Bitcoin price distributed with Student's t-distribution. 
Student's t-distribution is similar to normal distribution but it has heavier tails. 
Let us consider that the Bitcoin price is a stochastic variable which is distributed by Student's t-distribution
$$
price'_s \sim Student\_t(\mu,\sigma,\nu),
$$
where $\nu$ denote the degree of freedom, $\mu$ is a location parameter, $\sigma$ is a scale parameter. In Bayesian approach, we consider that 
$\mu=price'$, where $price'$ is defined by the regression model (\ref{e1}). 
Fig.~\ref{boxplot1} shows the box plots for received parameters of regression model.  Fig.~\ref{boxplot1corr} shows the box plots for the parameters in case of added expert correction term.
 For  scale parameter  $\sigma$, the value 0.14 was received in case of the absence of expert correction term and the value 0.1 was received in  case of added expert correction term. It shows that 
adding expert correction term into Bayesian regression model decreases the width of the probability distribution function for the target variable.

\subsection{Conclusions}
In our study, we considered the linear model for Bitcoin price which includes regression features based on Bitcoin currency statistics, 
mining processes, Google search trends, Wikipedia pages visits.
The pattern  of deviation of regression model prediction  from real prices is simpler comparing to price time series. So, we assume
that this pattern can be predicted by an experienced expert. In such a way, combining the regression model and 
expert correction, one can receive better results than with either regression model or expert opinion only.  
We assume that the expert can catch the complex time series patterns that relies on financial behavioral 
theories, economics and politics. These patterns cannot be caught from time series historical data since they exist for a short  time period. 
After these patterns are published or considered between investors, they soon become self-destructed. 
The received results show that the correct expert definition of time pivot points for the regression model deviation can essentially improve the prediction of Bitcoin price.  
In the proposed approach, the expert has to define time pivot points which describe the deviation of regression model based on the historical data comparing to real price time series.
With Bayesian inference, one can utilize the probabilistic approach using distributions with fat tails and take into account outliers in Bitcoin 
price time series. Having probability density distributions for price 

\section{Modeling COVID-19 Spread and Its Impact on Stock Market Using Different Types of Data}
In this case study,   we consider  different regression approaches for modeling COVID-19 spread  and its impact on the stock market. The logistic curve model was used with Bayesian regression for predictive analytics of the COVID-19 spread. Bayesian approach makes it possible to use informative prior distributions formed by experts that allows considering  to consider the results as a compromise between historical data and expert opinion. 
	The obtained results show that different crises with different reasons have different impact on the same stocks.  Bayesian inference makes it possible to analyze the uncertainty of crisis impacts. 
The impact of COVID-19 on the stock market using time series of visits on Wikipedia pages 
related to coronavirus was studied. Regression approach for modeling COVID-19 crises and other crises impact on stock market were investigated.
The analysis of semantic structure of tweets  related to coronavirus using  frequent itemsets and association rules  theory  was carried out. 

\subsection{Introduction}
At present time, there are different methods, approaches and data sets  for modeling  the COVID-19 
spread~\cite{covid_ref2,covid_ref3,covid_ref4,covid_ref5,covid_ref6,covid_ref7}.
The approach based on Bayesian inference allows us to receive a posterior distribution of model parameters using conditional likelihood and prior distribution. 
  In the Bayesian inference, we can use informative prior distributions which can be set up by an expert. So, the result can be considered as a compromise between 
historical data and expert opinion. It is important in the cases when we have a small number of historical data. 
 In ~\cite{  
pavlyshenko2019bitcoin,pavlyshenko2016linear,pavlyshenko2016machine,pavlyshenko2020using} we consider 
 different approaches of using  Bayesian regression.
For solving Bayesian models, numerical Monte-Carlo methods are used. Gibbs and Hamiltonian sampling are the popular methods of finding posterior distributions for the parameters of probabilistic mode
~\cite{kruschke2014doing,gelman2013bayesian, carpenter2017stan}.
Bayesian inference makes it possible to obtain probability density functions for model parameters and estimate the uncertainty that is important in risk assessment analytics.
Different problems caused by COVID-19 are being widely considered in social networks especially Twitter.   
So, the analysis of tweet trends  can reveal the semantic structure of users' opinions related to COVID-19. 

In this work, we consider the use of  Bayesian regression for modeling the COVID-19 spread. 
We also consider the impact of COVID-19 on the stock market using time series of visits on Wikipedia pages 
related to coronavirus. We study a regression approach for modeling the impact of different crises  on the stock market.
The usage of frequent itemsets and  association rules theory for analysing tweet sets was considered. 

\subsection{Bayesian Model for COVID-19 Spread Prediction}
\begin{figure}[htb]
\center
\includegraphics[width=0.7\linewidth]{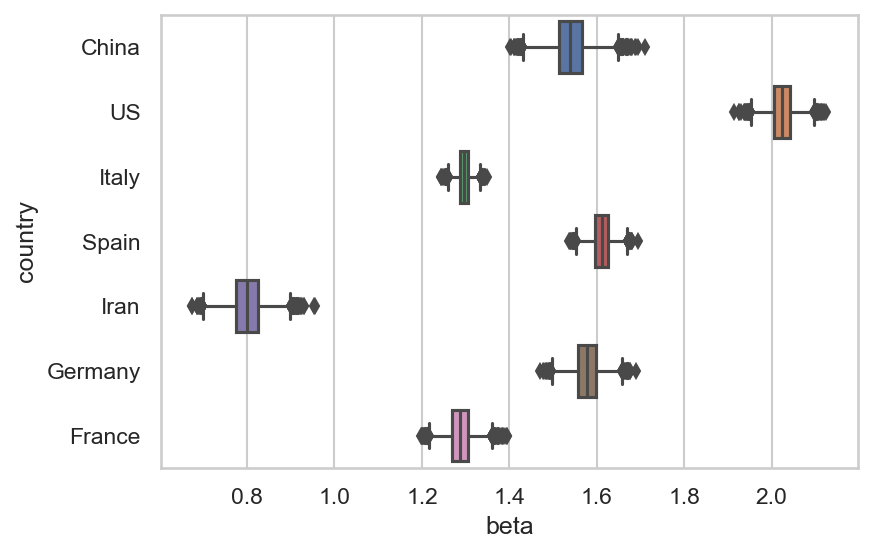}
\caption{Box plots for beta coefficients of coronavirus spread model for different countries}
\label{covid_beta_boxplots_countries}
\end{figure}

For the predictive analytics of the COVID-19 spread, we used a logistic curve model. 
Such model is very popular nowadays. 
To estimate model parameters, we  used Bayesian regression~\cite{kruschke2014doing,gelman2013bayesian, carpenter2017stan}. 
A logistic curve model with Bayesian regression approach can be written as follows:
 \begin{equation}
 \begin{split}
 &n \sim \mathcal{N}(\mu, \sigma) \\
 &\mu=\frac{\alpha}{1+exp(-\beta(t-t_0))}10^5 \\
 &t=t_{weeks}(Date-Date_0)
 \end{split}
\end{equation}
where $Date_0$ is a start day for observations in the historical data set, it is measured in weeks,
 $\alpha$ parameter denotes maximum cases of coronavirus, $\beta$ parameter is an empirical coefficient which denotes the rate of coronavirus spread. The data for the analysis were taken from ~\cite{covid_ref3}.
 For Bayesian inference calculations, we used a 'pystan' package for Stan platform for statistical modeling \cite{carpenter2017stan}.
Figure ~\ref{covid_beta_boxplots_countries} shows the box plots for calculated $\beta$ parameters of the coronavirus spread model for different countries.

\begin{figure}[htb]
\center
\includegraphics[width=0.85\linewidth]{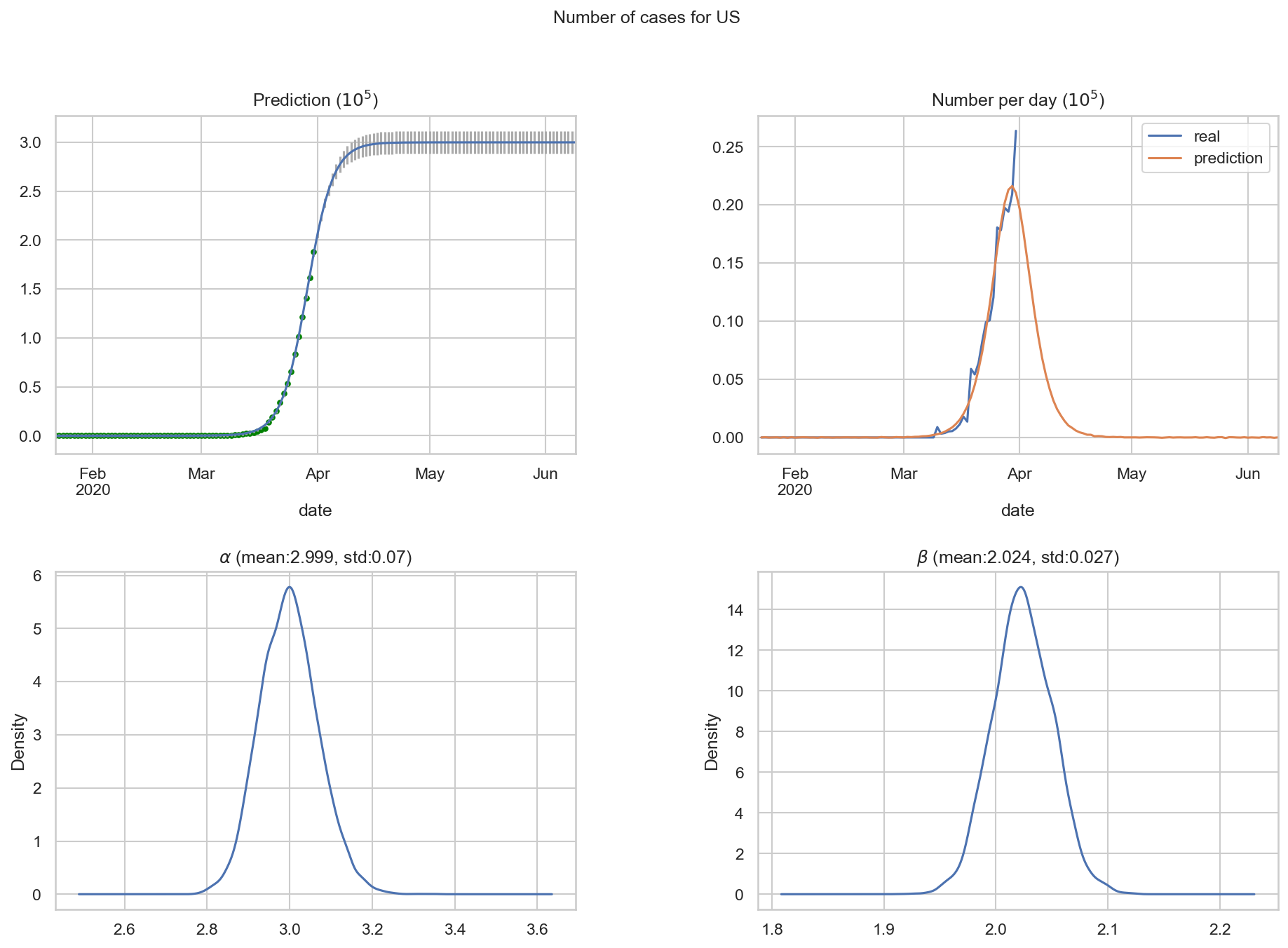}
\caption{Modeling of COVID-19 spread  for USA.}
\label{covid_fig3}
\end{figure}

\begin{figure}[htb]
\center
\includegraphics[width=0.85\linewidth]{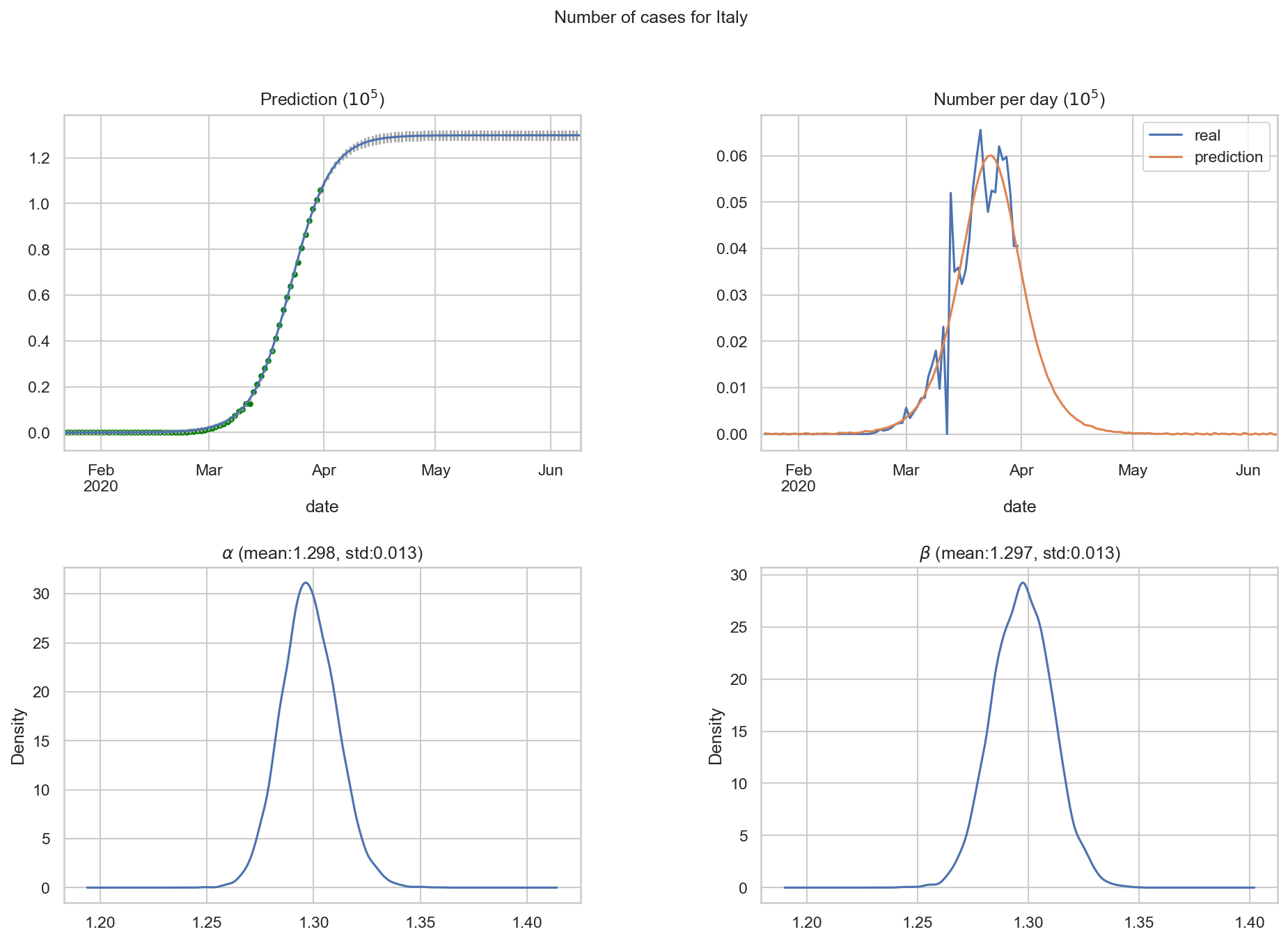}
\caption{Modeling of COVID-19 spread  for Italy.}
\label{covid_fig4}
\end{figure}

Figures~\ref{covid_fig3},~\ref{covid_fig4} show the predictions for  coronavirus spread cases  using current historical data. In  practical analytics, it is important to find the maximum of coronavirus cases per day, this point means the estimated half time of coronavirus spread in the region under investigation. New historical data will correct the distributions for model parameters and forecasting results. 

The results show that the Bayesian regression model using logistic curve can be effectively used for predictive analytics of the  COVID-19 spread.  In Bayesian regression approach, we can take into account expert opinions via information prior distribution, so the results can be treated as a compromise between historical data and expert opinion that  is important in the case of small amount of historical data or in the case of a non-stationary process.
It is important to mention that new data and expert prior distribution for model can essentially correct previously received results. 
\FloatBarrier
\subsection{Using Wikipedia Pages Visits Time Series for Prediction Stock Market Movements}
\begin{figure}[htb]
\center
\includegraphics[width=0.85\textwidth]{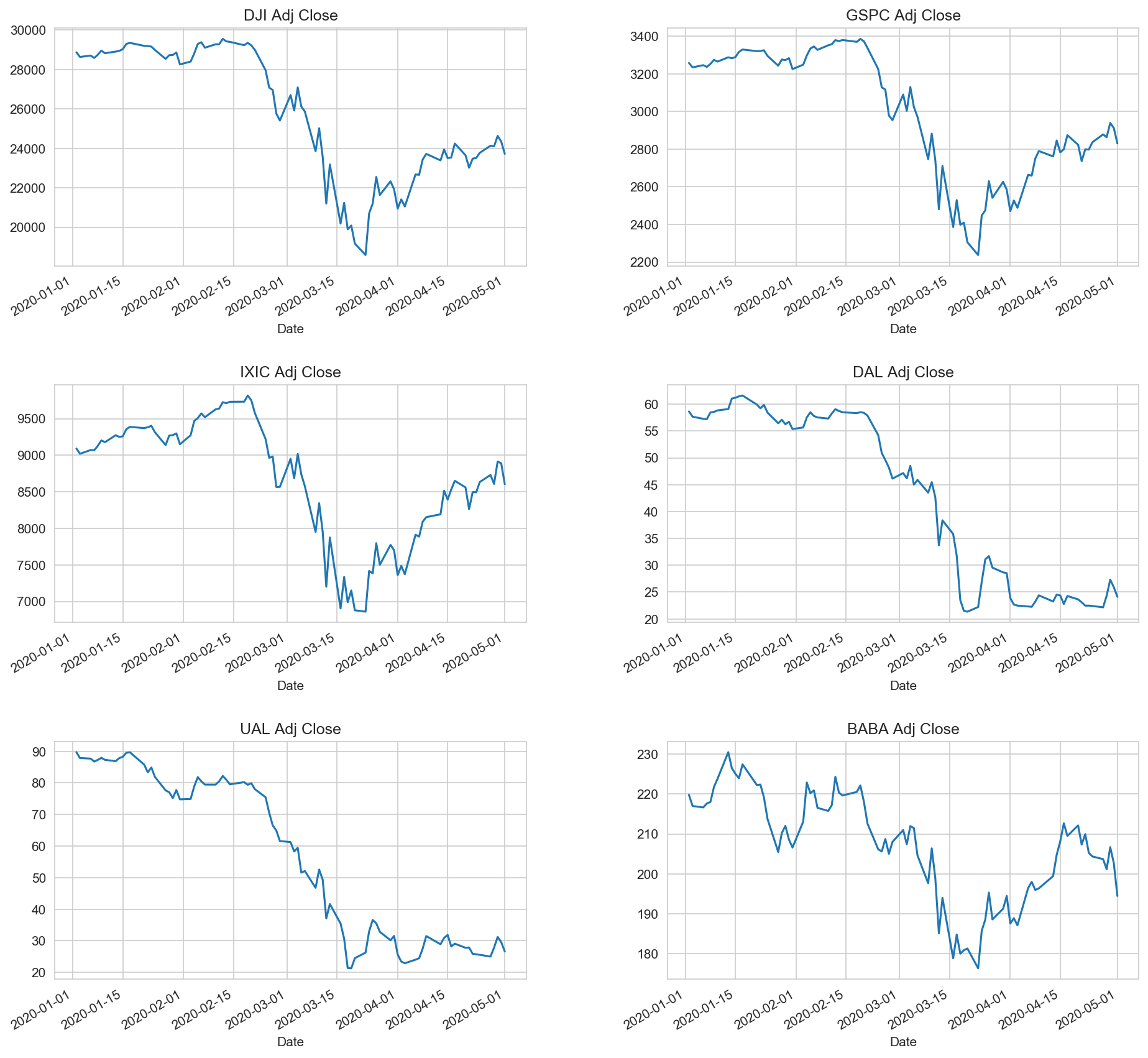}
\caption{Stock market indexes and  stock price  time series.}
\label{fig_covid_ts_stock_prices}
\end{figure}

\begin{figure}[htb]
\center
\includegraphics[width=0.85\textwidth]{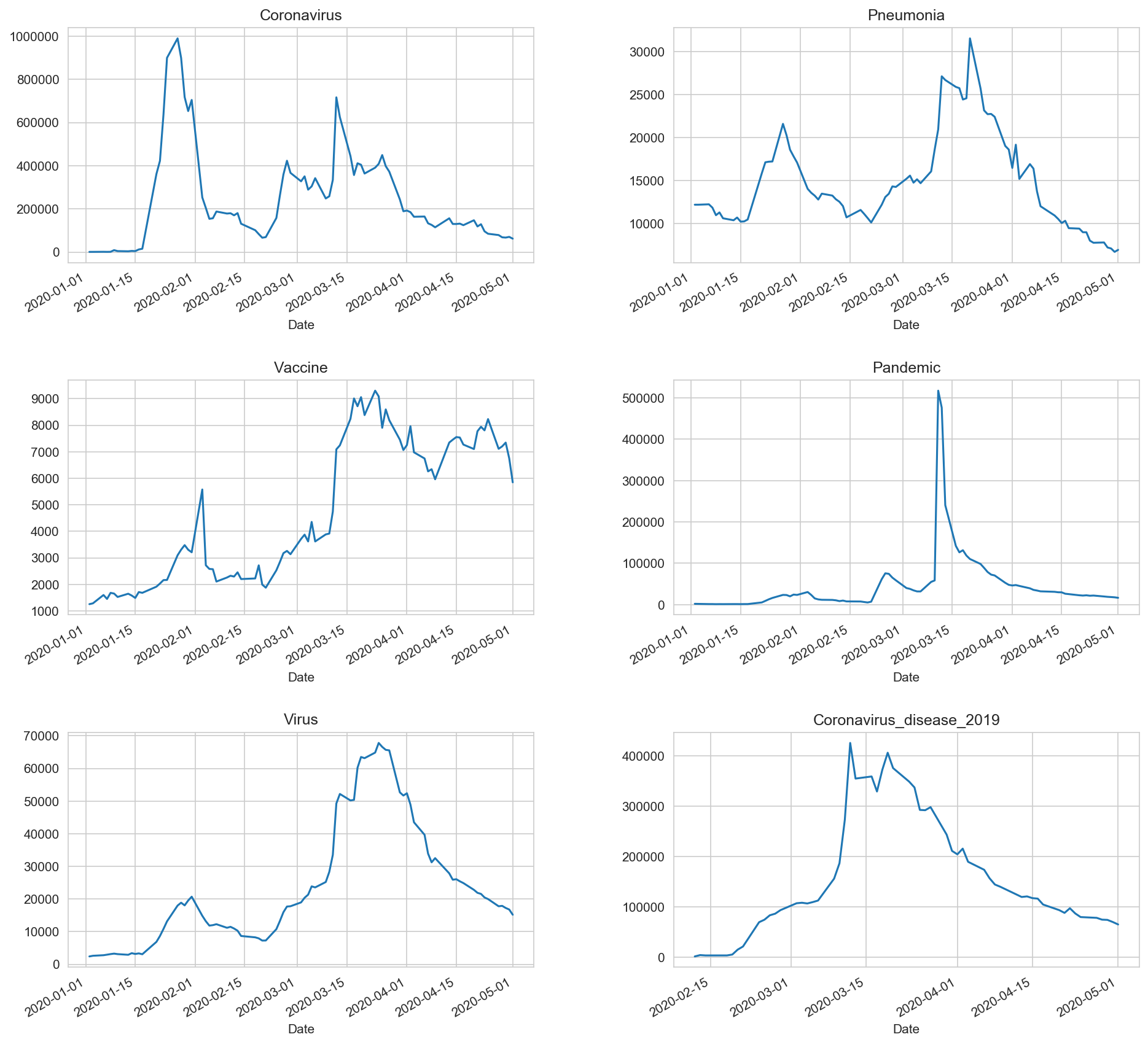}
\caption{Time series of numbers of visits to Wikipedia pages.}
\label{fig_covid_wiki_ts}
\end{figure}

\begin{figure}[htb]
\center
\includegraphics[width=0.85\textwidth]{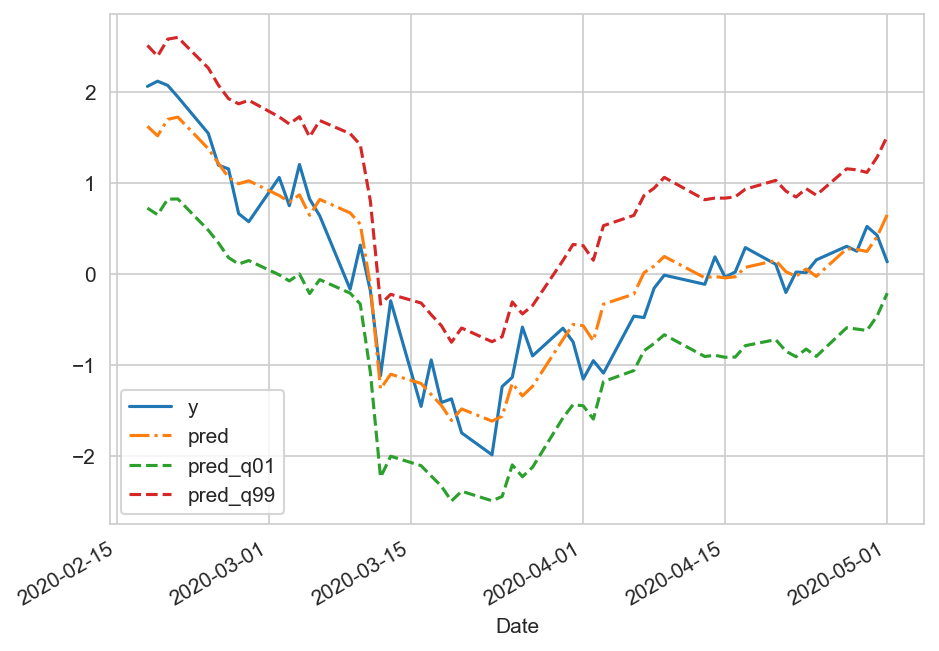}
\caption{Mean values and 0.01, 0.99 quantiles of probability density function of predictions for  S\&P-500 index.}
\label{fig_covid_sp500_pred}
\end{figure}

\begin{figure}[htb]
\center
\includegraphics[width=0.75\textwidth]{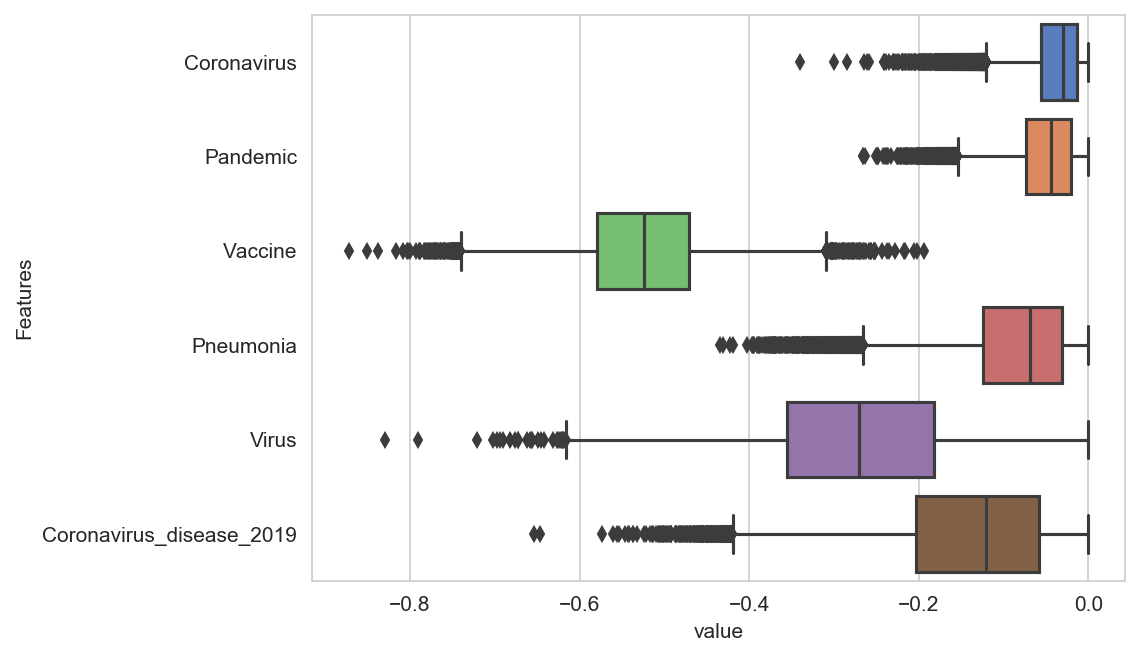}
\caption{Boxplots for PDF of linear regression coefficients.}
\label{fig_covid_boxplots_coef_wp}
\end{figure}

\begin{figure}[htb]
\center
\includegraphics[width=0.75\textwidth]{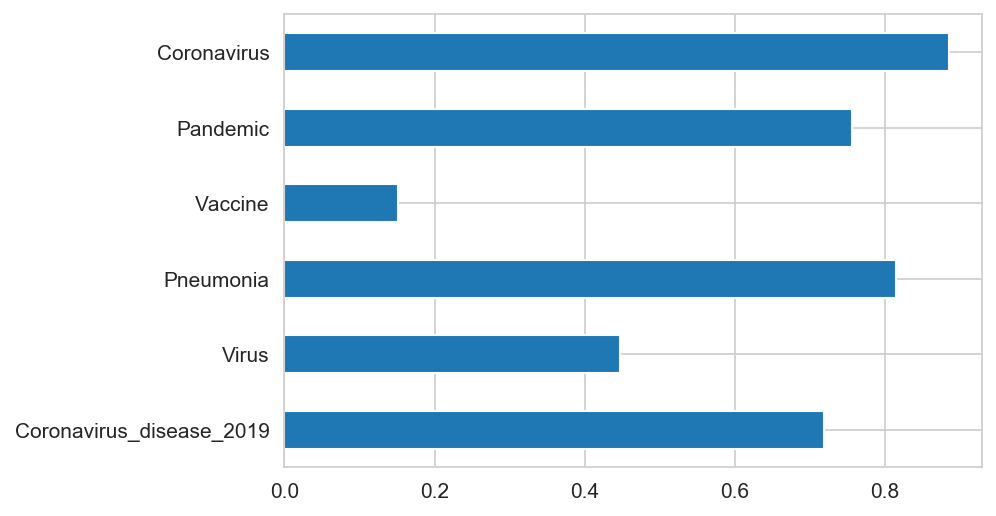}
\caption{Variation coefficient  for features.}
\label{fig_coef_std_mean_wp}
\end{figure}

Let us consider the influence of the COVID-19 spread on the stock market movement. 
The influence of impact factors can be described by alternative data, such as 
characteristics of search trends, users' activity in social networks, etc.
 Figure~\ref{fig_covid_ts_stock_prices} shows the stock market indexes and some stock price time series 
in the period of the biggest  impact of COVID-19 on the stock market. 
In this study, as alternative data, we consider the time series of visits to  Wikipedia pages which are related to  COVID-19.
The figure~\ref{fig_covid_wiki_ts} shows the time series of numbers of visits to Wikipedia pages. 
For our analysis, we consider the time period of ['2020-02-15','2020-05-01']. 
For Bayesian regression, we used Stan platform for statistical modeling~\cite{carpenter2017stan}. 
As features, we used z-scores of the time series of Wikipedia page visit numbers.
As a target variable, we used z-scores of S\&P-500 index. 
We applied the constraints that linear regression coefficients cannot be positive. 
Figure ~\ref{fig_covid_sp500_pred} shows the mean values and 0.01, 0.99 quantiles of probability density function of predictions for  S\&P-500 index. 
Figure ~\ref{fig_covid_boxplots_coef_wp} shows the boxplots for PDF of linear regression coefficients.
Figure ~\ref{fig_coef_std_mean_wp} shows variation coefficient which is equal to the ratio between the standard deviation and absolute mean values  of regression coefficients. 
These coefficients  describe the  uncertainty of regression features. 
The obtained results show that different  features have different 
impact and uncertainty with respect to the target variable.
The most impactful and the least volatile among the considered features was the feature of the number of visits to the Wikipedia page about the vaccine.

\FloatBarrier
\subsection{Regression Approach for Modeling the  Impact of Different Crises  on the Stock Market}
\begin{figure}[htb]
\center
\includegraphics[width=0.75\linewidth]{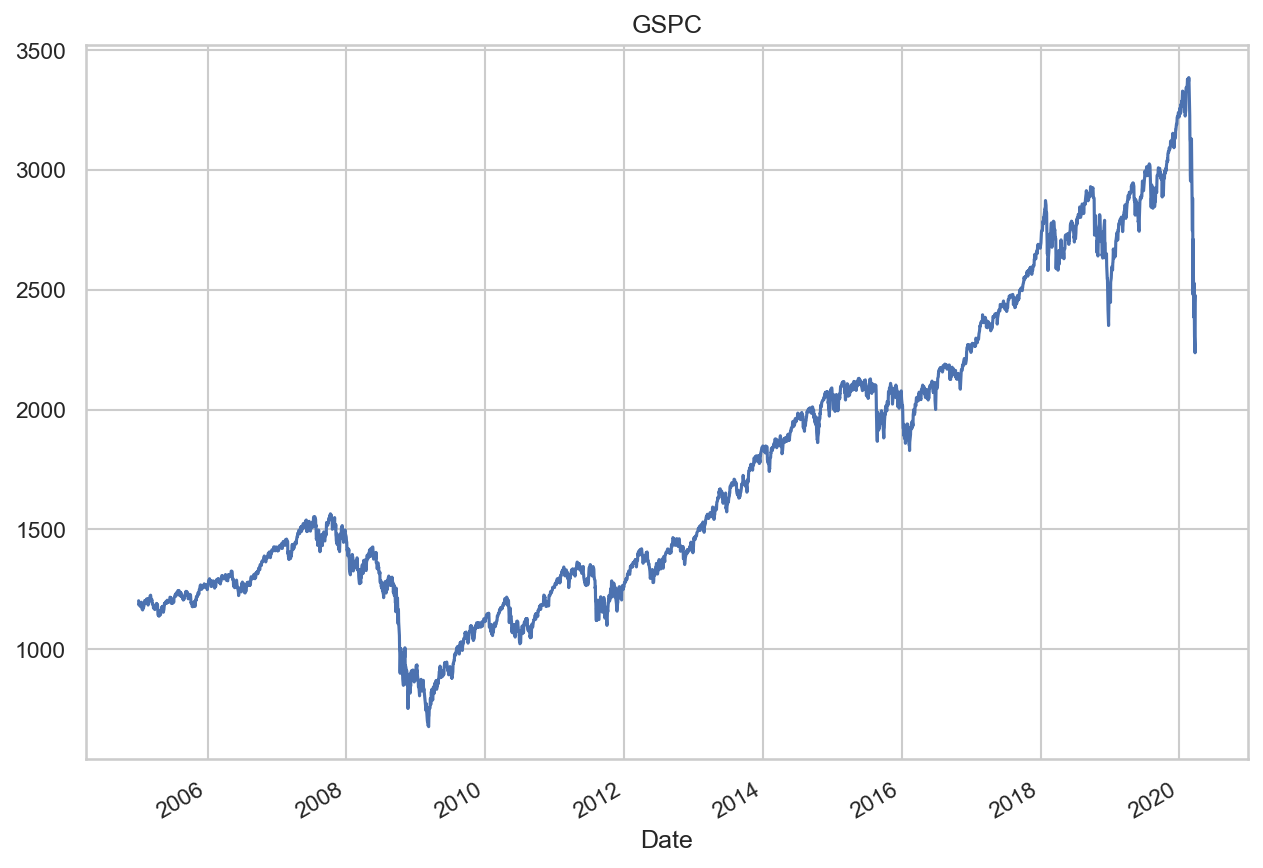}
\caption{Time series for S\&P500 composite index.}
\label{covid_fig10}
\end{figure}

	The coronavirus outbreak has a huge impact on the stock market. It is very important, e.g. for forming stable portfolios, to understand how different crises impact stock prices and the stock market as a whole. 
	We are going to consider the impact of coronavirus crisis on the stock market and compare it to the crisis of 2008 and market downturn of 2018. For this, we can use the regression approach using ordinary least squares (OLS)  regression and Bayesian regression. Bayesian inference makes it possible to obtain probability density functions for coefficients of the factors under investigation and estimate the uncertainty that is important in the risk assessment analytics. In Bayesian regression approach,  we can analyze extreme target variable values using non-Gaussian distributions with fat tails.
	We took the following time periods for each of crises -  crisis\_2008: [2008-01-01,2009-01-31], down\_turn\_2018: [2018-10-01,2019-01-03], coronavirus: [2020-02-18,2020-03-25].  For each of the above mentioned crises, we created a regression variable which is equal to 1 in the crisis time period and 0 in other cases.
	Figure~\ref{covid_fig10} shows the time series for S\&P500 composite index.

\begin{figure}[htb]
\center
\includegraphics[width=0.5\linewidth]{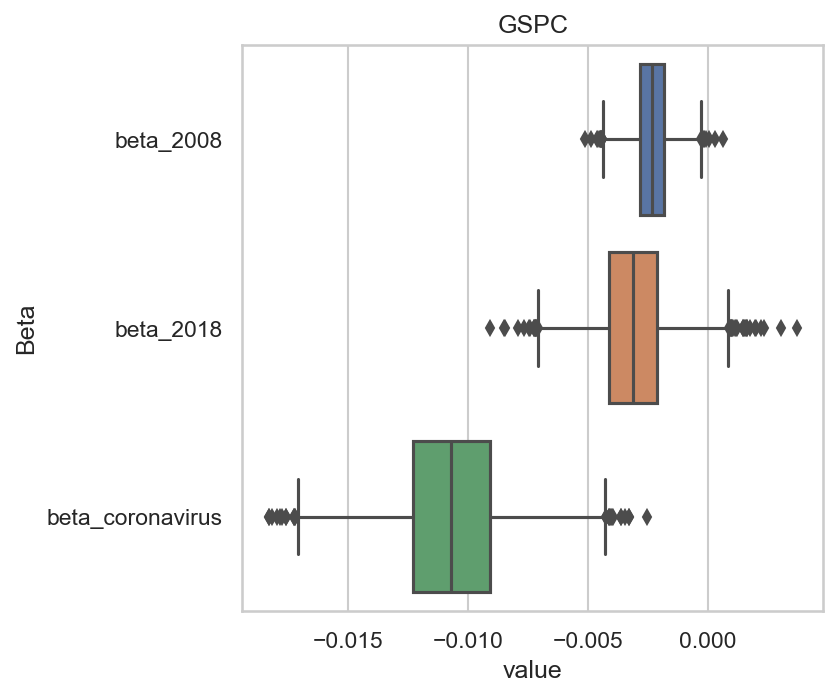}
\caption{Box plots of impact weights of each crisis on  S\&P composite index.}
\label{covid_fig11}
\end{figure}

\begin{figure}[htb]
\center
\includegraphics[width=0.5\linewidth]{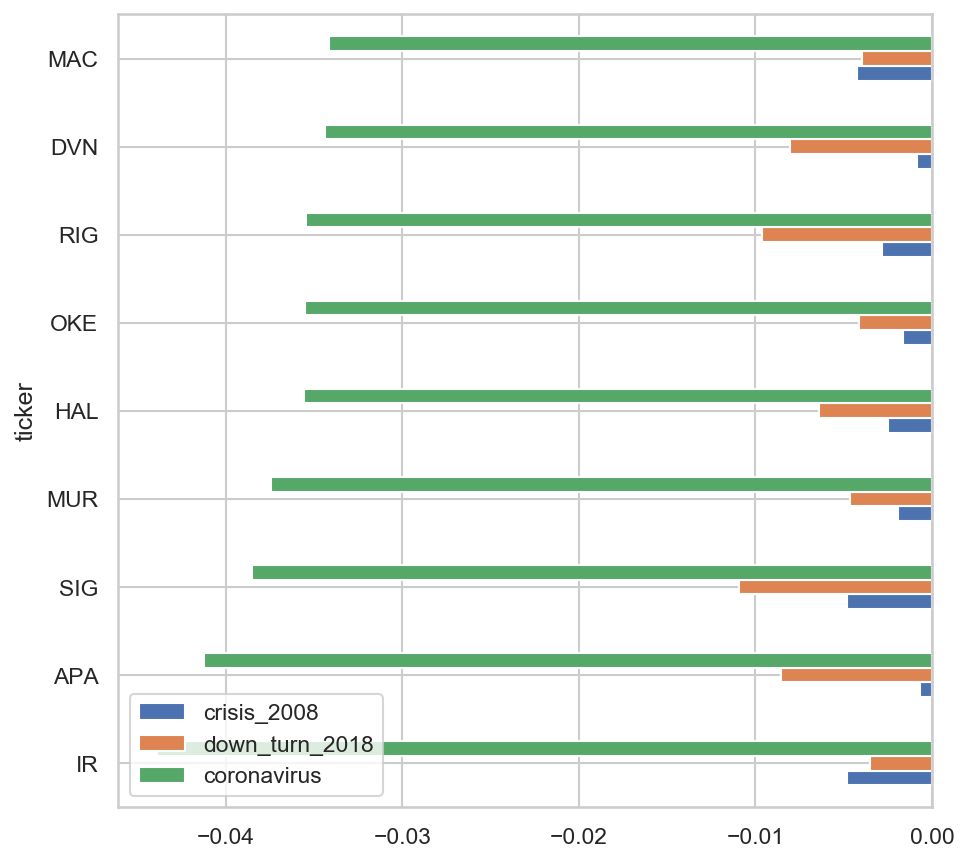}
\caption{Top negative price returns in COVID-19 crises.}
\label{covid_fig12}
\end{figure}

\begin{figure}[htb]
\center
\includegraphics[width=0.5\linewidth]{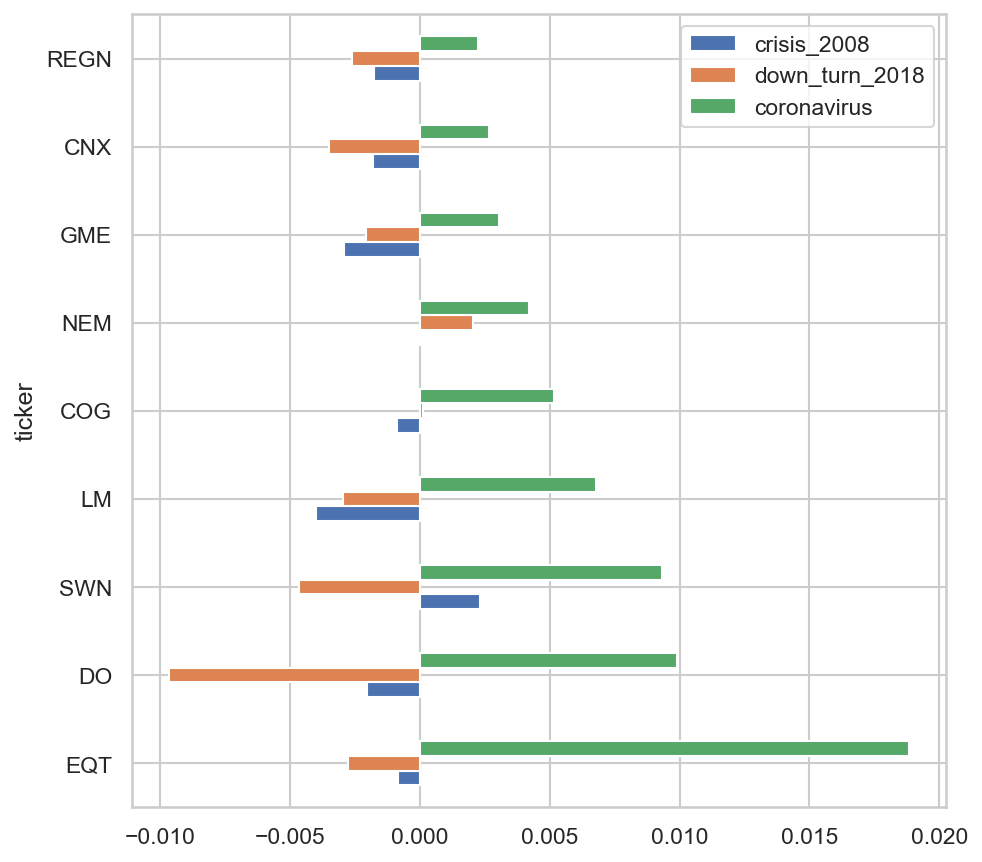}
\caption{Tickers with positive price returns in COVID-19 crises.}
\label{covid_fig13}
\end{figure}

\begin{figure}[htb]
\center
\includegraphics[width=0.5\linewidth]{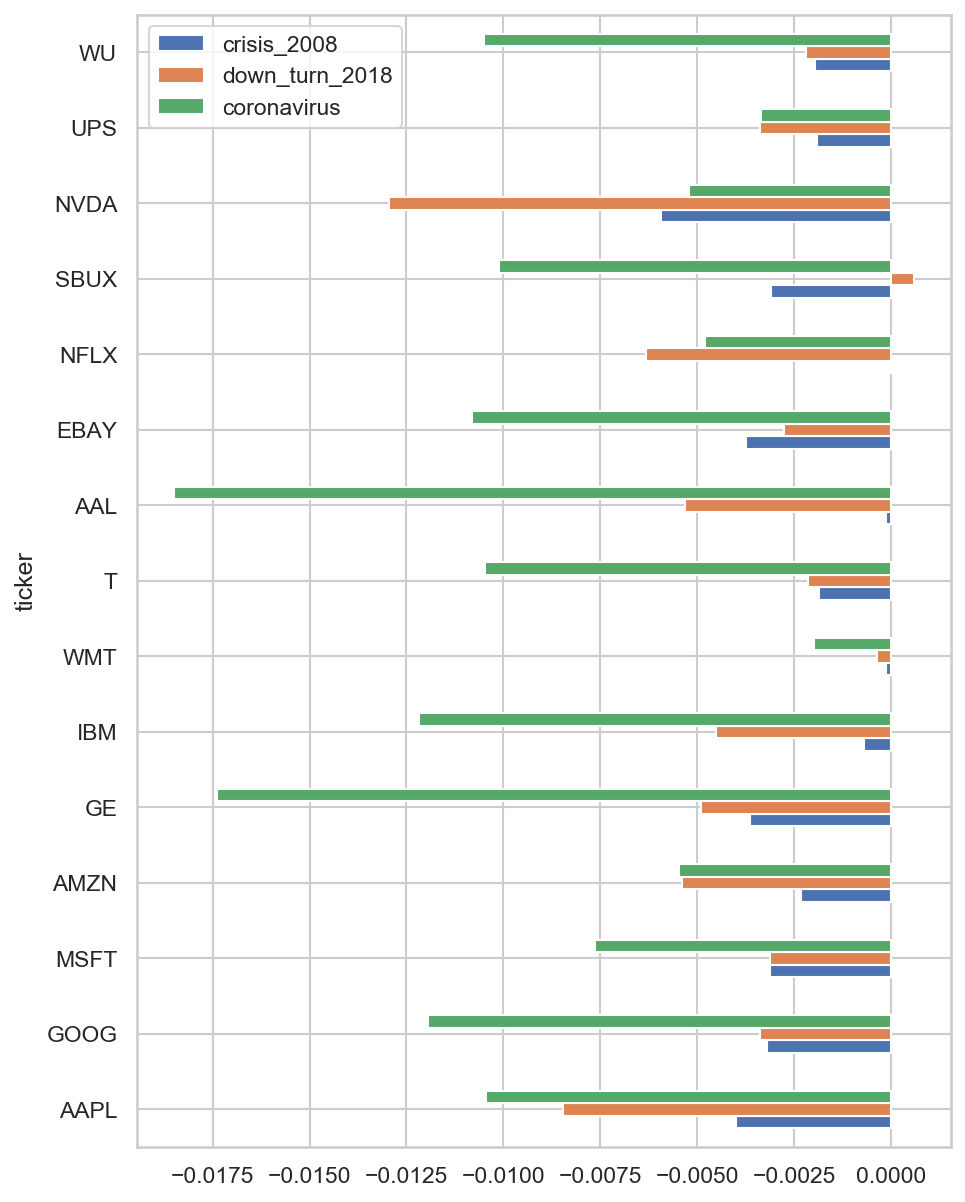}
\caption{The weights for different crises for arbitrarily chosen stocks.}
\label{covid_fig14}
\end{figure}

\begin{figure}[htb]
\center
\includegraphics[width=0.85\linewidth]{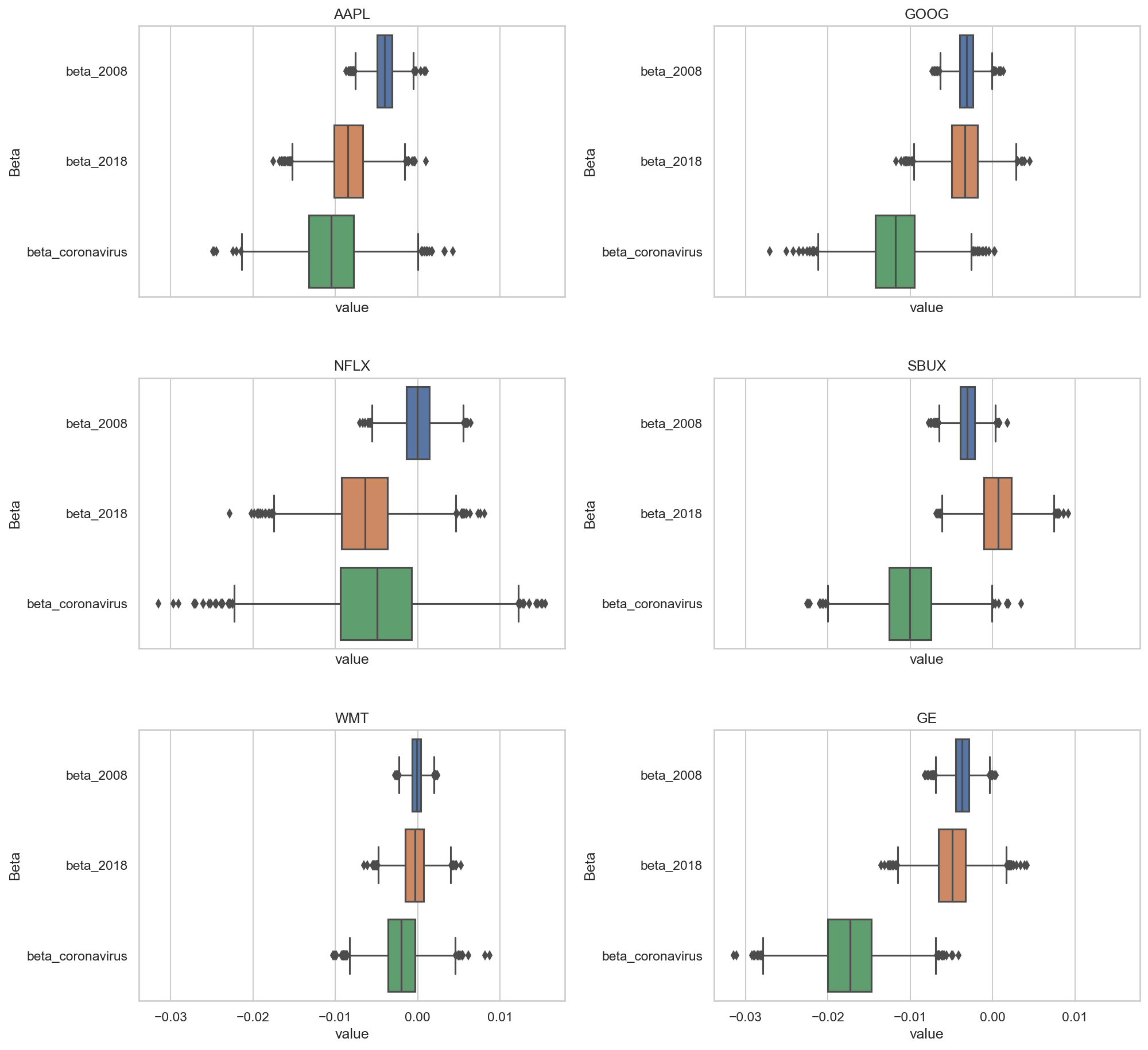}
\caption{Box plots for crisis weights for different stocks.}
\label{covid_fig15}
\end{figure}

	As a target variable, we consider the daily price return. Knowing the daily price return, changes in crises periods, one can estimate the ability of investors to understand trends and recalculate portfolios.
	These results were received using Bayesian inference. For Bayesian inference calculations, we used Python \textit{'pystan'} package for \textit{Stan} platform for statistical modeling \cite{carpenter2017stan}.
	Figure~\ref{covid_fig11} shows the box plots of impact weights of each crisis on  S\&P500 composite index. 
	The wider box for coronavirus weight can be caused by shorter time period of investigation comparing with other crises and consequently larger uncertainty.
	For our investigations, we took a random set of tickers from S\&P list.
	Figure~\ref{covid_fig12} shows top negative price returns in coronavirus crises.
	Figure~\ref{covid_fig13} shows the tickers with positive price return in coronavirus crisis. 
	Figure~\ref{covid_fig14} shows the weights for different crises for arbitrarily chosen stocks. 
	We calculated the distributions for crises weights using Bayesian inference.
	Figure~\ref{covid_fig15} shows the box plots for crisis weights for different stocks.

\FloatBarrier

 \FloatBarrier
\subsection{Conclusions}
The received results show that 
the logistic curve model can be used with Bayesian regression for the predictive analytics of the COVID-19 spread. 
Such a model can be effective when  the exponential growth  of coronavirus confirmed cases 
takes place.
In practical analytics, it is important to find the maximum of coronavirus cases per day, this point means  the estimated half time of coronavirus spread in the region under investigation. New historical data will correct the distributions for model parameters and forecasting results. 
For conducting the modeling of COVID-19, we developed the ’Bayesian Model for COVID-19 spread Prediction’ Python package, which can be loaded at~\cite{covid_ref8} for free use.
 In Bayesian regression approach, we can take into account expert opinions via information prior distribution, so the results can be treated as a compromise between the historical data and expert opinion that  is important in the case of small amount of historical data or in the case of a non-stationary process.
It is important to mention that new data and expert prior distribution for model can essentially correct previously received results. The impact of COVID-19 on the stock market using time series of visits on Wikipedia pages related to coronavirus was studied. 
The obtained results show that different  features have different 
impact and uncertainty with respect to the target variable.
The most impactful and the least volatile among the considered features was the feature of the number of visits to the Wikipedia page about the vaccine.
The obtained results show that different crises with different reasons have different impact on the same stocks.  Bayesian inference makes it possible to analyze the uncertainty of crisis impacts.
	The uncertainty of crisis impact weights can be measured as a standard deviation for weight probability density functions.
	The uncertainty of coronavirus crisis is larger than other crises that can be caused by shorter analysis time. Knowing the uncertainty, allows risk assessment for portfolios and other financial and business processes.
   Using the graph theory, the users' communities and influencers can be revealed given vertices quantitative characteristics. 
   The analysis of tweets related to COVID-19  was carried out using frequent itemsets and association rules.
   Found frequent itemsets and association rules reveal the semantic structure of tweets related to COVID-19. The quantitative characteristics of frequent itemsets and association rules, e.g. value of support, can be used as features in the predictive analysis. 

\section{Forming Predictive Features of Tweets for  Decision-Making Support}

In this case study,   we consider  the approaches for forming different predictive features of tweet data sets and using them in the predictive analysis for decision-making support. 
The graph theory as well as frequent itemsets and association rules theory is used for  forming and retrieving different features from these datasests.  The use of  these approaches makes it possible to reveal a semantic structure in tweets related to a specified entity. 
 It is shown that quantitative characteristics of semantic frequent itemsets can be used in predictive regression models with specified target variables. 
 
\subsection{Introduction}

Tweets, the messages of Twitter microblogs,  have high density of semantically important keywords. It makes it possible to get semantically important information from the tweets and generate the features of predictive models for the decision-making support. Different studies of Twitter are considered in the papers~\cite{java2007we,kwak2010twitter,pak2010twitter, cha2010measuring,benevenuto2009characterizing,
bollen2011twitter,asur2010predicting,shamma2010tweetgeist, kraaijeveld2020predictive,
wang2020novel, balakrishnan2020improving}. 
In~\cite{pavlyshenko2019forecasting,pavlyshenko2019cantwitter}, we study the use of  tweet features for forecasting different kinds of events. 
In~\cite{pavlyshenko2020modelling}, we study the modeling of COVID-19 spread and its impact on the stock market using different types of data as well as  consider the features of tweets related to COVID-19 pandemic.

In this work, we study the predictive features of tweets using loaded datasets of tweets related  to Tesla company. 

\subsection{Graph structure of tweets}
\begin{figure}
\center
\includegraphics[width=0.65\linewidth]{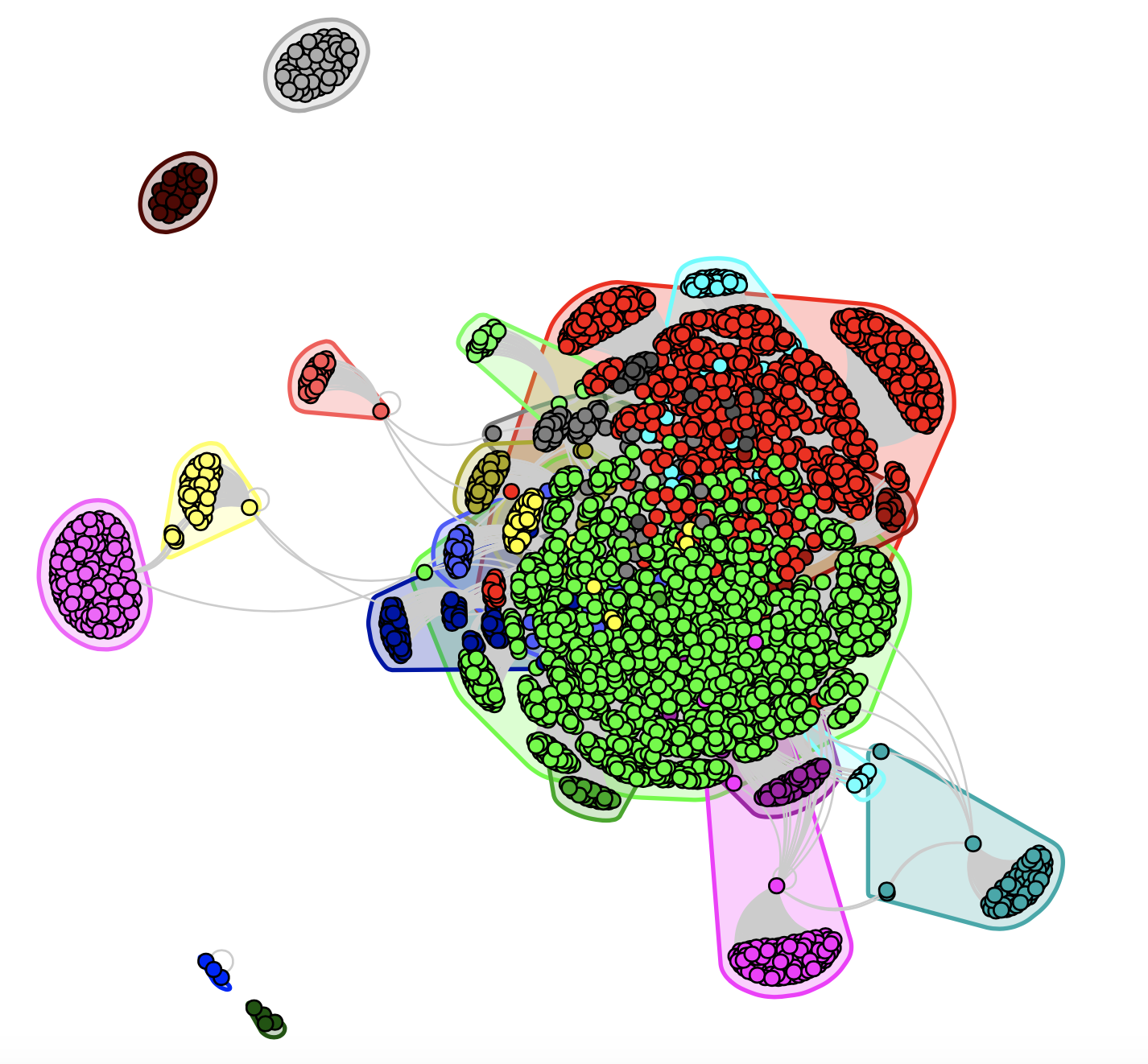}
\caption{Revealed users' communities for the subset of tweets}
\label{usr_graph}
\end{figure}

\begin{figure}
\centerline{\includegraphics[width=0.65\textwidth]{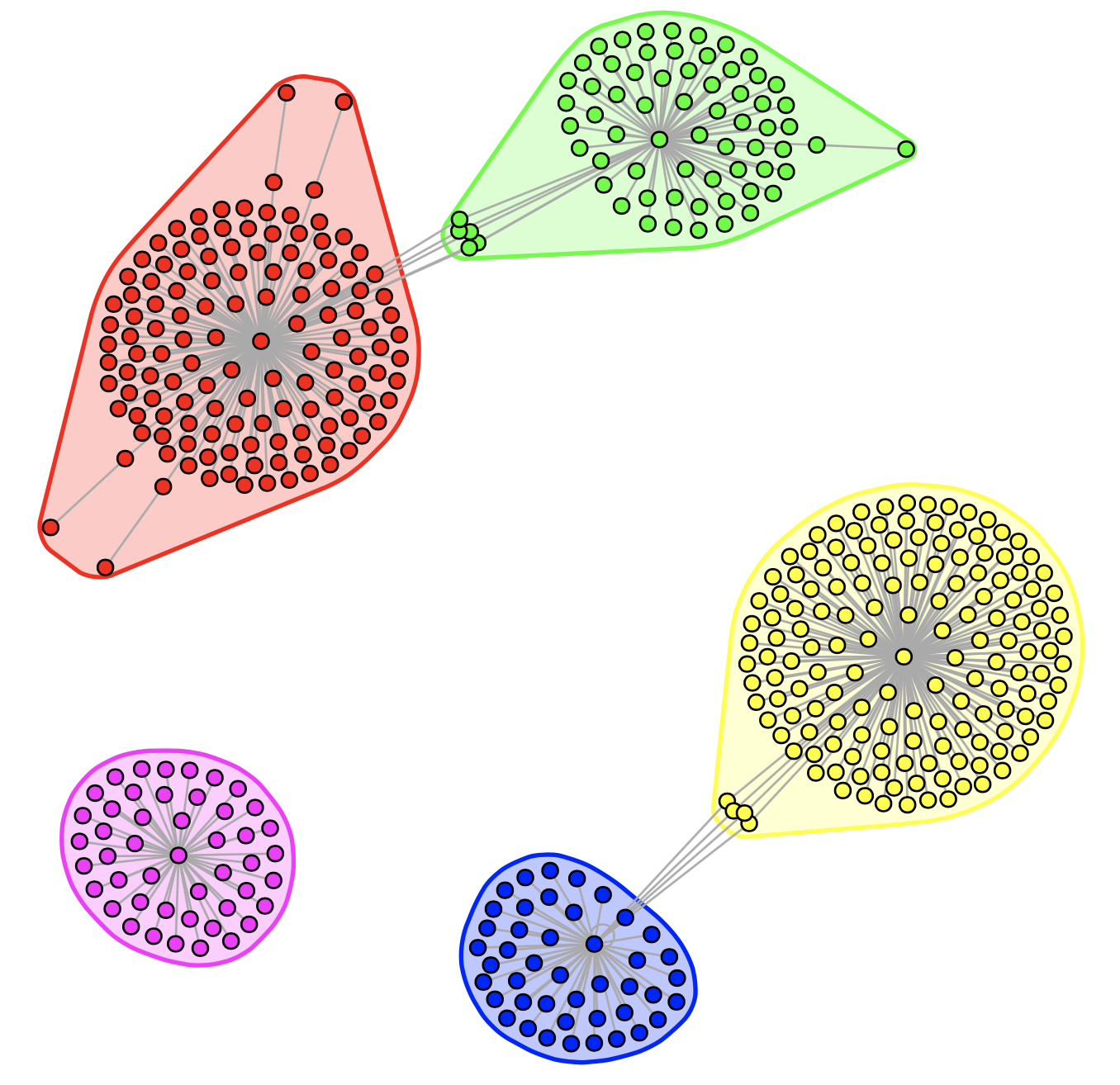}}
\caption{Subgraph for users of highly isolated communities}
\label{tesla_tw_p15}
\end{figure}

The  relationships among users can be considered as a graph, where vertices denote users and edges denote their connections.
Using graph mining algorithms, one can detect user communities and find ordered lists of users by various characteristics, such as
\textit {Hub, Authority, PageRank, Betweenness}. To identify user communities, we used the \textit{Community Walktrap Algorithm} algorithm, which is implemented in the package
 \textit{igraph}~\cite{csardi2006igraph} for the R programming language environment. We used the Fruchterman-Reingold algorithm from this package for visualization.
The \textit{Community Walktrap} algorithm searches for related subgraphs, also called communities, by random walk~\cite{pons2005computing}.
A graph which shows the relationships between users can be represented by
 Fruchterman-Reingold algorithm~\cite{fruchterman1991graph}.
We can assume that tweets could carry predictive information for different business processes. For our case study, we have loaded the tweets related to Tesla company for some time period. Qualitative structure can be used for aggregating different quantitative time series and, in such a way, creating new features for predictive models which can be used, for example, for stock prices forecasting. Let us consider which features we can retrieve from tweet sets for the predictive analytics. Figure~\ref{usr_graph} shows revealed users' communities for the subset of tweets.
Figure~\ref{tesla_tw_p15} shows the subgraph for users of highly isolated communities.
Revealing users' communities makes it possible to analyze different trends in tweet streams which are forming by different users' groups. 
\subsection{Analysis of tweets using frequent itemsets}
\begin{figure}
\center
\includegraphics[width=0.65\linewidth]{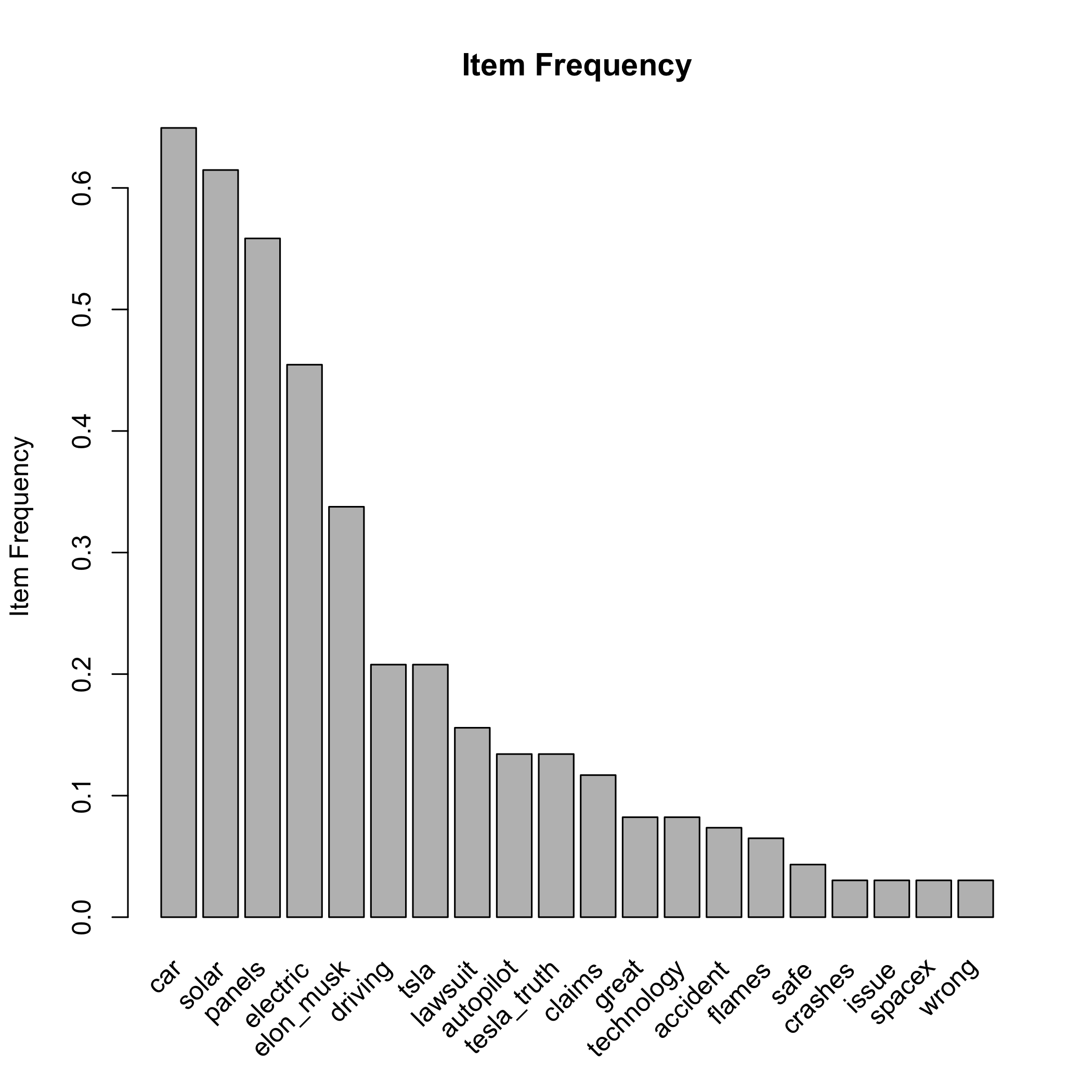}
\caption{Keyword frequencies for  the thematic field of frequent itemset analyis}
\label{tesla_freq}
\end{figure}

\begin{figure}
\centerline{\includegraphics[width=0.75\textwidth]{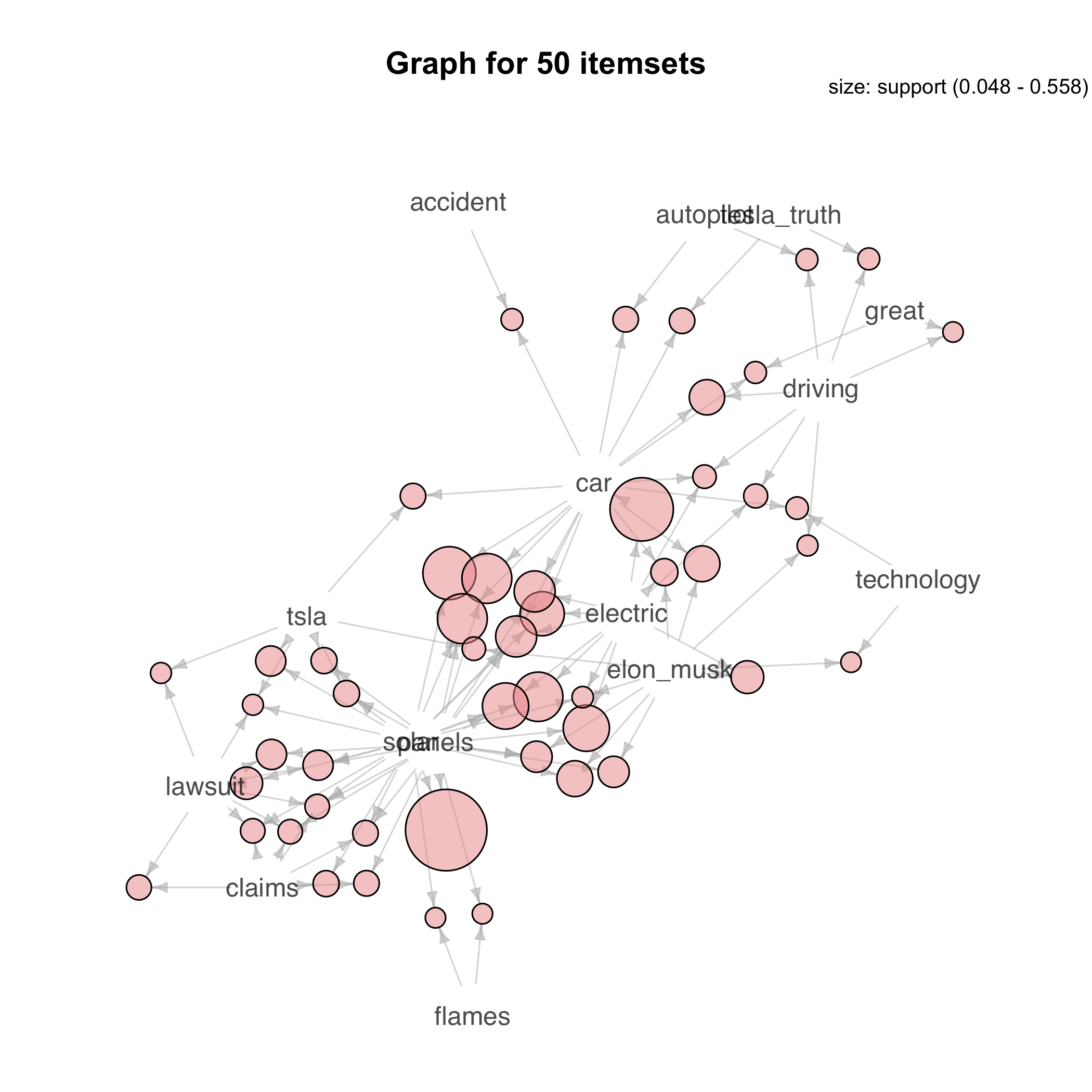}}
\caption{Semantic frequent itemsets}
\label{tesla_tw_p3}
\end{figure}

\begin{figure}
\centerline{\includegraphics[width=0.65\textwidth]{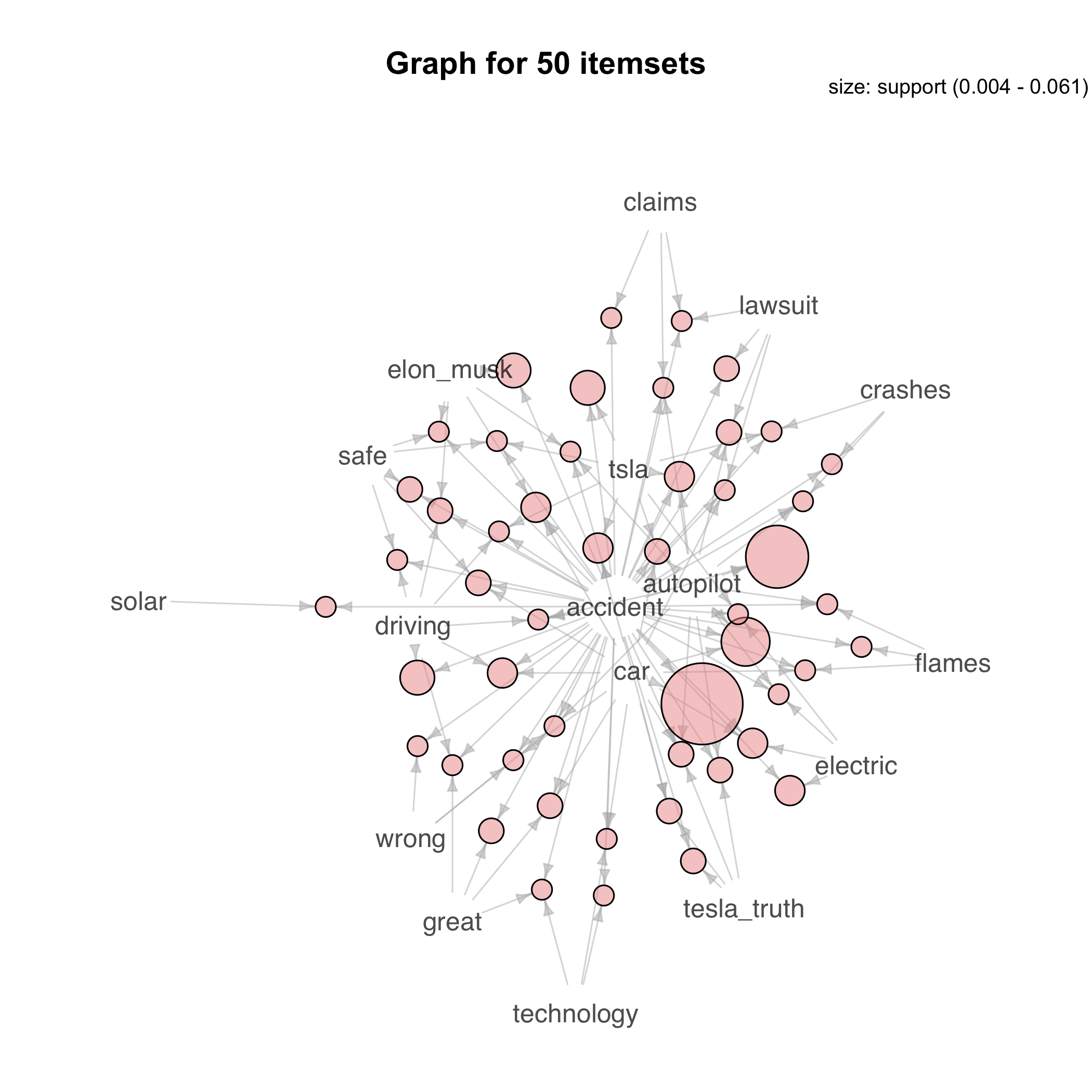}}
\caption{Semantic frequent itemsets}
\label{tesla_tw_p4}
\end{figure}

\begin{figure}
\centerline{\includegraphics[width=0.65\textwidth]{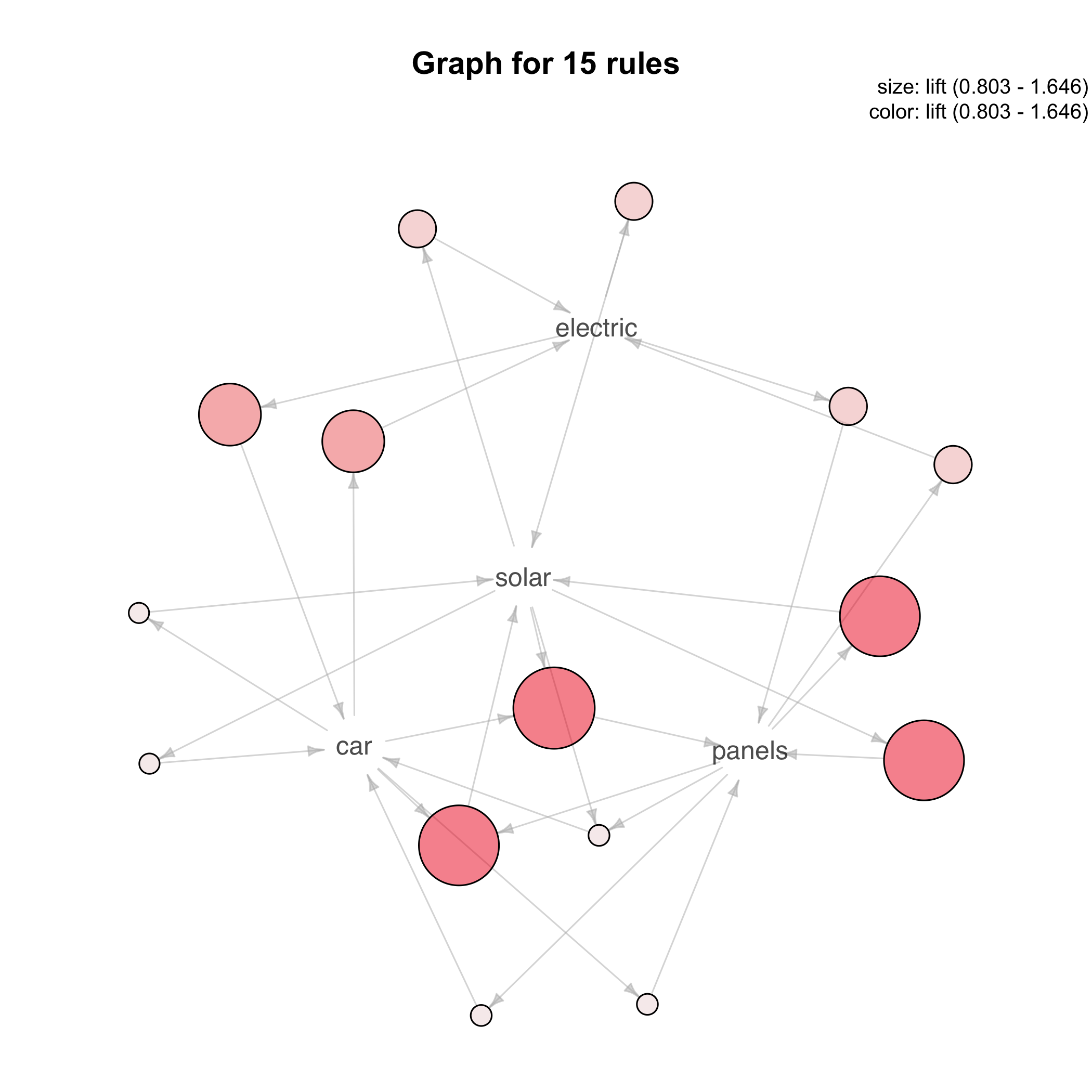}}
\caption{Associative rules, represented by a graph}
\label{tesla_tw_p5}
\end{figure}

\begin{figure}
\centerline{\includegraphics[width=0.65\textwidth]{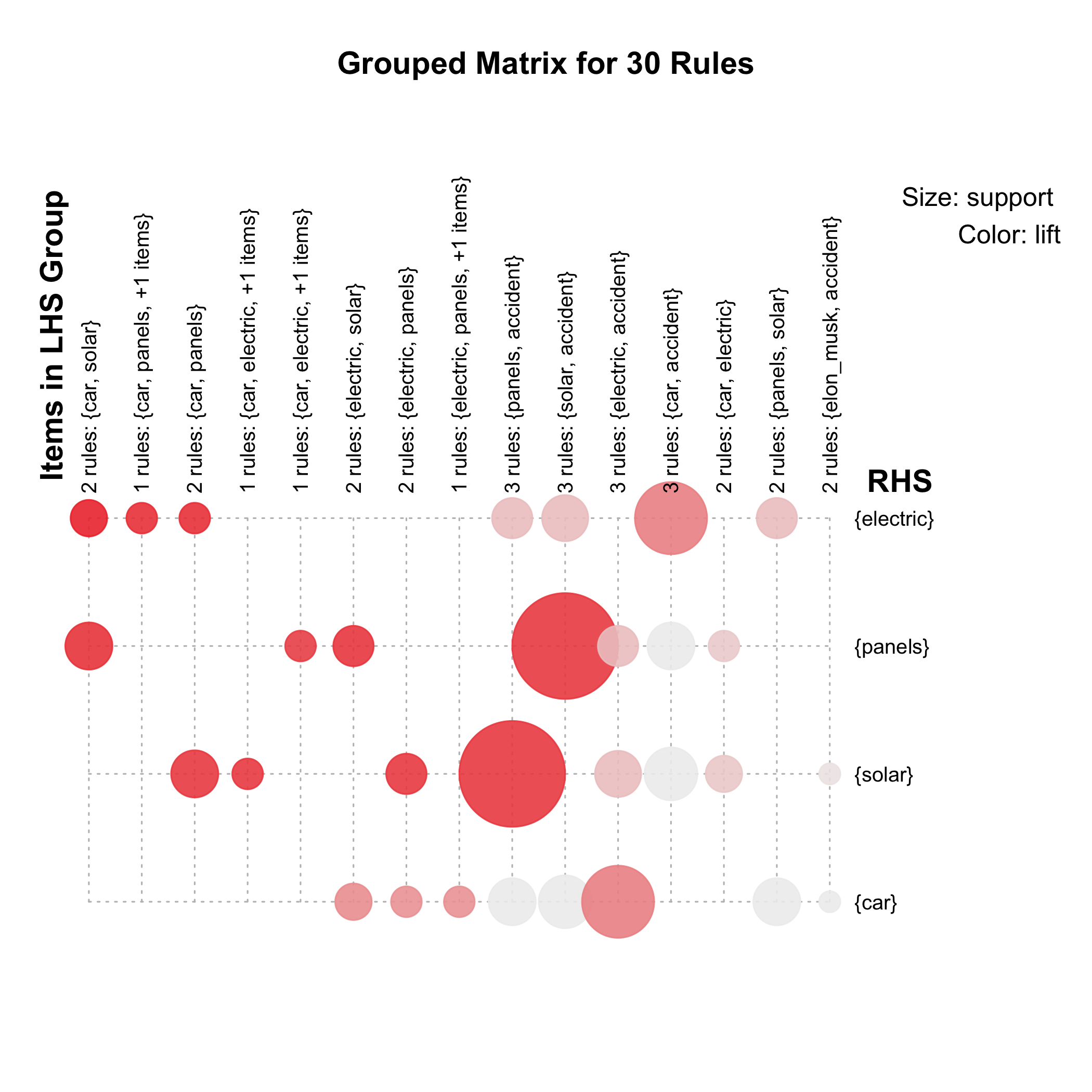}}
\caption{Associative rules represented by a grouped matrix}
\label{tesla_tw_p6}
\end{figure}

\begin{figure}
\center
\includegraphics[width=0.85\linewidth]{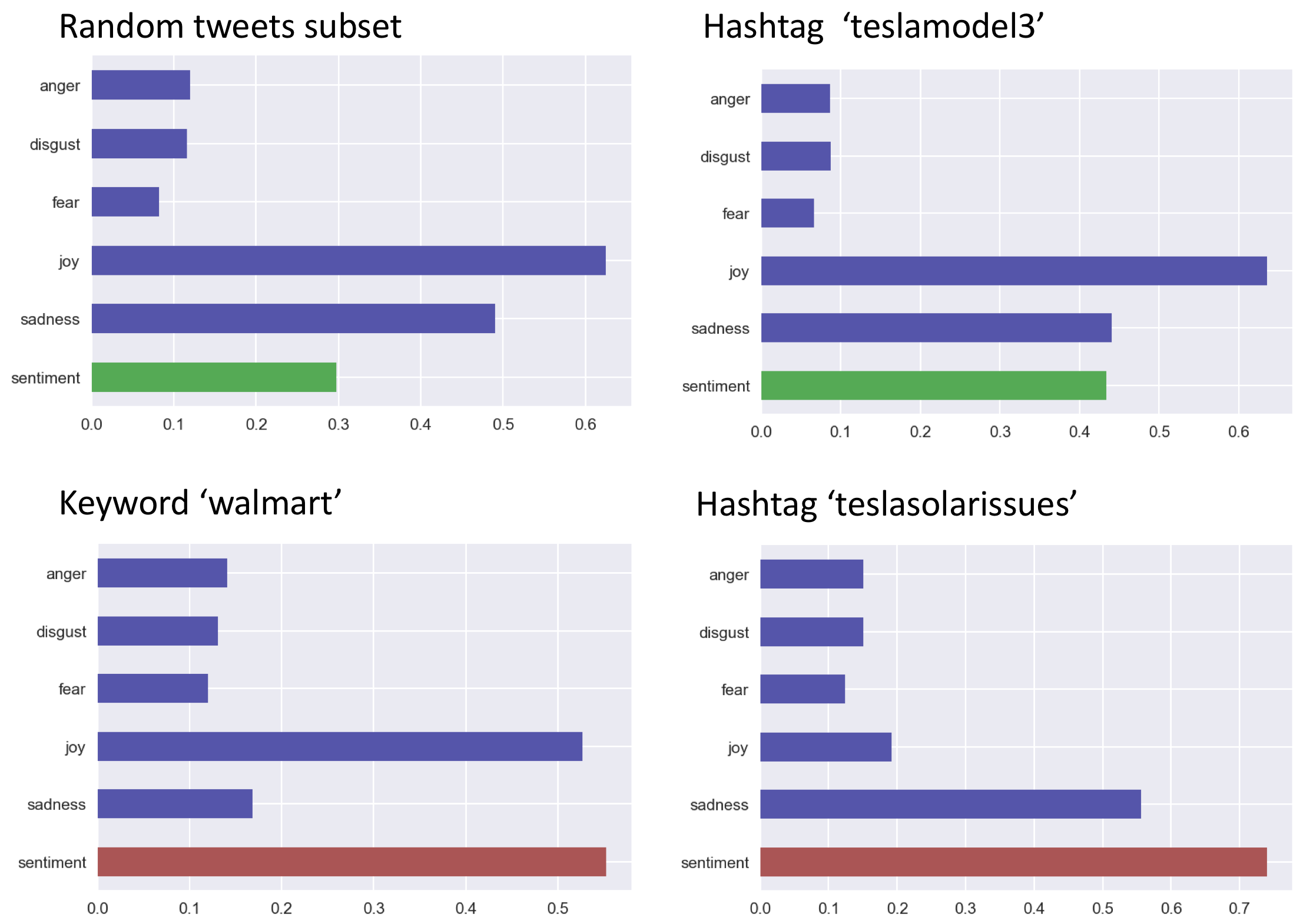}
\caption{Sentiment and personality analytics characteristics}
\label{sent_personalities}
\end{figure}

The frequent set  and associative rules theory is often used in the intelectual analysis~\cite{agrawal1994fast,agrawal1996fast,chui2007mining,gouda2001efficiently,
srikant1997mining,klemettinen1994finding,pasquier1999discovering,
brin1997beyond}.
 It can be used in a text data analysis to identify and analyze certain sets of objects, which are often found in large arrays and are characterized by certain features. 
Let's consider the algorithms for detecting frequent sets and associative rules on the example of processing microblog messages on Twitter. We can specify a thematic field which is a set of keywords semantically related to domain area under study.  Figure~\ref{tesla_freq} shows the frequencies of keywords   for  the thematic field of frequent itemsets analysis. This will make it possible to narrow the semantic analysis of messages to the given thematic framework. Based on the obtained frequent semantic sets, we are going to analyze  possible associative rules that reflect the internal semantic connections of thematic concepts in messages. In the time period when tweet dataset was being loaded, the accident with solar panels manufactured by Tesla on Walmart stores roofs took place. 
  It is important to consider the reflection of trends related to this topic in various processes, in particular, the dynamics of the company's stock prices in the financial market.
Using frequent itemsets and association rules, we can find a semantic structure in specified semantic fields of lexemes. 
Figures~\ref{tesla_tw_p3},~\ref{tesla_tw_p4}  shows semantic frequent itmesets for specified topics related to  Tesla company. Figures~\ref{tesla_tw_p5},~\ref{tesla_tw_p6} show association rules represented by graph and by grouped matrix.  
Figure~\ref{sent_personalities} shows sentiment and personality analytics characteristics received using IBM Watson Personality Insights~\cite{mahmud2016ibm}.
\subsection{Predictive analytics using tweet features}
\begin{figure}
\center
\includegraphics[width=1\linewidth]{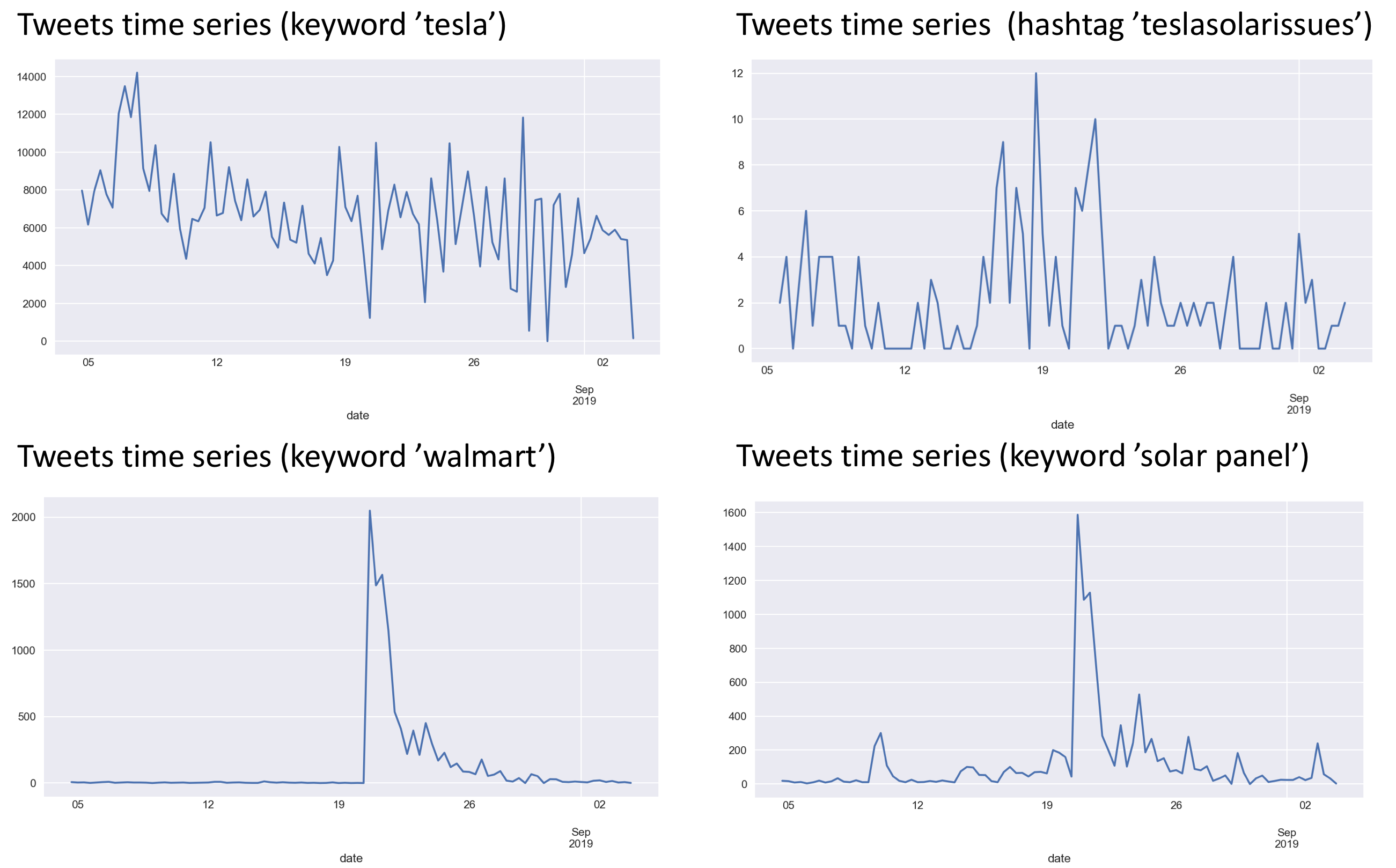}
\caption{Time series for different keywords and hashtags in the the tweets}
\label{kw_ts}
\end{figure}

\begin{figure}
\center
\includegraphics[width=1\linewidth]{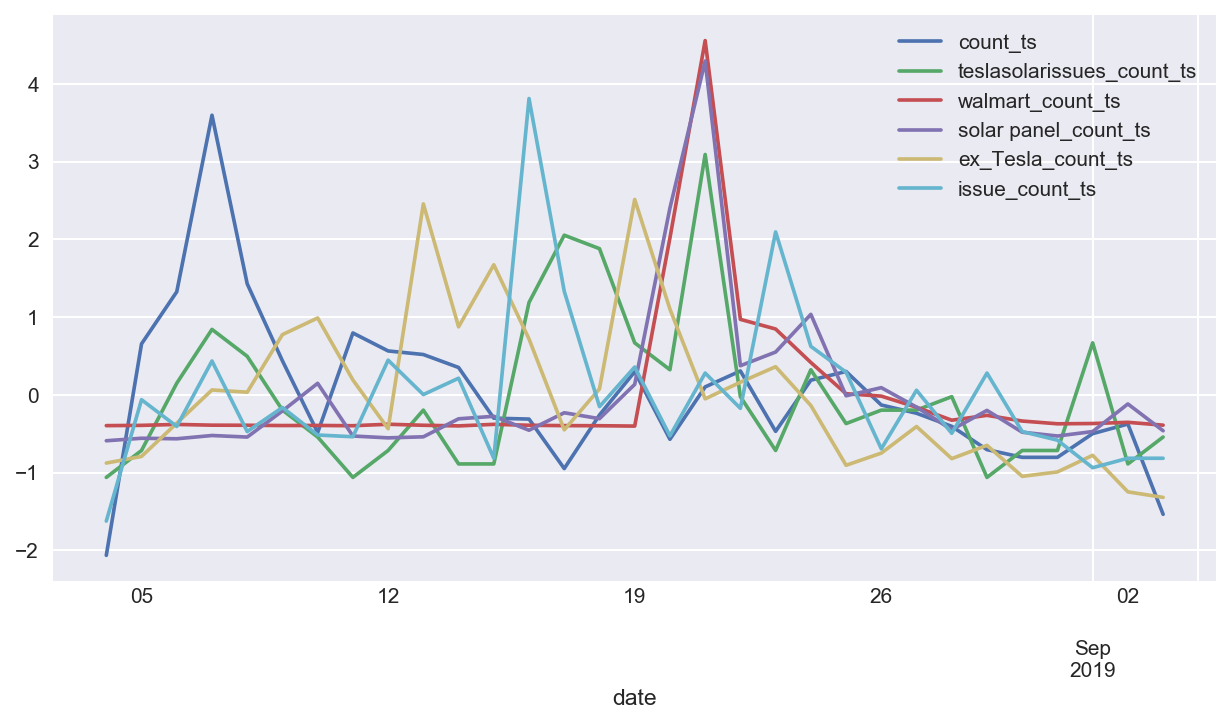}
\caption{Normalized keyword time series}
\label{tesla_kw_ts}
\end{figure}

\begin{figure}
\centerline{\includegraphics[width=0.85\textwidth]{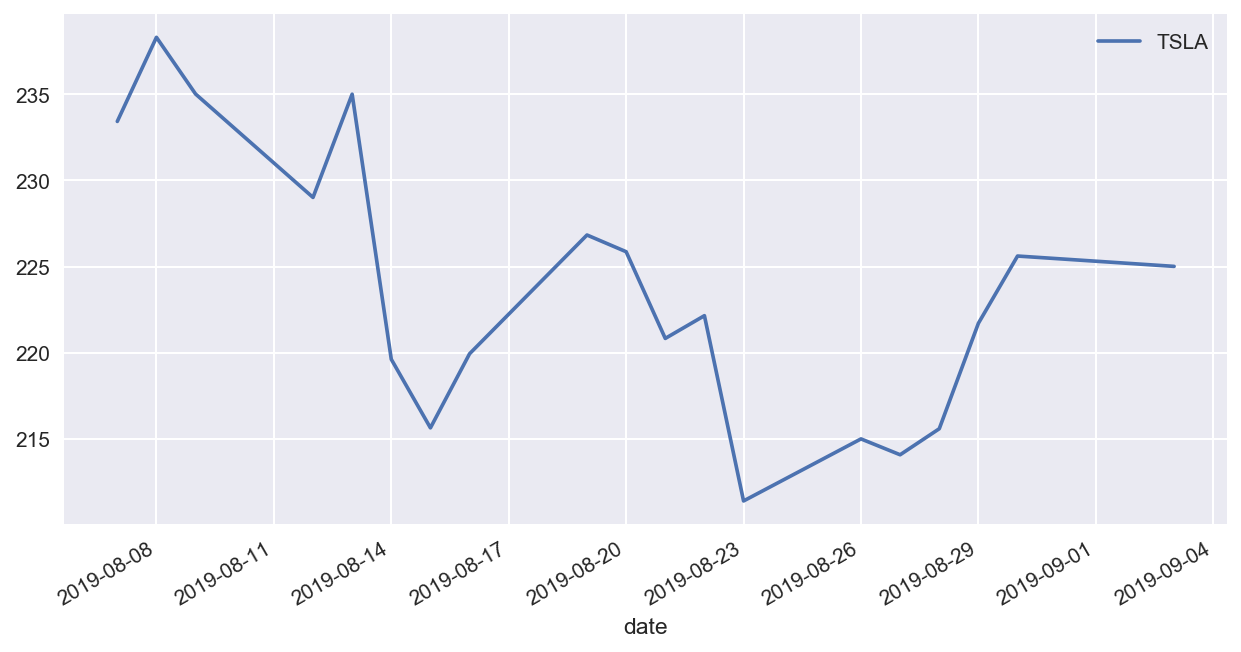}}
\caption{Dynamics of the stock price of \textit{Tesla} (TSLA ticker)}
\label{tesla_tw_p10}
\end{figure}

\begin{figure}
\center
\includegraphics[width=0.85\linewidth]{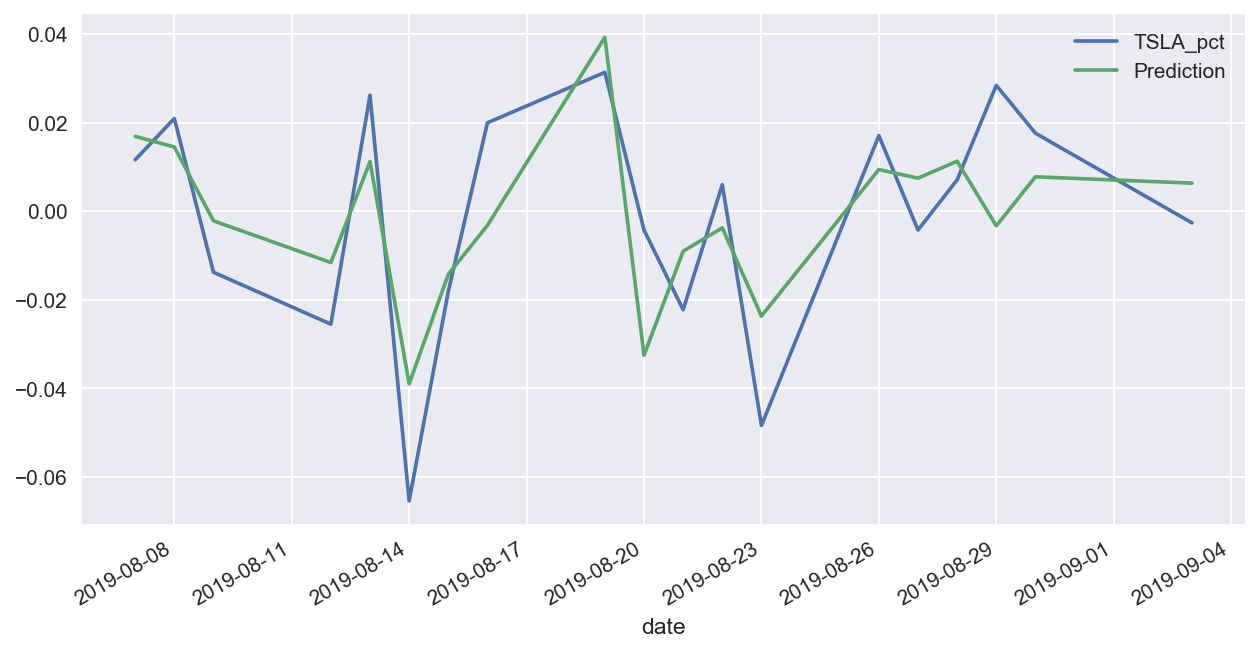}
\caption{Stock price return real and predicted values}
\label{tlsa_stock_ts_lasso_pred}
\end{figure}

\begin{figure}
\center
\includegraphics[width=0.75\linewidth]{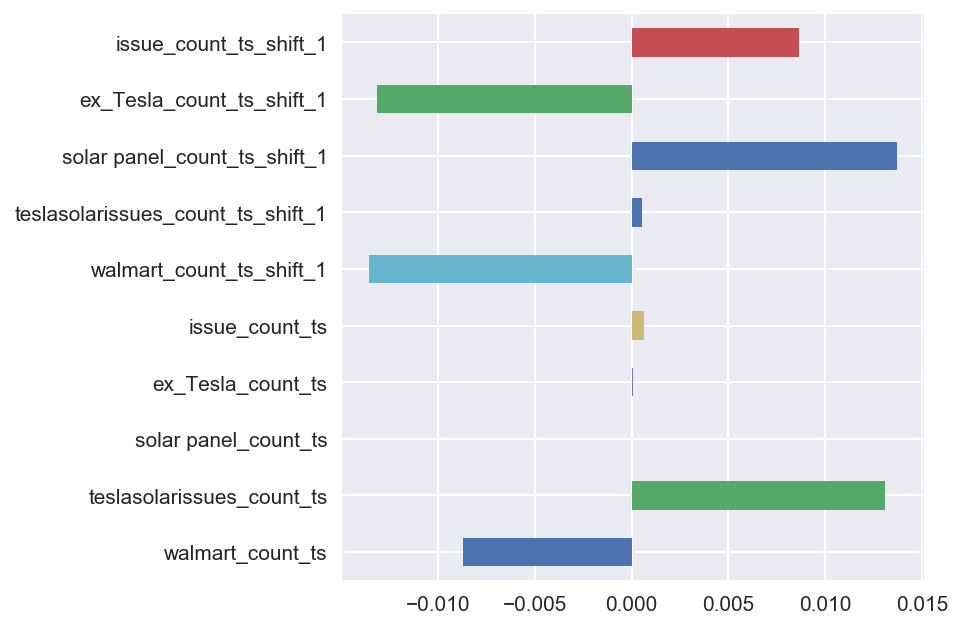}
\caption{Regression coefficients for the chosen features in the predictive model}
\label{tesla_stock_ts_lasso_coef}
\end{figure}

\begin{figure}
\center
\includegraphics[width=0.75\linewidth]{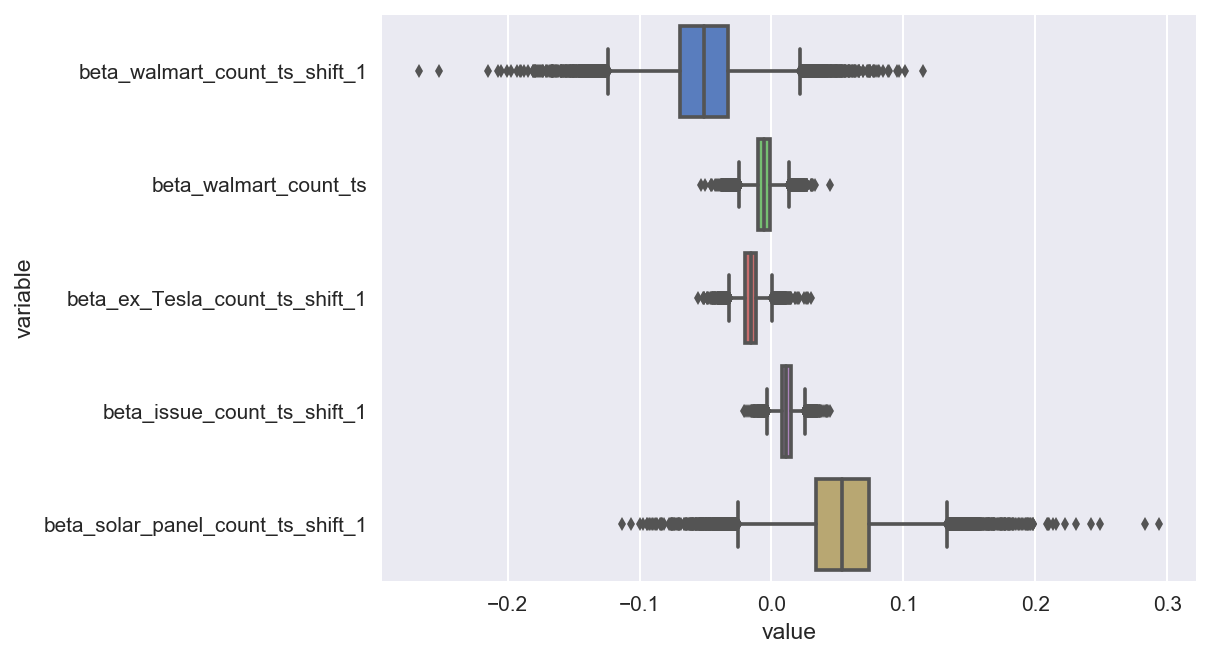}
\caption{Boxplots for feature coefficients in Bayesian regression model}
\label{boxplots_coef}
\end{figure}

Using revealed users' graph structure, semantic structure and topic related keywords and hashtags, one can receive keyword time series for tweet counts per day. These time series can be considered as features in the predictive models. In some time series, we can see when exactly the accident with solar panels on Walmart roof appeared and how long it was being considered in Twitter. Figure~\ref{kw_ts} shows the time series for different keywords and hashtags in the the tweets.
Figure~\ref{tesla_kw_ts} shows normalized keywords time series.
Social networks influence the formation of investment sentiment of potential stock market participants. Let us consider the dynamics of shares of the Tesla company  in the time period of the incident with solar panels  manufactured by Tesla.  It is reflected in the keywords time series on Figure~\ref{kw_ts}. One can see that at the time of the Tesla solar panel incident, the tweet activity is increasing over the time series of some keywords. Let us analyze how this incident affects the share price of Tesla. 
A linear model was created, where time series of keywords and their time-shifted values (lags) were considered as independent regression variables.
As a target variable, we considered the time series of the relative change in price during the day (price return).
Using LASSO regression, weights were found for the analyzed traits. 
Figure~\ref {tesla_tw_p10} shows the dynamics of the stock price Tesla (TSLA ticker) in the stock market.
We created a linear model where keyword time series and their lagged values were considered as covariates.  As a target variable, we considered stock price return time series for ticker TSLA. Using LASSO regression, we found weight coefficients for the features under consideration. Figure~\ref{tlsa_stock_ts_lasso_pred} shows the stock price return and predicted values.
Figure~\ref{tesla_stock_ts_lasso_coef} shows the regression coefficients for the chosen features in the predictive model.
We also conducted regression using Bayesian inference. Bayesian approach makes it possible to calculate the distributions for model parameters and for the target variable that is important for risk assessments~\cite{kruschke2014doing,gelman2013bayesian,carpenter2017stan}. 
Bayesian inference also makes it possible to  take into account non-Gaussian distribution of target variables that take place in many cases for financial time series. 
In~\cite{pavlyshenko2020bayesian}, we considered  different approaches of using Bayesian models for time series. Figure~\ref{boxplots_coef} shows the boxplots for feature coefficients in Bayesian regression model.
\subsection{Q-learning using tweet features}
\begin{figure}
\center
\includegraphics[width=0.85\linewidth]{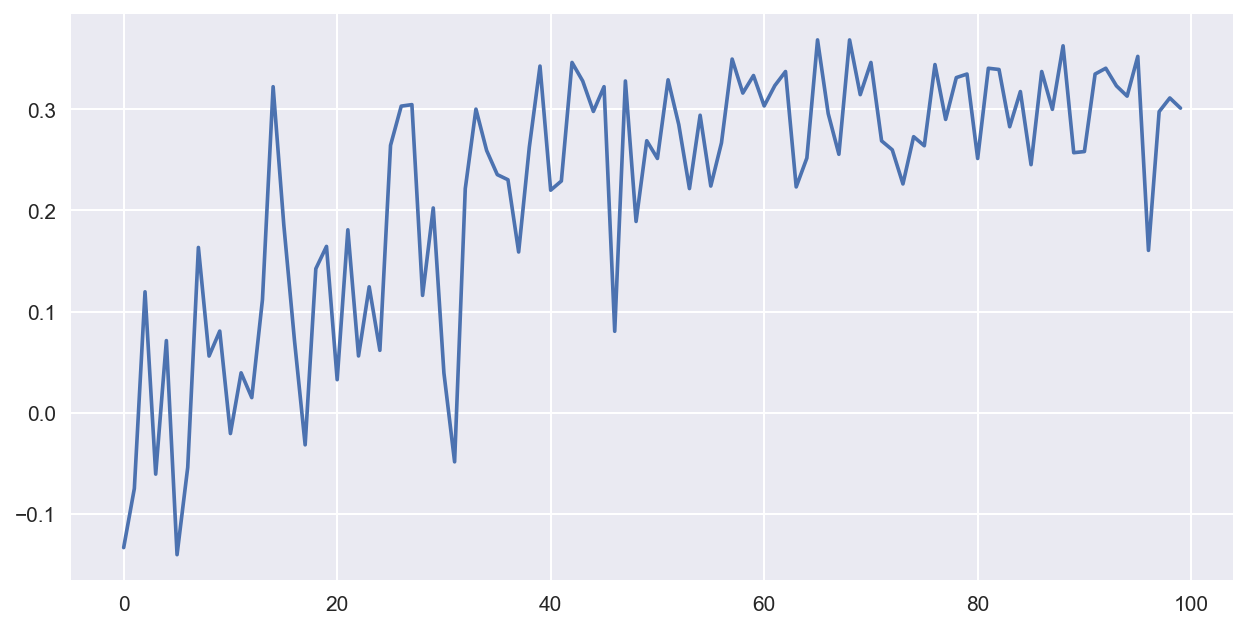}
\caption{Price return for the episodes for learning agent iterations}
\label{tsla_rl}
\end{figure}

It is interesting to use Q-learning to find an optimal trading strategy. 
Q-learning is an approach  based on the 
Bellman equation~\cite{sutton1998introduction,mnih2015human,mnih2013playing}.
In~\cite{pavlyshenko2020salests}, we considered different approaches for sales time series analytics using deep Q-learning.
Let us consider a simple trading strategy for the stocks with ticker TSLA.  In the simplest case of using deep Q-learning, we can apply three actions 'buy','sell','hold'. For state features, we used keyword time series. As a reward, we used stock price return. The environment for learning agent was modeled using keywords and reward time series. Figure~\ref{tsla_rl} shows the price return for the episodes for learning agent iterations.
The results show that an intelligent agent can find the an optimal profitable strategy. Of course, this is a very simplified case of analysis, where the effect of overfitting may occur, so this approach requires further study. The main goal is to show that, using reinforced learning and an environment model based on historical financial data and quantitative characteristics of tweets, it is possible to build a model in which an intelligent agent can find an optimal strategy that optimizes the reward function in episodes of  interaction of learning agent with the environment.  It was shown that time series of keywords features can be used as predictive features for different predictive analytics problems. Using Bayesian regression and tweets quantitative features one can estimate an uncertainty for the target variable that is important for the decision making support.
\subsection{Conclusions}
   Using the graph theory, the users' communities and influencers can be revealed given tweets characteristics. 
   The analysis of tweets, related to specified area, was carried out using frequent itemsets and association rules.
   Found frequent itemsets and association rules reveal the semantic structure of tweets related to a specified area. The quantitative characteristics of frequent itemsets and association rules, e.g. value of support, can be used as features in regression 
   models. Bayesian regression make it possible to assess the uncertainty of tweet features and target variable. It is shown 
   that tweet features  can also be used in deep Q-learning for forming the optimal strategy of learning agent e.g. in  the study of optimal trading strategies on the stock market.   
\FloatBarrier

 \phantomsection
\addcontentsline{toc}{section}{References}
\bibliographystyle{ieeetr}
\bibliography{references.bib}

\end{document}